\documentclass[journal]{IEEEtran}
\usepackage{times}
\usepackage{epsfig}
\usepackage{graphicx}
\usepackage{amsmath}
\usepackage{amssymb}
\usepackage{epstopdf}
\usepackage{array}
\newcolumntype{K}[1]{>{\centering\arraybackslash}m{#1}}
\newcommand{\ed}[1]{#1}

\graphicspath{ {figures/} }

\usepackage[pagebackref=true,breaklinks=true,letterpaper=true,colorlinks,bookmarks=false]{hyperref}
\ifCLASSINFOpdf
\else
\fi

\begin{document}
%
\title{\ed{Class-Aware Fully-Convolutional Gaussian and Poisson Denoising}}
%
%
%

\author{Tal~Remez,~\IEEEmembership{Member,~IEEE,}
        Or~Litany,~\IEEEmembership{Member,~IEEE,}
        Raja~Giryes,~\IEEEmembership{Member,~IEEE,}
        and~Alex~M. Bronstein,~\IEEEmembership{Member,~IEEE}
}

\maketitle

\begin{figure*}[bth!]
	\centering
    \begin{tabular}{c@{\hskip 0.01\textwidth}c@{\hskip 0.01\textwidth}c@{\hskip 0.01\textwidth}c}
		\includegraphics[width = 0.23\textwidth]{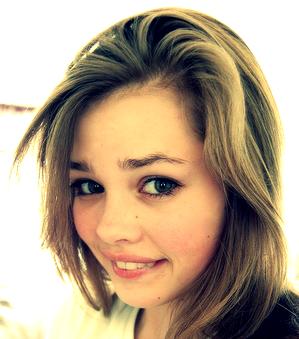} &
		\includegraphics[width = 0.23\textwidth]{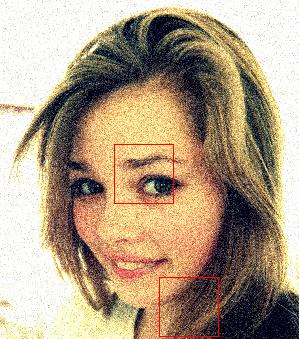} &
        \includegraphics[width = 0.23\textwidth]{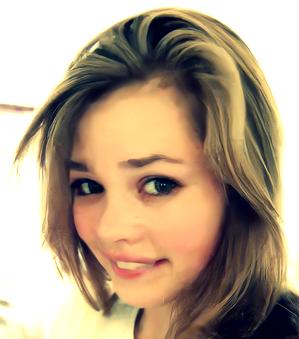} &
		\includegraphics[width = 0.23\textwidth]{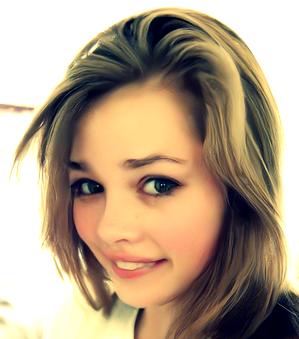} \\
        
	\end{tabular}  
   	\begin{tabular}{c@{\hskip 0.01\textwidth}c@{\hskip 0.01\textwidth}c@{\hskip 0.01\textwidth}c@{\hskip 0.01\textwidth}c@{\hskip 0.01\textwidth}c@{\hskip 0.01\textwidth}c@{\hskip 0.01\textwidth}c}
		\includegraphics[width = 0.11\textwidth]{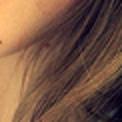} &
        \includegraphics[width = 0.11\textwidth]{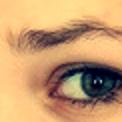} &
		\includegraphics[width = 0.11\textwidth]{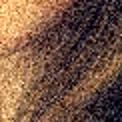} &
        \includegraphics[width = 0.11\textwidth]{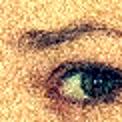} &
   		\includegraphics[width = 0.11\textwidth]{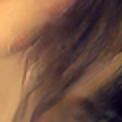} &
        \includegraphics[width = 0.11\textwidth]{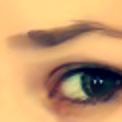} &
   		\includegraphics[width = 0.11\textwidth]{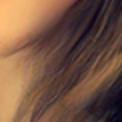} &
        \includegraphics[width = 0.11\textwidth]{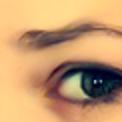} \\  
        
        \multicolumn{2}{c}{Ground truth image}&
        \multicolumn{2}{c}{Noisy image}&
        \multicolumn{2}{c}{\ed{Class-agnostic denoiser \cite{zhang2017learning}}}&
        \multicolumn{2}{c}{Our class-aware method}\\
        
        \multicolumn{2}{c}{}&
        \multicolumn{2}{c}{}&
        \multicolumn{2}{c}{\ed{$30.50$ dB}}&
        \multicolumn{2}{c}{\ed{$30.88$ dB}}\\
        
	\end{tabular}      
    \smallskip 
	\caption{\small \textbf{Perceptual comparison of class-aware and standard denoising.} Our proposed face-specific denoiser produces a visually pleasant result and avoids common artifacts caused by general-purpose denoisers. \ed{The input image is contaminated by Gaussian noise with $\sigma=25$.}}
\end{figure*}

\begin{abstract}
	We propose a fully-convolutional neural-network architecture for image denoising which is simple yet powerful. Its structure allows to exploit the gradual nature of the denoising process, in which shallow layers handle local noise statistics, while deeper layers recover edges and enhance textures. Our method advances the state-of-the-art when trained for different noise levels and distributions (both Gaussian and Poisson). 
In addition, we show that making the denoiser class-aware by exploiting semantic class information boosts performance, enhances textures and reduces artifacts.

\end{abstract}



\section{Introduction}
Denoising of additive white Gaussian noise and shot noise (Poisson-distributed) are fundamental problems in image enhancement. The Gaussian model is typically used for describing thermal noise and approximating shot noise in high- and medium-light imaging. In the low-light regime, which is largely dominated by shot noise, the Gaussian model fails to accurately represent this noise, and the more accurate Poisson \ed{model} is used instead. In addition to image denoising purposes, it has been shown in \cite{Chan16Plug, Romano16Little, Sreehari16Plug, venkatakrishnan2013plug, Tirer18Image} that a good Gaussian denoising algorithm may serve as a prior for efficiently solving many other image processing inverse problems such as deblurring or inpainting within a single generic framework.

In recent years, state-of-the-art in denoising of images has been achieved by techniques based on artificial neural networks \cite{burger2012image,Feng15Fast,Chen16Trainable,vemulapalli2016deep, zhang2017learning, Zhang17Beyond}. Following this promising trend, we propose a new fully-convolutional network architecture for image denoising that advances the state-of-the-art for both types of noise and most noise levels.
In contrast to previous methods, our network estimates the noise gradually through the composition of noise estimators extracted at intermediate layers. 

The latter fact makes intermediate results directly useful and allows \ed{a well timed} termination of the denoising process. 
In addition, it allows to inspect the inner workings of the network and gain \ed{an} insight into its operation. Interestingly, the network produces a monotonically decreasing error without being explicitly trained to do so.
%
%
Furthermore, being fully-convolutional it does not suffer from blocking artifacts characterizing \ed{non-overlapping patch-based approaches, or from the additional complexity required when overlapping patches are processed and averaged to improve quality and avoid the blockiness \cite{Guleryuz05Nonlinear1, Guleryuz05Nonlinear2, Elad06Image}}.

Having a good denoiser at hand, one may ask how it can be improved even further.
Patch-based image denoising theory suggests that existing methods have practically converged to the theoretical bound of the achievable performance \cite{Levin12Patch, Chatterjee10Denoising, levin2011natural,chatterjee2011practical}.
Yet, it turns out that two possibilities to break this barrier still exist. The first is to use larger patches \ed{which despite the \textit{law of diminishing return} for complex patches described in \cite{Levin12Patch}, } has been proven useful in \cite{burger2012image}, where the use of $39 \times 39$ patches allowed to outperform the popular BM3D denoising algorithm \cite{dabov2007image}. The second possibility is to use a better image prior, such as narrowing down the space of images to a more specific class. These two possibilities are not mutually exclusive, and indeed we show an approach that exploits them both.

To exploit larger patches, our network has a receptive field of size $41 \times 41$, which is bigger than the \ed{common} existing practice. The convolutional architecture is instrumental to prevent the network from becoming prohibitively large. 
To exploit a narrower image subspace, we draw inspiration from previous studies showing the benefit of designing a strategy for a specific class of images  \cite{baker2002limits,Bryt08Compression,Wang14Comprehensive,Joshi10Personal,Iizuka16Let, zhang2016colorful, Teodoro16Image, Niknejad17ClassGaussian, Niknejad17ClassPoisson, Ljubenovic17Blind, Fang17Automatic, Fang17Segmentation}.
To this end, we propose a class-aware denoising framework, in which the denoiser is fine-tuned to best fit a particular class of images. At inference, the class information can be provided manually by the user, for example, choosing face denoising for cleaning a personal photo collection, or automatically via a classification algorithm. We show that class-awareness indeed boosts performance significantly when the class is given by an oracle. Moreoever, combining our denoiser with an off-the-shelf classifier, fine-tuned to our classes, instead of the oracle leads to comparable performance. 

\medskip 

\noindent {\bf Contribution.} \, 
Our contribution is threefold:
\begin{enumerate}
\item  We propose a novel fully-convolutional neural-network architecture that is \ed{comparable to} the state-of-the-art for Gaussian and Poisson image denoising and Poisson video denoising.
\item Our network design grants easy access to the noise being removed at intermediate layers, allowing interesting insights into its inner workings.
\item We demonstrate an additional boost in performance when classifying the input image before routing it to a class-specific denoiser. 
\end{enumerate} 

\noindent While this paper focuses on denoising, our methodology can be easily extended to much broader class-aware image enhancement and restoration problems, rendering it applicable to many image processing and computer vision tasks.
Preliminary results of this paper for Gaussian image denoising have been presented in \cite{Remez17Deep}. We show how the proposed deep architecture can be further used for Poisson denoising, demonstrating examples of simulated and real low-light images and videos.

The rest of the paper is organized as follows: Section \ref{sec_related} surveys related work. In Section \ref{sec_method}, we describe our denoising architecture and show how it can be made class-aware. Section \ref{sec_implementation} includes a  description of our implementation. Section \ref{sec_exp} presents an experimental evaluation of class-agnostic and class-aware denoising for Gaussian and Poisson noise.

\section{Related work}
\label{sec_related}
Numerous methods have been proposed for removing Gaussian noise from images, including
$k$-SVD \cite{Aharon06KSVD}, non-local means \cite{Buades05Non}, BM3D \cite{dabov2007image} non-local $k$-SVD \cite{Mairal09Non}, field of experts (FoE) \cite{Schmidt10Generative}, Gaussian mixture models (GMM) \cite{Yu12Solving}, non-local Bayes \cite{Lebrun13Nonlocal}, global image denoising \cite{talebi2014global}, nonlocally centralized sparse representation (NCSR) \cite{Dong13Nonlocally} and simultaneous sparse coding combined with Gaussian scale mixture (SSC-GSM) \cite{Dong15Image}.

A popular strategy for recovering images contaminated by Poisson noise relies on variance-stabilizing transforms (VST), such as Anscombe and Fisz \cite{anscombe48transformation,Fisz55Limiting,Zhang08Wavelets}, which convert Poisson noise to be approximately  Gaussian with unit variance. Thus, it is possible to perform Poisson denoising using Gaussian denoisers, e.g., by VST+BM3D \cite{makitalo2011optimal}. Recently, it has been shown that it is possible to improve this method by iteratively applying the VST and the Gaussian denoiser (I+VST+BM3D) \cite{Azzari16Variance}. However, such approaches are limited as their performance deteriorate significantly for low intensities.
Another approach is to cast the Poisson denoising problem to a Gaussian one using the plug-and-play framework \cite{venkatakrishnan2013plug} as suggested in the P$^4$IP method \cite{rond2016poisson}.

Alternatively, there are methods that are applied directly to the Poisson noisy data 
such as the non-local sparse PCA (NLSPCA) technique \cite{Salmon13Poisson} or the sparse coding based Poisson denoising algorithm (SPDA)  \cite{Giryes14Sparsity}.

A popular strategy improving the performance of many Poisson denoising algorithms is binning \cite{Salmon13Poisson}. Instead of processing the noisy image directly, a low-resolution version of the image with higher SNR is generated by aggregation of nearby pixels. A Poisson denoising technique is then applied followed by simple linear interpolation to recover the original high resolution image. The binning technique trades-off spatial resolution and SNR, and has been shown to be useful in the very low SNR regimes. 

Most aforementioned techniques are designed based on some properties of natural images such as the recurrence of patches at different locations and scales, or their sparse representation in some (possibly trained) dictionary.
In the past few years, however, the state-of-the-art in image denoising has been achieved by techniques based on artificial neural networks. The first such method is the multilayer perceptron (MLP) proposed in \cite{burger2012image}. It is based on a fully connected architecture and therefore requires a 
long training time, a large amount of memory, and has a high arithmetic complexity at inference.
In \cite{Feng15Fast} (TRDPD) and \cite{Chen16Trainable} (TNRD), nonlinear diffusion based neural networks have been proposed for Poisson and Gaussian denoising respectively. Another work \cite{vemulapalli2016deep}, proposes a neural network based on a deep Gaussian Conditional Random Field (DGCRF) model. \ed{More recently, in parallel to our work, a convolutional neural network with batch normalization and without intermediate noise estimations has been proposed (IRCNN) \cite{zhang2017learning}.}

Designing a strategy for a specific class of images has been shown to be beneficial for many image enhancement applications.
For example, in \cite{baker2002limits}, the authors set a bound on super-resolution performance and showed it can be broken when a face-prior is used. In \cite{Bryt08Compression}, a compression algorithm for facial images was proposed. Face hallucination, super-resolution, and sketch-photo synthesis methods were developed by \cite{Wang14Comprehensive}. In \cite{Joshi10Personal}, the authors showed that given a collection of photos of the same person, it is possible to obtain a more realistic reconstruction of the face from a blurry image.  In \cite{Iizuka16Let, zhang2016colorful}, class labeling at a pixel-level was used for the colorization of gray-scale images. 

The closest class-specific approach to our work is the one proposed in \cite{Teodoro16Image, Niknejad17ClassGaussian, Niknejad17ClassPoisson, Ljubenovic17Blind}. It shows improvement in image restoration tasks (denoising and deblurring) by adapting the reconstruction algorithm to a specific class of images. The proposed strategy is patch based and assumes \emph{a priori} \ed{knowledge} of the class. Our solution circumvents these two limitations by (i) using a fully convolutional neural network that processes the image as a whole; and (ii) a classification network that can detect the class of the image. 

\section{Method}
\label{sec_method}
\subsection{Class-agnostic network architecture}
\label{sec_method_agnostic}
Our network structure was designed according to the following principles: First, it is fully convolutional to facilitate denoising images of varying size. This eliminates the use of image patches which, in turn, entails carefully choosing an aggregation procedure for the denoised overlapping patches. 
Two advantages achieved by such a paradigm are the relatively small number of parameters and fast execution time as demonstrated in the experiments in Section \ref{sec_exp}. A second requirement was to have a gradual denoising process since this has been shown by \cite{Romano15Boosting, Zoran11From, Sulam15Expected} to yield better denoising.

\begin{figure}[tb] 
\includegraphics[width=0.97\linewidth]{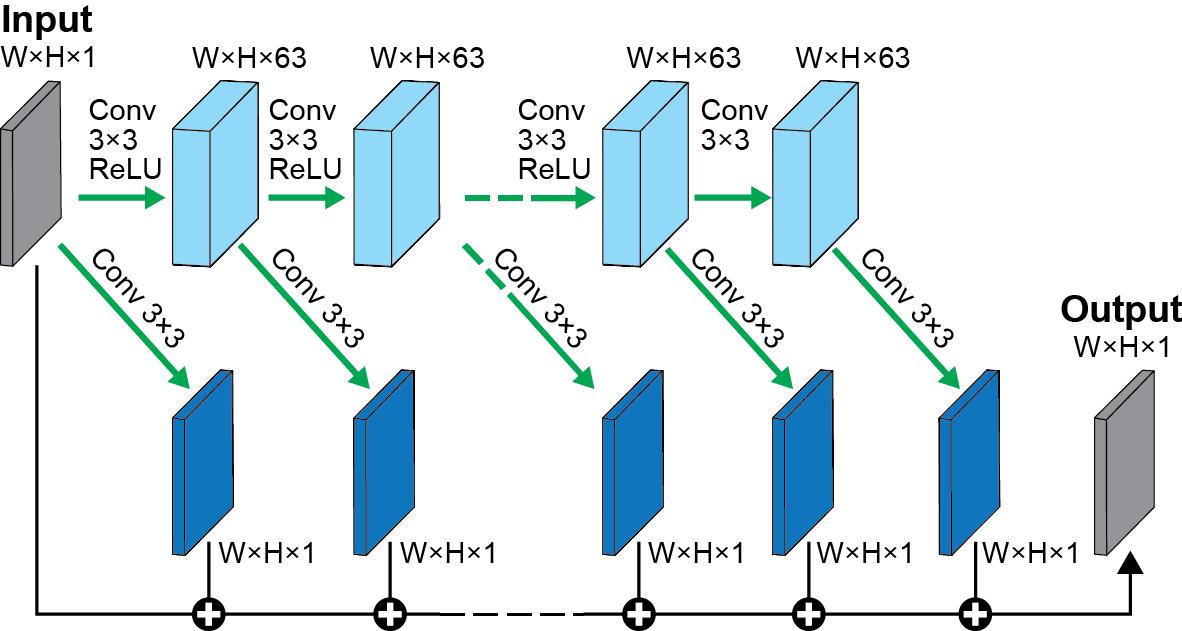}
\caption{\small \textbf{Network architecture.} 
Each layer performs $63$ convolutions followed by a \textit{ReLU} and forwarded to the next layer (light blue). An additional convolution yields a matrix (dark blue), which is added to the input image. Tensor sizes are listed as $Width \times Height \times \# Channels$.}
\label{fig_denoiseNet}
\end{figure}

Our network architecture is presented in Fig.~\ref{fig_denoiseNet}. In each layer, we perform $63$ convolutions (marked in light blue) followed by a \textit{ReLU} and forwarded to the next layer. We additionally perform a single \ed{$3\times 3$ convolution, whose weights are learned together with the other parameters of the network. This convolution yields} a matrix of the same size as the input image (marked in dark blue in the figure). This matrix is added to the noisy input and is not forwarded to the next layer. We refer to these single channel matrices as \emph{noise estimates} because their sum cancels out the noise. \ed{In all layers we use convolutions with a fixed size of $3\times3$ and stride $1$}. In Section \ref{sec_visualizing}, we visualize the noise estimates at different layer depths to gain insights into what the network has learned.

\ed{As the sum of the layers is added to the noisy input image in order to produce the clean image, the effect these layers produce is noise removal instead of directly estimating the clean signal. The fact that only a fraction of the noise is removed in each layer of the network makes the propagation of the information easier as the signal is recovered gradually and not all at once. This provides some kind of boosting in the recovery, which is shown to be helpful also in the context of conventional image denoising \cite{Romano15Boosting, Zoran11From}.}


In the case of Poisson noise, we do not apply the binning or the Anscombe transform as the common practice suggests \cite{makitalo2011optimal, Salmon13Poisson, Makitalo14Noise}, since this pre-processing did not show any performance improvement of the network. 

\subsection{Class-aware denoising}
\label{sec_method_aware}
Class-aware denoising requires handling each class in a specialized manner. Our method comprises two stages: First, the image is classified into one of the supported classes using a classifier. Then, it is fed into a class-specific denoising network of the chosen class. In all class-aware experiments, we used the following five semantic classes: \emph{face}, \emph{pet}, \emph{flower}, \emph{living-room}, and \emph{street}.

As a classifier, we used the pretrained weights of the convolution layers of VGG16 \cite{simonyan2014very} and trained four additional fully connected layers of sizes $1024$, $1024$, $1024$, and $5$ with \textit{ReLU} and drop-out, and a soft-max layer at the very end.


Each of our class-specific densoisers has the same architecture as described before and is trained using ImageNet images \cite{ImageNet15} from a specific (semantic) class. 
In the experiments we show that the described two-stage system boosts performance significantly
when the class is given by an oracle, and that
replacing the oracle with a classifier does not deteriorate the gain in the performance.

\section{Implementation details}
\label{sec_implementation}
In all experiments we used denoising networks with $20$ layers implemented in TensorFlow \cite{abadi2015tensorflow} and trained for $160K$ mini-batches on a Titan-X GPU. The mini-batches contained $64$ patches of size $128 \times 128$. Color images were converted to YCbCr, and the Y channel was used as the input grayscale image after it had been scaled and shifted to the range $[-0.5,0.5]$ and contaminated with simulated noise. During training, image patches were randomly cropped and flipped about the vertical axis. 
\ed{As the receptive field of the network is of size 41, the outer 21 pixels of the input suffer from convolution artifacts. To avoid these artifacts in training, we calculate the loss of the network only at the central part of each patch used for training, not taking into account the outer 21 pixels in the calculation of the loss. At test time, we pad the image symmetrically by 21 pixels before applying our network and remove these added shoulders from the denoised image. We used an $\ell_2$-loss in all experiments.}
Training was performed with the ADAM optimizer \cite{DBLP:journals/corr/KingmaB14} with a learning rate $\alpha=10^{-4}$, $\beta_1=0.9$, $\beta_2=0.999$ and $\epsilon=10^{-8}$.\footnote{\ed{Code available at \url{github.com/TalRemez/deep_class_aware_denoising}}.}

In the class-aware setup, the classifier was trained on noisy grayscale images with the same noise statistics used at test time after downsampling the image-resolution to a fixed size of $128 \times 128$ pixels. Downsampling the image allows faster classification and better robustness to noise. The classifier was trained for $10K$ mini-batches of $64$ images using ADAM optimizer with a learning rate of $\alpha=10^{-4}$, $\beta_1=0.9$, $\beta_2=0.999$ and $\epsilon=10^{-8}$, and the categorical cross-entropy loss. Keep probability of $0.5$ was used with dropout. 

\section{Experiments}
\label{sec_exp}

\subsection{Class-agnostic denoising}
\label{sec_exp_agnostic}
This section evaluates the performance of our class-agnostic method on Gaussian and Poisson noise using the following test sets: (i) images from PASCAL VOC \cite{pascal-voc-2010}; (ii) the commonly used $68$ test images chosen by \cite{roth2009fields} from the Berkeley segmentation dataset \cite{MartinFTM01}; and (iii) for Poisson noise we additionally evaluated performance on the commonly used $10$ images test set as well as on a real low-light image. In all experiments of this section, a separate network was trained for each noise type and level using $8K$ images from PASCAL \cite{pascal-voc-2010}. 

\subsubsection{Gaussian denoising}
\label{sec_agnostic_gaussian}
\begin{table}[t]
\centering
\small
\begin{tabular}{ l@{\hskip 0.01\textwidth}c@{\hskip 0.01\textwidth}c@{\hskip 0.01\textwidth}c@{\hskip 0.01\textwidth}c@{\hskip 0.01\textwidth}c@{\hskip 0.01\textwidth}c@{\hskip 0.01\textwidth}c  }
    \hline\hline
    $\sigma$	& $10$ 	  			& $15$ 			& $25$ 				& $35$ 				& $50$ 				& $65$ 				& $75$ \\ \hline
    BM3D		& $34.26$ 			& $32.10$ 		& $29.62$ 			& $28.14$ 			& $26.61$ 			& $25.64$ 			& $25.12$ \\  
    MLP 		& $34.29$ 			& $-$ 			& $29.95$ 			& $28.49$ 			& $26.98$ 			& $26.07$ 			& $25.54$ \\  
    TNRD		& $-$     			& $32.35$ 		& $29.90$ 			& $-$ 				& $26.91$ 			& $-$ 				& $-$ \\  
    IRCNN		& $-$     	   		& $32.66$ 		& $30.25$ 			& $-$ 				& $27.22$ 			& $-$ 				& $-$ \\ 
    Ours 		& $\textbf{34.87}$ & $\textbf{32.79}$ & $\textbf{30.33}$ & $\textbf{28.88}$ & $\textbf{27.32}$ & $\textbf{26.30}$ & $\textbf{25.75}$	\\ \hline \hline
  \end{tabular}  
  \vspace{2mm}
\caption{\small \textbf{Gaussian denoising on PASCAL.} Average PSNR values on a $1K$ image test-set.}
\label{tab_pascal_gaussian}	
\end{table}

\begin{figure}[t] 
\vspace{-0.2in}
\includegraphics[width=0.48\textwidth]{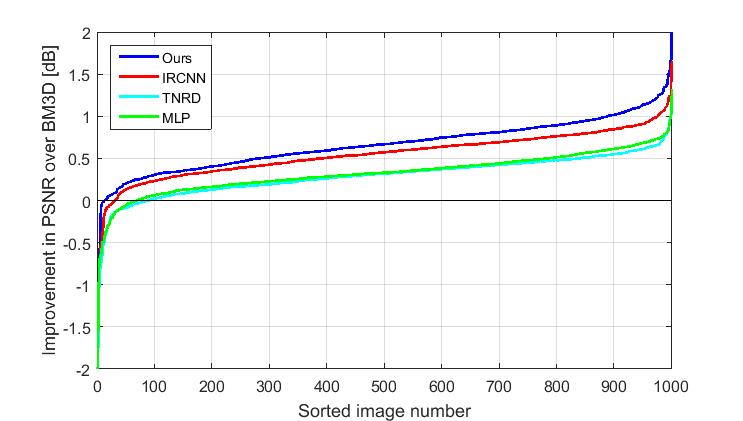}
\caption{\small\textbf{Gaussian denoising performance profile relative to BM3D on PASCAL.}  Image indices are sorted in ascending order of performance gain relative to BM3D. Our improvement is demonstrated by (i) decrease of the zero-crossing point, and (ii) consistently higher values of gain. The comparison was made on $1K$ images for $\sigma=25$.}
\label{pascal_s_curve_g}
\vspace{-0.15in}
\end{figure}

\begin{table}[tb]
\centering
\small
\begin{tabular}{ l@{\hskip 0.01\textwidth}c@{\hskip 0.01\textwidth}c@{\hskip 0.01\textwidth}c@{\hskip 0.01\textwidth}c@{\hskip 0.01\textwidth}c@{\hskip 0.01\textwidth}c@{\hskip 0.01\textwidth}c  }
    \hline\hline
    $\sigma$ & $10$ 	& $15$ 		& $25$ 		& $35$ 		& $50$ 		& $65$ 		& $75$ \\ \hline
    BM3D	 & $33.31$ 	& $31.10$ 	& $28.57$ 	& $27.08$ 	& $25.62$ 	& $24.68$ 	& $24.20$ \\  
    MLP 	 & $33.50$ 	& $-$ 		& $28.97$ 	& $27.48$ 	& $26.02$ 	& $25.10$ 	& $24.58$ \\  
    TNRD	 & $-$ 		& $31.41$ 	& $28.91$ 	& $-$ 		& $25.95$ 	& $-$ 		& $-$ \\ 
    IRCNN	 & $-$    	& $\textbf{31.63}$ 		& $\textbf{29.14}$ 	& $-$ 		& $\textbf{26.17}$ 	& $-$ 		& $-$ \\ 
    Ours 	 & $\textbf{33.58}$ & $31.44$ & $29.05$ & $\textbf{27.56}$ & $26.06$ & $\textbf{25.12}$ & $\textbf{24.61}$	\\ \hline \hline
  \end{tabular}   
  \vspace{2mm}
\caption{\small \textbf{Gaussian denoising on $68$ image set from \cite{roth2009fields}.} Average PSNR values are presented. 
\vspace{-4mm}}
\label{tab_bsds_gaussian}	
\end{table}

We compare our network performance at removing Gaussian noise with standard deviation values between $10$ and $75$ against the following leading methods: (i) BM3D \cite{dabov2007image}; (ii) multilayer perceptrons (MLP) \cite{burger2012image}; (iii) TNRD \cite{Chen16Trainable}; and (iv) IRCNN \cite{zhang2017learning} \ed{using the pretrained models provided online}. For MLP, TNRD and IRCNN we used the publicly available pre-trained, noise level-specific models. We tested our denoising algorithms on a test set of $1K$  images from PASCAL \cite{pascal-voc-2010}. Table \ref{tab_pascal_gaussian} summarizes the performance in terms of average PSNR. 

\ed{Note that the IRCNN pre-trained network for Gaussian noise provided by the authors of \cite{zhang2017learning} is trained on the BSD dataset and not PASCAL. 
However, as shown in \cite{godard2017deep}, our network still gets better performance than IRCNN even when the latter is trained on  PASCAL. In the following sections we show that this result is consistent when we re-train the IRCNN network for other scenarios (such as Poisson noise) and compare to our network.}

Figure \ref{pascal_s_curve_g} emphasizes the statistical significance of the improvement achieved by our method for Gaussian denoising with $\sigma=25$. It compares the gain in performance over BM3D achieved by our method with the one achieved by MLP, TNRD and IRCNN. The plot clearly visualizes the consistent improvement in PSNR achieved by our method. Additionally, for $\sigma=25$, our method outperforms competing methods on $82\%$ of the images, whereas IRCNN, MLP, BM3D, and TNRD win on $13\%, 4\%, 1\%$ and $0\%$ respectively. 


Next, we tested how well our network performs on the widely used test set of $68$ images selected by \cite{roth2009fields} from Berkeley segmentation dataset \cite{MartinFTM01}. Table \ref{tab_bsds_gaussian} shows that our method generalizes well to this dataset, for all $\sigma$ values.

\subsubsection{Poisson denoising}
\label{sec_agnostic_poisson}
We evaluated our method on peak values ranging from $1$ up to $30$ and compared its performance to leading methods. Whenever code for other techniques was not publicly available, we evaluated our method on the same test set they had been tested on, and compared our performance with their reported scores. 
\ed{We reimplemented IRCNN and trained it on the exact same dataset that we used with our network. We tried it with and without batch normalization. We found that batch normalization reduced its PSNR performance by $0.15$ dB on average for the different peak values. A similar phenomenon was observed in \cite{godard2017deep} in the context of image burst denoising. Therefore we only present the results for the architecture that does not have batch normalization.}

\begin{figure}[t] 
\includegraphics[width=0.48\textwidth]{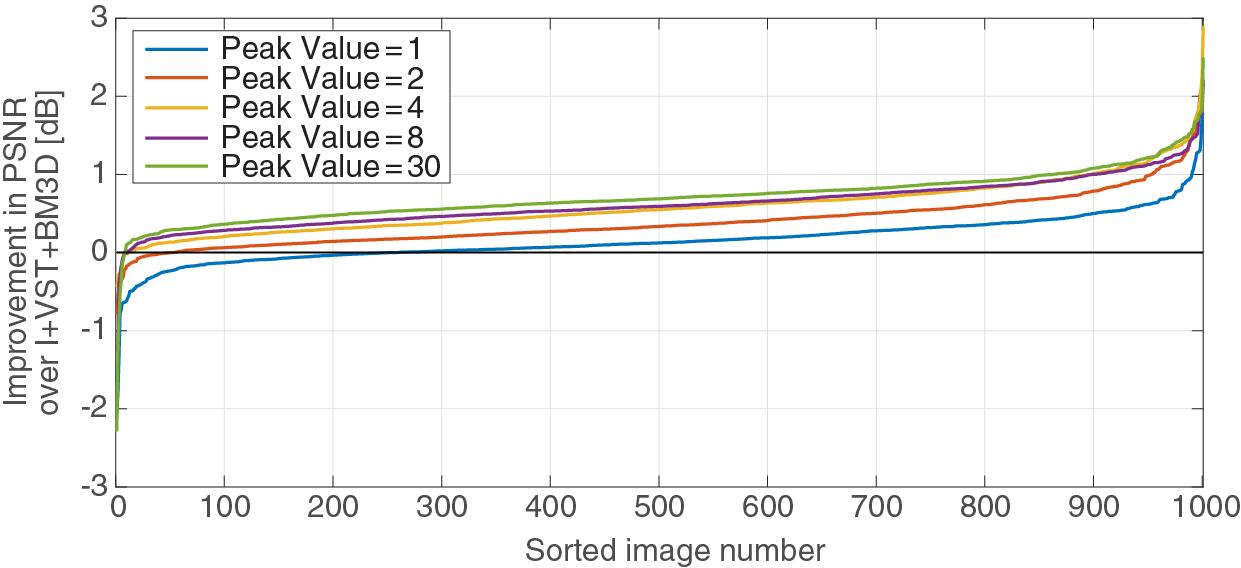}
\caption{\textbf{Poisson denoising performance profile relative to I+VST+BM3D on PASCAL.}  Image indices are sorted in ascending order of our performance gain relative to I+VST+BM3D. The improvement of our method is demonstrated by (i) small zero-crossing point, and (ii) consistently higher PSNR values.
The comparison was made using $15$ noise realizations per image. Our method outperforms I+VST+BM3D on $74\%$, $94.8\%$, $98.9\%$, $99$, and $99.2\%$ of the images for peak values $1$ to $30$ respectively.} 
\label{pascal_s_curve_poisson}
\end{figure}

Similarly to the Gaussian noise, we first evaluated our method on $1K$ test images from PASCAL \cite{pascal-voc-2010}. Table \ref{tab_pascal_poisson} summarizes performance in terms of average PSNR.
\begin{table}[t]
\centering
\small
  \begin{tabular}{ l@{\hskip 0.01\textwidth}c@{\hskip 0.01\textwidth}c@{\hskip 0.01\textwidth}c@{\hskip 0.01\textwidth}c@{\hskip 0.01\textwidth}c@{\hskip 0.01\textwidth}c@{\hskip 0.01\textwidth}c@{\hskip 0.01\textwidth}c  }
    \hline \hline
    	Peak 		& $1$ 	& $2$ 	& $4$ 	& $8$ 	& $30$ \\ \hline
		I+VST+BM3D  & $22.71$ & $23.70$ 	& $24.78$ 	& $26.08$ 	& $28.85$ \\
        IRCNN		& \ed{$22.70$} & \ed{$23.98 $} & \ed{$25.22$} & \ed{$26.58$} & \ed{$29.44$} \\
    	Ours    & $\textbf{22.87}$ & $\textbf{24.09}$ 	& $\textbf{25.36}$ 	& $\textbf{26.70}$ 	& $\textbf{29.56}$ \\
     \hline\hline
  \end{tabular}  
  \vspace{2mm}
\caption{\small \textbf{Poisson denoising on PASCAL.} Average PSNR values for different peak values on $1K$ test images and $15$ noise realizations per image.} 
\label{tab_pascal_poisson}	
\end{table}
Figure \ref{pascal_s_curve_poisson} shows a profile of the gain in performance of our method with respect to I+VST+BM3D \ed{\cite{Azzari16Variance}} for all peak values. 

\begin{table}[tb]
\centering
\small
  \begin{tabular}{ l@{\hskip 0.01\textwidth}c@{\hskip 0.01\textwidth}c@{\hskip 0.01\textwidth}c@{\hskip 0.01\textwidth}c@{\hskip 0.01\textwidth}c@{\hskip 0.01\textwidth}c@{\hskip 0.01\textwidth}c  }
    \hline \hline
		Peak 				 & $1$ 				& $2$ 				& $4$ 				& $8$ 				\\ \hline
		NLSPCA 				 & $20.90$			& $21.60$			& $22.09$			& $22.38$			\\
        NLSPCA bin  		 & $19.89$			& $19.95$	 		& $19.95$			& $19.91$			\\
        VST+BM3D 			 & $21.01$			& $22.21$			& $23.54$			& $24.84$			\\
        VST+BM3D bin 		 & $21.39$			& $22.14$			& $22.87$			& $23.53$			\\
        I+VST+BM3D 		 	 & $21.66$			& $22.59$			& $23.69$			& $24.93$			\\
        TRDPD$^8_{5\times5}$ & $21.49$			& $22.54$			& $23.70$			& $24.96$			\\
        TRDPD$^8_{7\times7}$ & $21.60$			& $22.62$			& $23.84$			& $25.14$	 		\\
        \ed{IRCNN}			 &\ed{$21.66$}		&\ed{$22.86$}		&\ed{$\textbf{24.00}$}	&\ed{$25.27$} \\
    	Ours    		 & $\textbf{21.79}$ & $\textbf{22.90}$ 	& $ 23.99 $ 	& $\textbf{25.30}$ 	\\ 
     \hline\hline
  \end{tabular}  
  \vspace{2mm}
\caption{\small \textbf{Poisson denoising on $68$ image set from \cite{roth2009fields}.} Average PSNR values of $15$ noise realizations for different peak values on $68$ images. Results for NLSPCA, VST+BM3D and TRDPD reported in \cite{Feng15Fast} were reproduced with the addition of the values for our method and for I+VST+BM3D. 
} 
\label{tab_bsds_poisson}	
\end{table}

\begin{table*}[h]
\small
\centering
  \begin{tabular}{ l@{\hskip 0.01\textwidth}|c@{\hskip 0.01\textwidth}|c@{\hskip 0.01\textwidth}c@{\hskip 0.01\textwidth}c@{\hskip 0.01\textwidth}c@{\hskip 0.01\textwidth}c@{\hskip 0.01\textwidth}c@{\hskip 0.01\textwidth}c@{\hskip 0.01\textwidth}c@{\hskip 0.01\textwidth}c@{\hskip 0.01\textwidth}c@{\hskip 0.01\textwidth}|l@{\hskip 0.01\textwidth}  }
    \hline \hline
    Method &  Peak &  Flag &  House &  Cam &  Man &  Bridge &  Saturn &  Peppers &  Boat &  Couple &  Hill &  Time \\
 \hline
 \hline

NLSPCA &  	& $19.68$ & $21.57$ & $20.25$ & $21.46$ & $19.02$ & $24.75$ & $19.5$ & $21.19$ & $21.14$ & $21.94$ & $86$s \\

NLSPCA bin &	& $15.77$ & $20.78$ & $18.4$ & $19.87$ 	& $18.26$ & $22.83$ & $17.78$ & $20.19$ & $20.11$ & $20.82$ & $16$s \\

SPDA &			& $\textbf{22.97}$ & $22.14$ & $20.15$ & - 		& $19.30$  & $27.05$ & $19.97$ & - 		& - 		& - 	& $5$h \\

SPDA bin &		& $18.99$ & $20.99$ & $19.43$ & $21.15$ & $18.84$ & $27.40$ & $18.93$ & $21.19$ & $20.97$ & $21.5$ & $25$min \\

P$^4$IP &		$1$	& $19.07$ & $22.67$ & $20.54$ & - 		& $19.31$ & $27.05$ & $20.07$ & - 		& - 		& -	 	& few mins \\

VST+BM3D &		& $18.46$ & $21.64$ & $20.19$ & $21.62$ & $19.43$ & $25.82$ & $19.71$ & $21.47$ & $21.14$ & $21.92$ & $0.78$s \\

VST+BM3D bin &	& $19.28$ & $22.53$ & $20.69$ & $22.07$ & $19.59$ & $\textbf{27.59}$ & $20.22$ & $21.97$ & $21.81$ & $22.72$ & $0.10$s \\

I+VST+BM3D &	& $19.74$ & $\textbf{23.04}$ & $21.07$ & $22.30$ & $\textbf{19.86}$  & $27.27$ & $20.44$ & $22.17$ & $22.08$ & $\textbf{22.85}$ & $0.82$s \\

Ours &		& $19.45$ & $22.87$ & $\textbf{21.59}$ & $\textbf{22.49}$ & $19.83$ & $26.26$ & $\textbf{21.43}$ & $\textbf{22.38}$ & $\textbf{22.11}$ & $22.82$ & $0.04$s/$1.3$s \\
\hline

NLSPCA & 		& $19.70$ & $23.16$ & $20.64$ & $22.37$ & $19.43$ & $26.88$ & $20.48$ & $21.83$ & $21.75$ & $22.68$ & $87$s \\

NLSPCA bin &	& $15.52$ & $20.85$ & $18.35$ & $19.87$ & $18.32$ & $21.27$ & $17.78$ & $20.29$ & $20.21$ & $20.98$ & $12$s \\

SPDA &			& $\textbf{24.72}$ & $24.37$ & $21.35$ & - 		& $20.17$ & $29.13$ & $21.18$ & - 		& - 	  & - 		& $6h$ \\

SPDA bin &		& $19.26$ & $21.12$ & $19.53$ & $21.66$ & $18.87$ & $28.54$ & $19.17$ & $21.43$ & $21.24$ & $21.94$ & $25$min \\

P$^4$IP &	$2$ 	& $21.04$ & $24.65$ & $21.87$ & - 		& $20.16$ & $28.93$ & $21.33$ & - & - & - & few mins \\

VST+BM3D &		& $20.79$ & $23.79$ & $21.97$ & $23.11$ & $20.49$ & $27.95$ & $22.02$ & $22.90$ & $22.65$ & $23.34$ & $0.82$s \\

  VST+BM3D bin && $19.91$ & $24.10$ & $21.43$ & $23.03$ & $20.36$ & $\textbf{29.26}$ & $21.45$ & $22.92$ & $22.84$ & $23.75$ & $0.10$s \\

I+VST+BM3D &	& $21.18$ & $24.62$ & $22.25$ & $23.40$ & $20.69$ & $28.85$ & $21.93$ & $23.30$ & $23.12$ & $23.88$ & $0.82$s \\

Ours &		& $21.38$ & $\textbf{24.77}$ & $\textbf{23.25}$ & $\textbf{23.64}$ & $\textbf{20.80}$ & $28.37$ & $\textbf{23.19}$ & $\textbf{23.66}$ & $\textbf{23.30}$ & $\textbf{23.95}$ & $0.04$s/$1.3$s \\
\hline

NLSPCA & 	 	& $20.15$ & $24.26$ & $20.97$ & $22.93$ & $20.21$ & $27.99$ & $21.07$ & $22.49$ & $22.33$ & $23.51$ & $123$s \\

NLSPCA bin &	& $15.52$ & $20.94$ & $18.27$ & $19.88$ & $18.32$ & $22.02$ & $17.72$ & $20.29$ & $20.25$ & $20.99$ & $13$s \\

SPDA &			& $\textbf{25.76}$ & $25.3$ & $21.72$ & -& $20.53$ & $\textbf{31.13}$ & $22.2$ & - & -& - & $8$h\\

SPDA bin &		& $19.42$ & $22.07$ & $19.95$ & $22.18$ & $19.26$ & $29.71$ & $20.19$ & $21.76$ & $21.69$ & $22.82$ & $31$min \\

P$^4$IP &		$4$	& $22.49$ & $26.33$ & $23.29$ & $24.66$ & $21.11$ & $30.82$ & $23.88$ & $24.10$ & $23.99$ & $25.28$ & few mins \\

VST+BM3D &		& $22.93$ & $25.49$ & $23.82$ & $24.32$ & $21.51$ & $29.41$ & $24.01$ & $24.16$ & $24.10$ & $24.47$ & $0.74$s \\

VST+BM3D bin &	& $20.43$ & $25.49$ & $22.22$ & $23.99$ & $21.13$ & $30.87$ & $22.57$ & $23.92$ & $23.84$ & $24.69$ & $0.10$s \\

I+VST+BM3D &	& $23.51$ & $26.07$ & $24.10$ & $24.52$ & $21.71$ & $30.38$ & $24.04$ & $24.53$ & $24.34$ & $24.82$ & $1.41$s \\

Ours &		& $23.18$ & $\textbf{26.59}$ & $\textbf{24.87}$ & $\textbf{24.77}$ & $\textbf{21.81}$ & $30.02$ & $\textbf{24.83}$ & $\textbf{24.86}$ & $\textbf{24.60}$ & $\textbf{25.01}$ & $0.04$s/$1.3$s \\
\hline

NLSPCA & 	 	& $14.87$ & $20.87$ & $18.21$ & $19.76$ & $18.23$ & $21.44$ & $17.67$ & $20.20$ & $20.21$ & $20.93$ & $60$s \\

SPDA &			& $\textbf{26.85}$ & $26.36$ & $22.24$ & $24.36$ & $21.05$ & $32.39$ & $22.89$ & $23.50$ & $23.37$ & $24.93$ & days \\ 

P$^4$IP &		$8$	& $23.10$ & $27.36$ & $24.49$ & $24.96$ & $21.68$ & $\textbf{32.88}$ & $24.94$ & $25.03$ & $25.06$ & $24.50$ & $167$s \\

I+VST+BM3D &	& $25.54$ & $27.95$ & $25.74$ & $25.81$ & $22.72$ & $32.35$ & $25.90$ & $25.95$ & $25.79$ & $26.06$ & $5.1$s \\ 

Ours &		& $25.73$ & $\textbf{28.42}$ & $\textbf{26.35}$ & $\textbf{26.10}$ & $\textbf{22.91}$ & $32.28$ & $\textbf{26.45}$ & $\textbf{26.23}$ & $\textbf{26.11}$ & $\textbf{26.26}$ & $0.04$s/$1.3$s \\
\hline 

NLSPCA & 		&  $14.78$ & $18.83$ & $17.98$ & $19.39$ & $18.03$ & $21.41$ & $17.06$ & $19.92$ & $19.98$ & $20.60$ & $92$s \\

SPDA &			& $27.10$ & $27.06$ & $22.47$ & $25.02$	& $21.22$ & $35.08$ & $23.61$ & $24.55$ & $24.06$ & $25.88$ & days \\ 

P$^4$IP &	 $30$	& $27.02$ & $29.85$ & $27.28$ & $26.52$ & $23.07$ & $36.03$ & $27.33$ & $26.98$ & $27.22$ & $27.01$ & $149$s\\

I+VST+BM3D &	& $\textbf{29.09}$ & $31.35$ & $28.55$ & $28.37$ & $25.08$ & $36.03$ & $29.08$ & $28.79$ & $28.80$ & $28.62$ & $4.5$s \\

Ours &		& $28.94$ & $\textbf{31.67}$ & $\textbf{29.21}$ & $\textbf{28.74}$ & $\textbf{25.42}$ & $\textbf{36.20}$ & $\textbf{29.77}$ & $\textbf{29.06}$ & $\textbf{29.13}$ & $\textbf{28.71}$ & $0.04$s/$1.3$s \\

\hline\hline 
 
  \end{tabular}  
  \vspace{2mm}
\caption{\small \textbf{Poisson denoising performance on standard images.} Numeric values represent PSNR in dB averaged over five noise realizations. Values for prior art algorithms for peak values of $1-4$ were taken from \cite{Azzari16Variance}. For the rest of the peak values we ran the code published by the authors; in the absence of optimal parameter settings, we used those for peak$=4$. Timing values presented are averages for images of size $256\times256$, for our method we present the run-time on Titan-X GPU/Intel E5-2630 $2.20$GHz CPU.}
\label{tab_10_image_psnr}	
\end{table*}

Next, we tested our method on the $68$ test images from the Berkeley dataset \cite{MartinFTM01}. Note that we did not fine-tune our network to fit this dataset but rather used it after it had been trained on PASCAL. Results are summarized in Table \ref{tab_bsds_poisson}. 

Finally, we evaluated our method on the standard 10 image set as presented in Table \ref{tab_10_image_psnr}.

\subsubsection{Darmstadt noise dataset}
\label{sec_darmstadt}
\ed{To test our network on real noisy data, 
we have evaluated its performance on the dataset presented in \cite{plotz2017benchmarking} using their official benchmark\footnote{https://noise.visinf.tu-darmstadt.de} for sRGB images. For the evaluation, we denoised each of the RGB color channels separately (to make a fair comparison to BM3D, MLP, DnCNN and TNRD that also have been applied on each channel separately when tested on this dataset). The networks used were the same ones used for the experiments in Section \ref{sec_agnostic_gaussian}, that were trained to denoise images with white additive Gaussian noise as explained in Section \ref{sec_implementation}. Since an estimate of the noise standard deviation is provided along with each test image we used the network trained for noise with the closest larger standard deviation.
The results are presented in Table \ref{tab_darmstadt}, where it can be seen that our methods compares favorably to MLP, DnCNN, and TNRD (the full table can be found on the official benchmark website).}

\begin{table}[t]
\centering
\small
\begin{tabular}{ c c c c c c }
\hline
\hline
    & BM3D & MLP & DnCNN & TNRD & Ours \\ \hline
    PSNR & $34.51$ & $34.23$ & $32.43$ & $33.65$ & $35.08$ \\
    SSIM & $0.85$ & $0.83$ & $0.79$ & $0.83$ & $0.87$ \\
    \hline \hline
  \end{tabular}  
  \vspace{2mm}
\caption{\small \textbf{\ed{Darmstadt noise dataset.}} \ed{Average PSNR and SSIM.}}
\label{tab_darmstadt}	
\end{table}

\subsubsection{Real low-light image}
\label{sec_real_pics}
To further test our network on real low-light images, we captured an image with a relatively short exposure ($1/13 sec$ and ISO 6400) introducing shot noise as well as all other camera noise sources. To achieve a noise-free reference image, we took another image of the same scene using a very long exposure ($15 sec$ and ISO 200). Images were captured using a Canon EOS 600D. The lighting and exposure time settings resulted in peak values of approximately $30$. 

\begin{figure*}[tb]
	\small
	\centering
    \begin{tabular}{c@{\hskip 0.01\textwidth}c@{\hskip 0.01\textwidth}c@{\hskip 0.01\textwidth}c@{\hskip 0.01\textwidth}c@{\hskip 0.01\textwidth}c}
		\includegraphics[width = 0.19\textwidth]{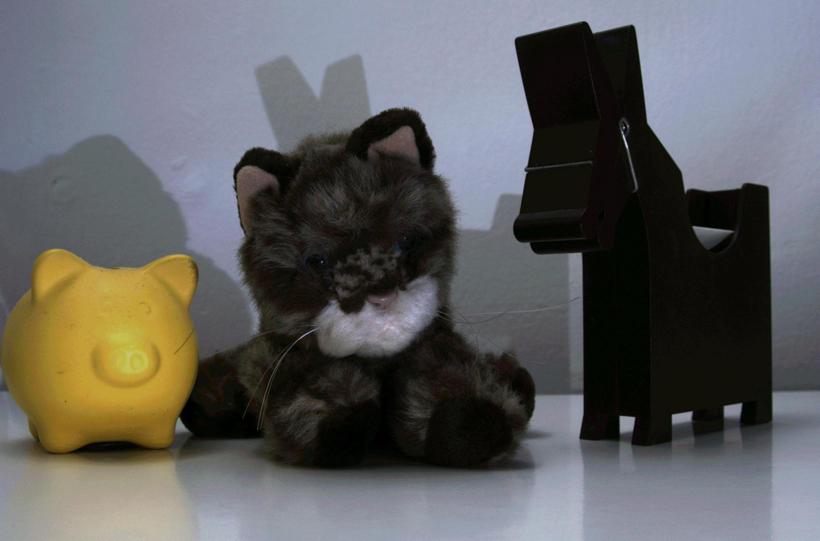} &
		\includegraphics[width = 0.19\textwidth]{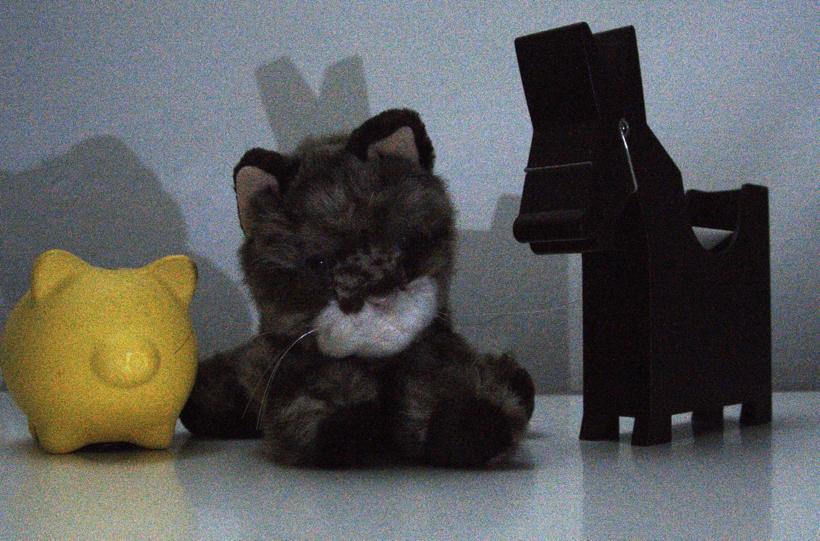} &
        \includegraphics[width = 0.19\textwidth]{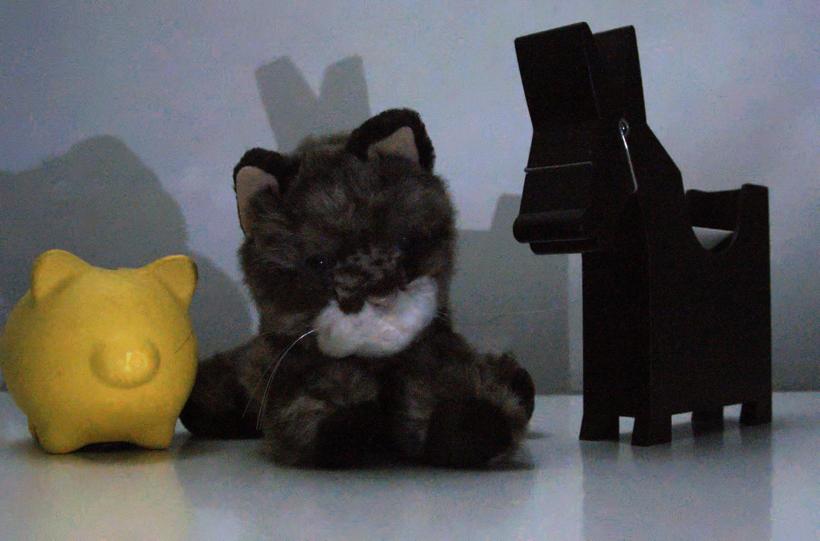} &
		\includegraphics[width = 0.19\textwidth]{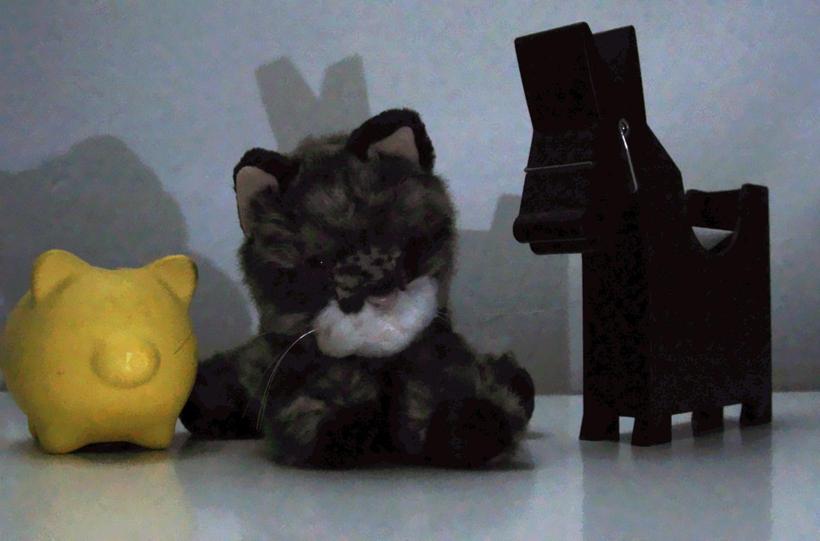} &
        \includegraphics[width = 0.19\textwidth]{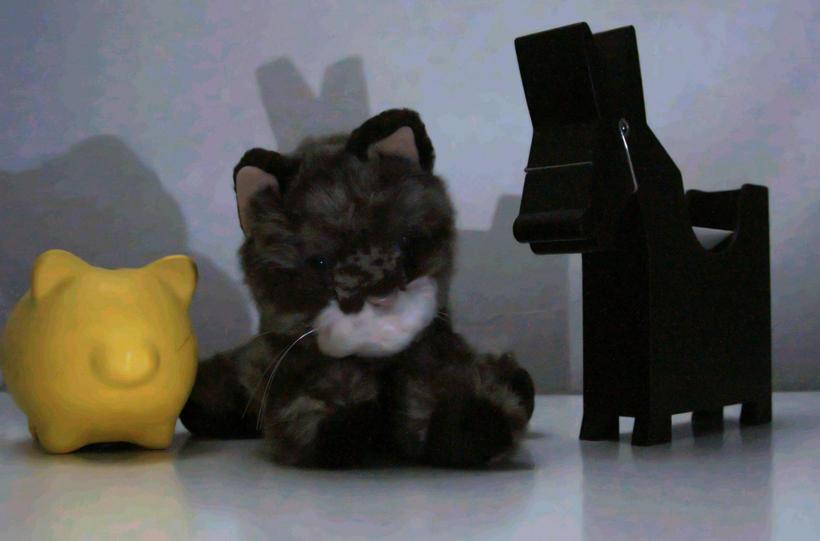} \\
        
	\end{tabular}  
   	\begin{tabular}{c@{\hskip 0.01\textwidth}c@{\hskip 0.01\textwidth}c@{\hskip 0.01\textwidth}c@{\hskip 0.01\textwidth}c@{\hskip 0.01\textwidth}c@{\hskip 0.01\textwidth}c@{\hskip 0.01\textwidth}c@{\hskip 0.01\textwidth}c@{\hskip 0.01\textwidth}c}
		\includegraphics[width = 0.09\textwidth]{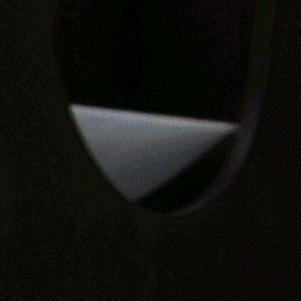} &
        \includegraphics[width = 0.09\textwidth]{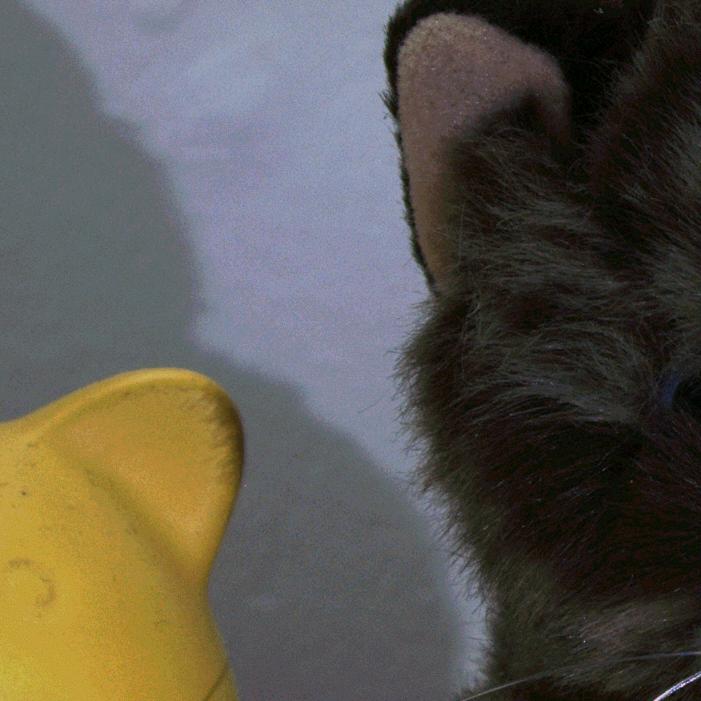} &
		\includegraphics[width = 0.09\textwidth]{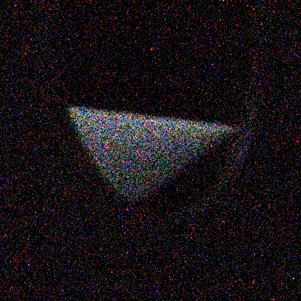} &
        \includegraphics[width = 0.09\textwidth]{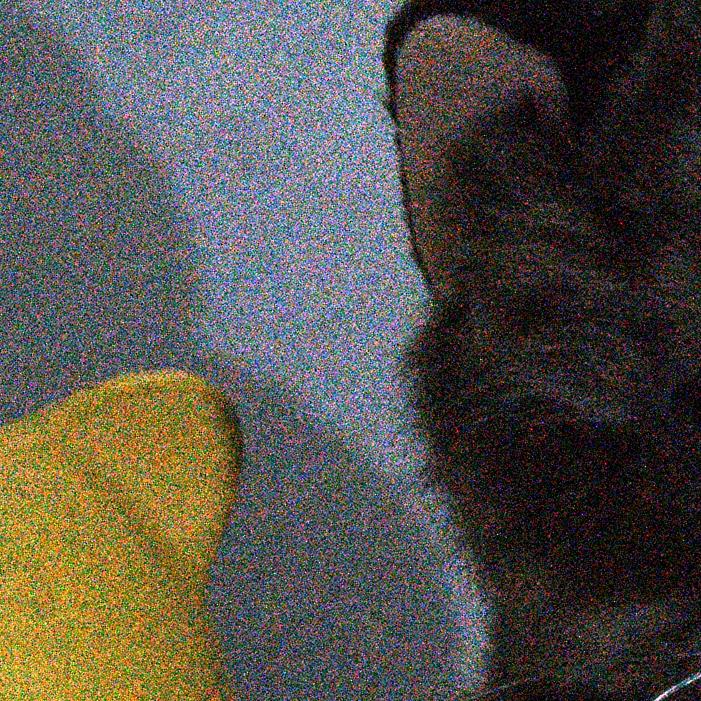} &
   		\includegraphics[width = 0.09\textwidth]{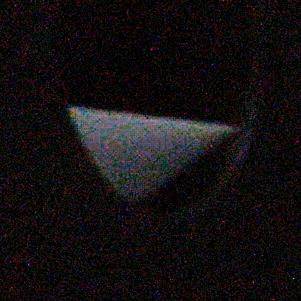} &
        \includegraphics[width = 0.09\textwidth]{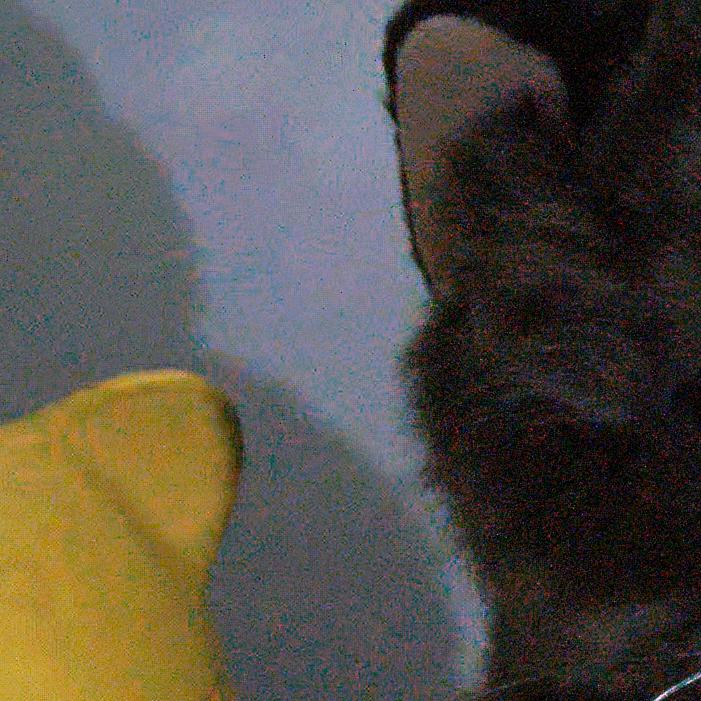} &
   		\includegraphics[width = 0.09\textwidth]{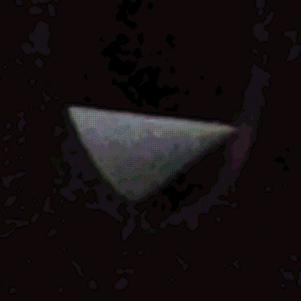} &
        \includegraphics[width = 0.09\textwidth]{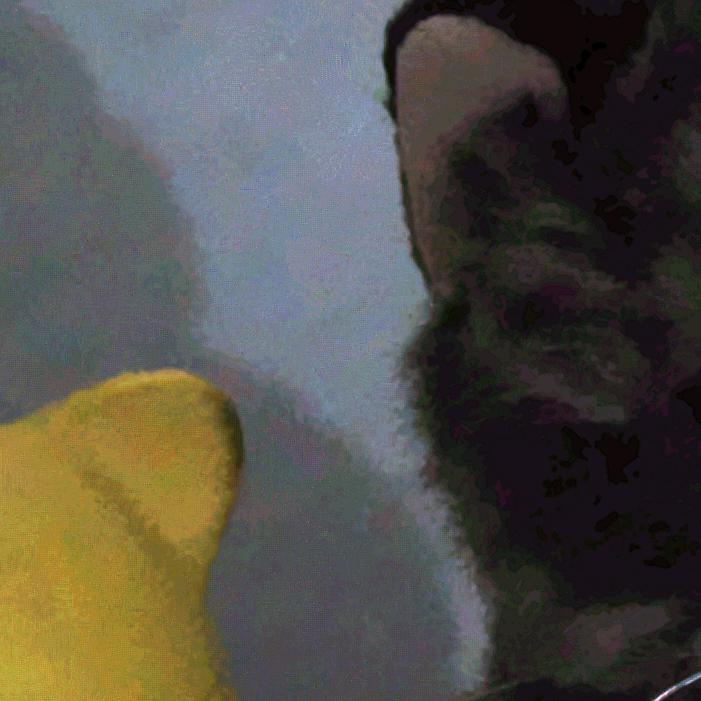} &
        \includegraphics[width = 0.09\textwidth]{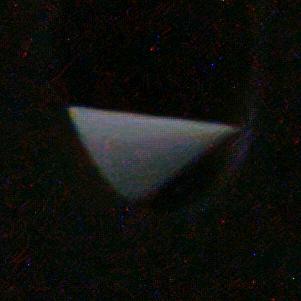} &
        \includegraphics[width = 0.09\textwidth]{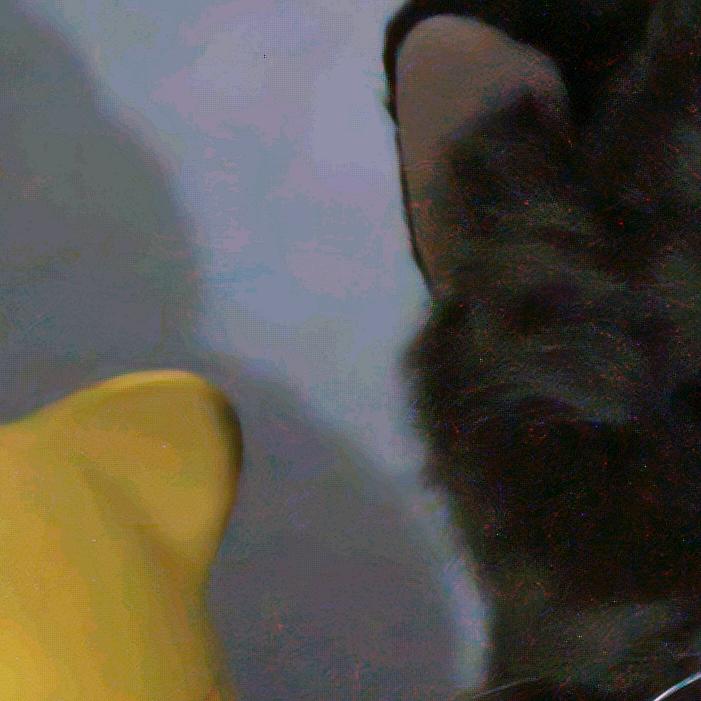}\\  
        
        \multicolumn{2}{c}{Clean reference}&
        \multicolumn{2}{c}{Noisy input}&
        \multicolumn{2}{c}{I+VST+BM3D}&
        \multicolumn{2}{c}{BM3D CFA} &
        \multicolumn{2}{c}{Ours} \\
        
        \multicolumn{2}{c}{} &
        \multicolumn{2}{c}{$15.16$ dB} &
        \multicolumn{2}{c}{$22.52$ dB} &
        \multicolumn{2}{c}{$24.74$ dB} &
        \multicolumn{2}{c}{$25.80$ dB} \\        
	\end{tabular}      
    \smallskip 
	\caption{\textbf{Denoising a real low-light image. } Denoising comparison of a low-light image with a peak value of approximately 30, intensities were scaled for display purposes. High resolution images are available in the supplementary material.  
    }
\label{fig_real_exp}
\end{figure*}

Figure \ref{fig_real_exp} shows the reconstruction results using a network trained with peak value of $30$. Although trained solely on simulated noise, our network achieves  visually appealing reconstruction results.
A quantitative evaluation shows an improvement of over $1$dB compared to BM3D CFA and I+VST+BM3D. Run time was $32$ minutes for I+VST+BM3D, $7$ minutes for BM3D, and $4$ minutes for our network running on an Intel E5-2630 $2.20$GHz CPU for an $18$ mega-pixel image. Our networks takes only $8$ seconds on a Titan X GPU. 

For I+VST+BM3D and our method, each color channel was denoised separately. Cross-color BM3D filtering (BM3D CFA) \cite{danielyan2009cross} was applied on all channels together with its optimal parameters. The default MATLAB demosaicing and a Gray world white blanching algorithms were used. 

\subsection{Visualizing the denoising process}
\label{sec_visualizing}
\begin{figure*}[htb]
	\centering	
    \small
        \begin{tabular}{ c@{\hskip 0.005\textwidth}c@{\hskip 0.005\textwidth}c@{\hskip 0.005\textwidth}c@{\hskip 0.005\textwidth}c}
		\includegraphics[width = 0.19\textwidth]{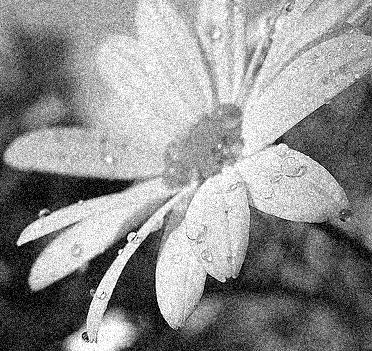} &
		\includegraphics[width = 0.19\textwidth]{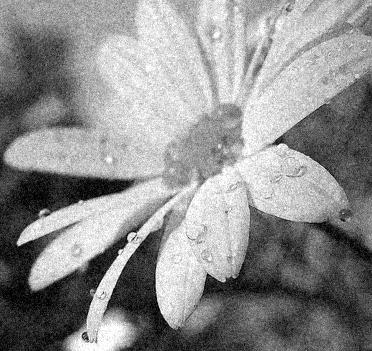} &
        \includegraphics[width = 0.19\textwidth]{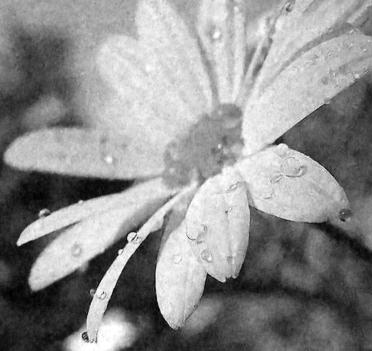} &
        \includegraphics[width = 0.19\textwidth]{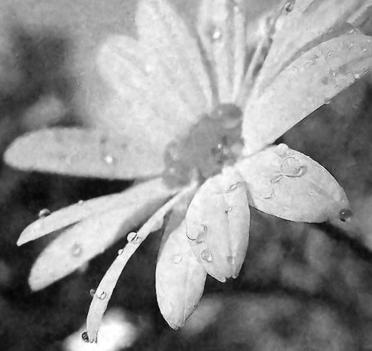} &
		\includegraphics[width = 0.19\textwidth]{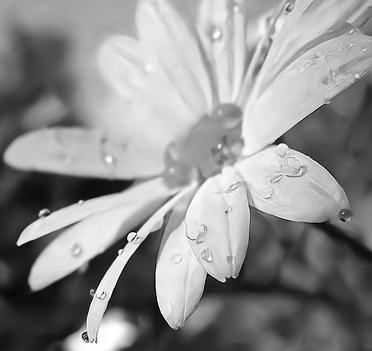} \\
        
        Noisy input \ed{$20.16$ dB}& \ed{$24.52$ dB} & \ed{$29.81$ dB} & \ed{$31.08$ dB} & Output \ed{$32.84$ dB} \\
        
		\includegraphics[width = 0.19\textwidth]{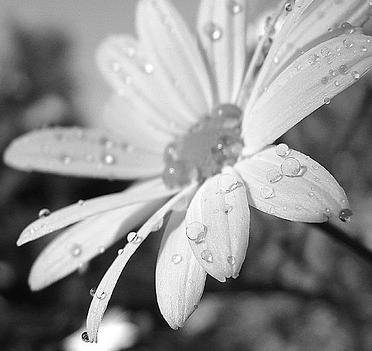} &
		\includegraphics[width = 0.19\textwidth]{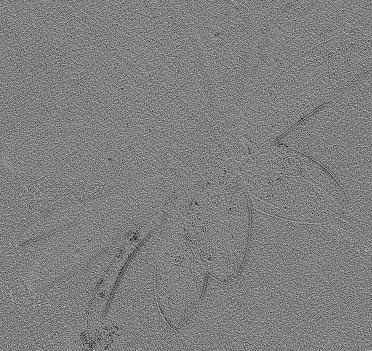} &
        \includegraphics[width = 0.19\textwidth]{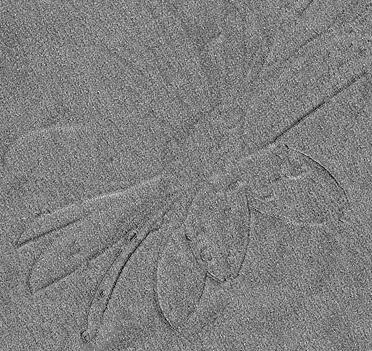} &
        \includegraphics[width = 0.19\textwidth]{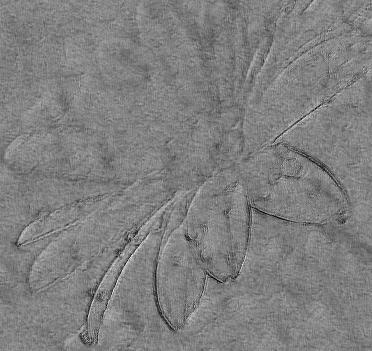} &
		\includegraphics[width = 0.19\textwidth]{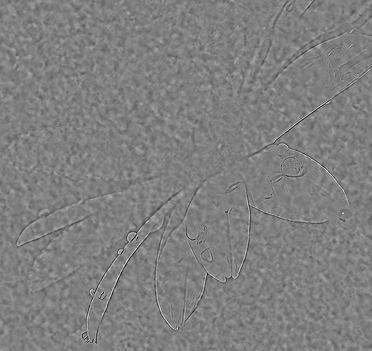} \\  
        Ground truth & Layer 5 & Layer 10 & Layer 15 & Layer 20 \\
        
		\includegraphics[width = 0.19\textwidth]{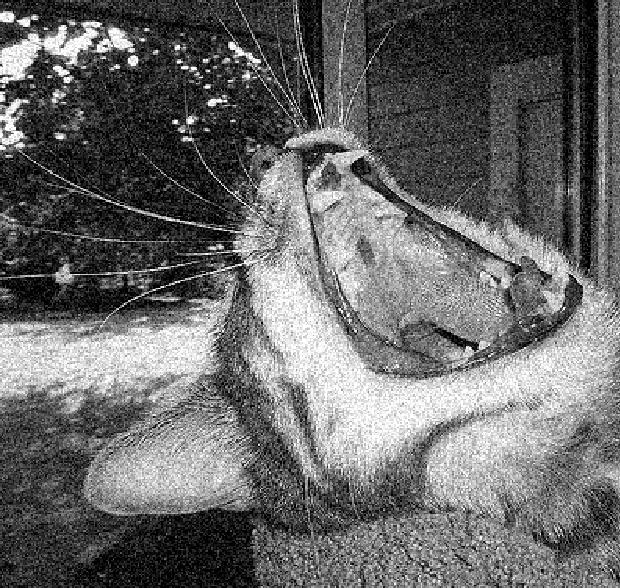} &
		\includegraphics[width = 0.19\textwidth]{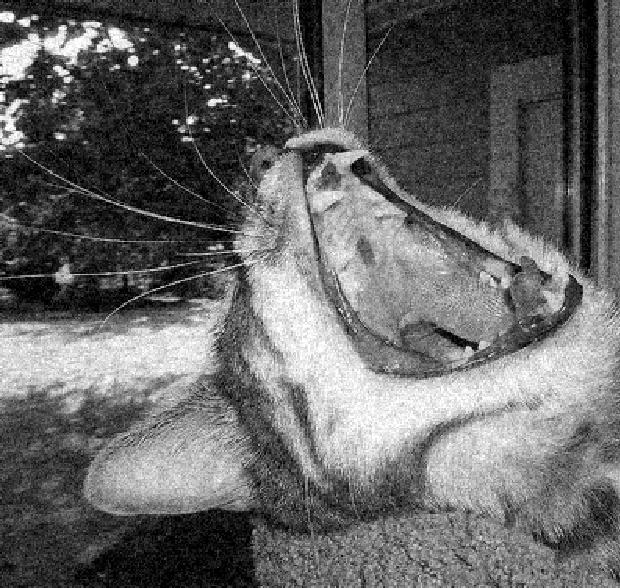} &
        \includegraphics[width = 0.19\textwidth]{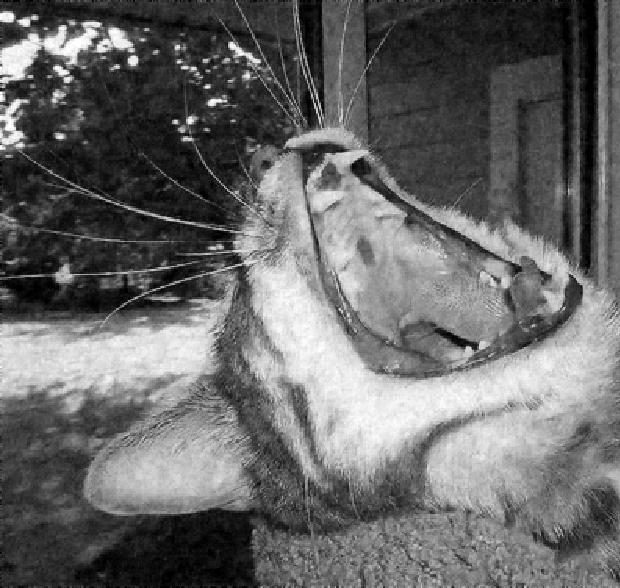} &
        \includegraphics[width = 0.19\textwidth]{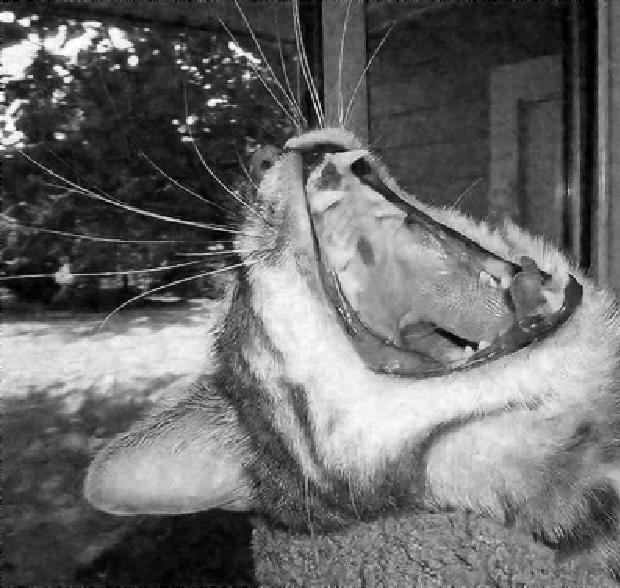} &
		\includegraphics[width = 0.19\textwidth]{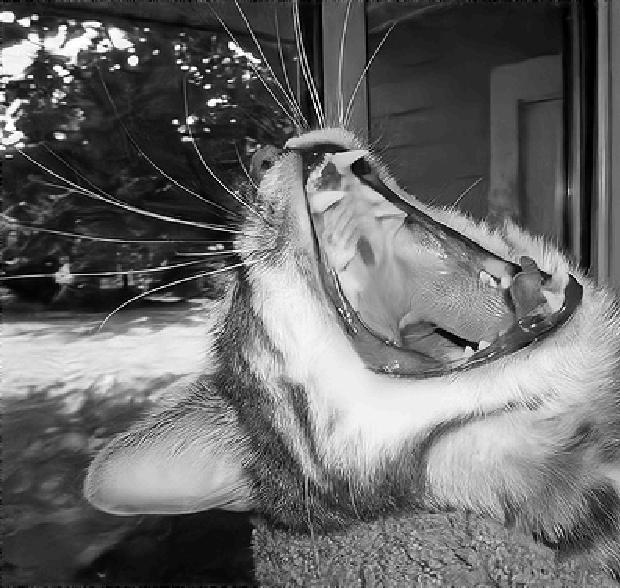} \\
        Noisy input \ed{$20.19$ dB}& \ed{$23.33$ dB} & \ed{$24.29$ dB} & \ed{$24.52$ dB} & Output \ed{$26.05$ dB} \\
        
		\includegraphics[width = 0.19\textwidth]{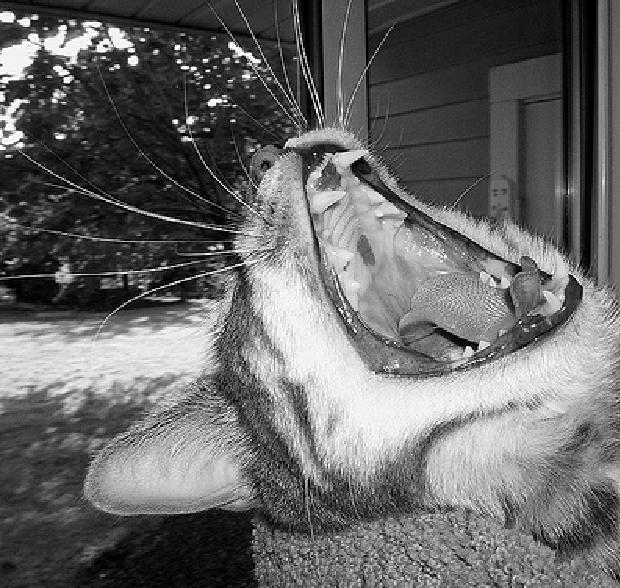} &
		\includegraphics[width = 0.19\textwidth]{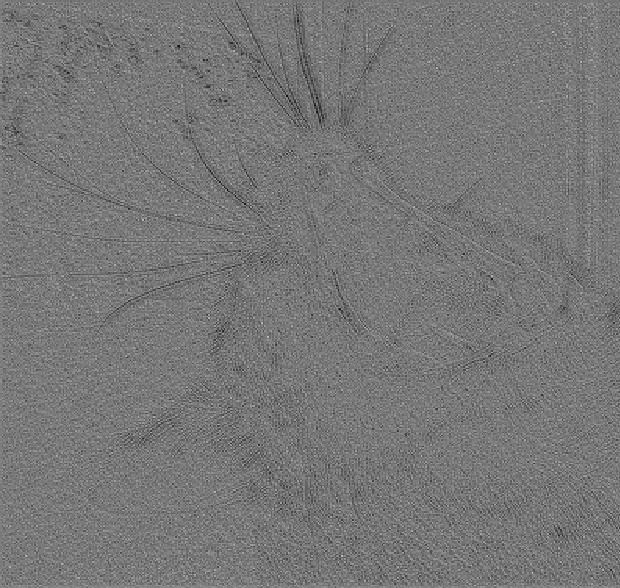} &
        \includegraphics[width = 0.19\textwidth]{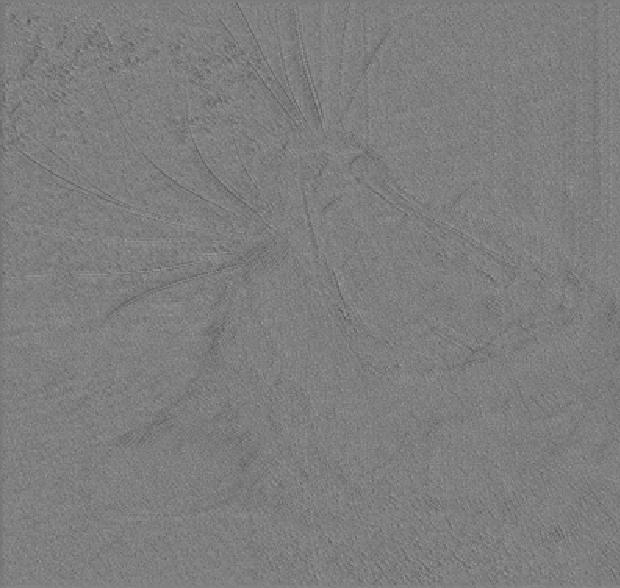} &
        \includegraphics[width = 0.19\textwidth]{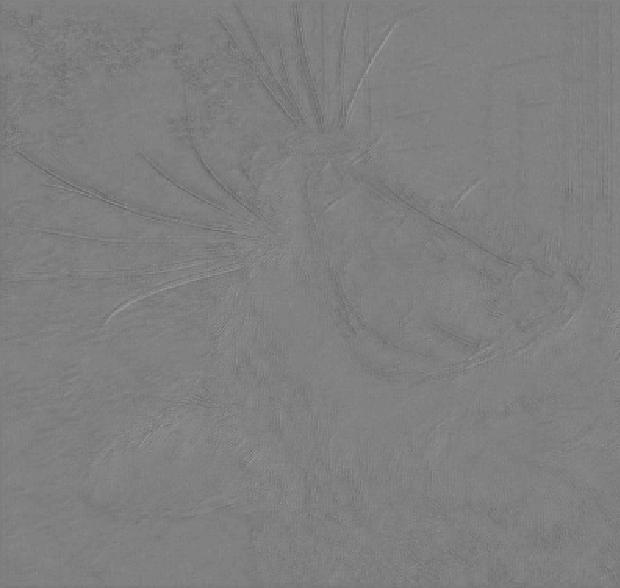} &
		\includegraphics[width = 0.19\textwidth]{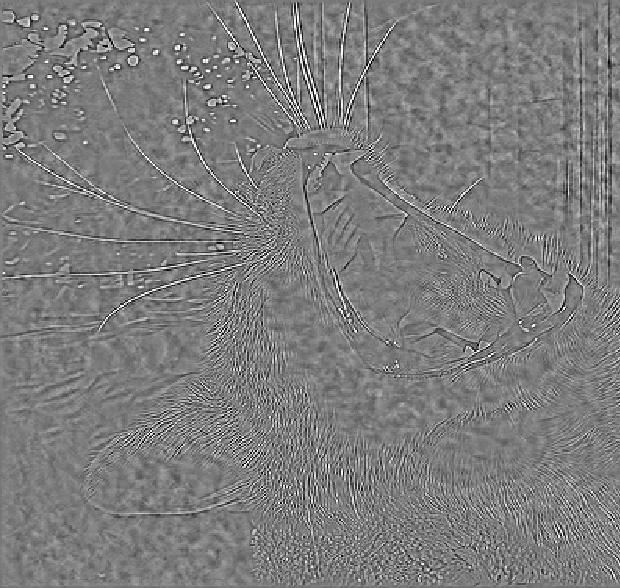} \\  
        Ground truth & Layer 5 & Layer 10 & Layer 15 & Layer 20 \\
        
	\end{tabular}   \\
	\caption{\small\textbf{Gradual denoising process.} The top row presents the noisy image (left) and the intermediate result obtained by removing the noise estimated up to the respective layer depth. 
The second row presents the ground truth image (left) and the noise estimates produced by individual layers; noise estimation images have been scaled for display purposes. Zoom-in for a better view of the fine details and noise. \ed{Images have been contaminated by Gaussian noise with $\sigma=25$. PSNR values are presented below each image.}}
    \label{fig_denoise_flow}
\end{figure*}

\begin{figure}[tb]
\begin{centering}
    \small
	\begin{tabular}{c@{\hskip 0.005\textwidth}c@{\hskip 0.005\textwidth}c}
    
   		\includegraphics[height = 0.21\textwidth]{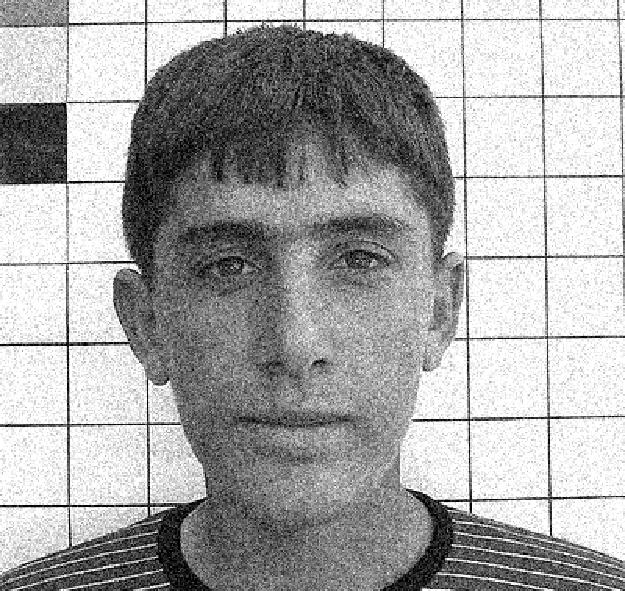} &
		\includegraphics[height = 0.21\textwidth]{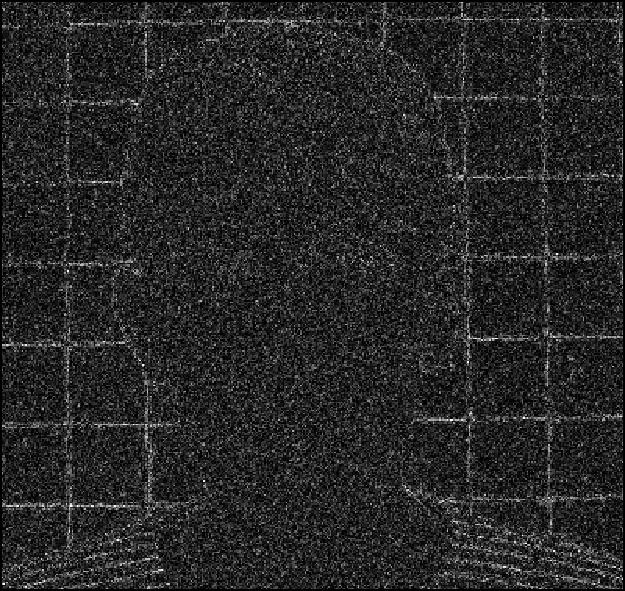} &\\
        Noisy image & Error after 5 layers  \\
		\includegraphics[height = 0.21\textwidth]{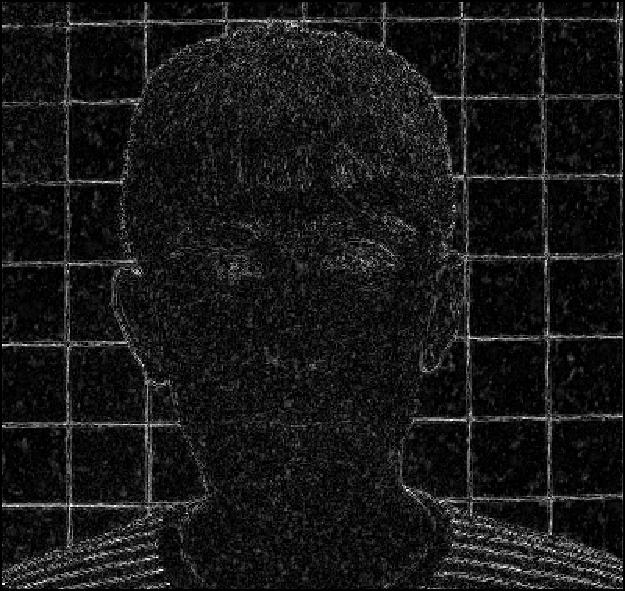} &
		\includegraphics[height = 0.21\textwidth]{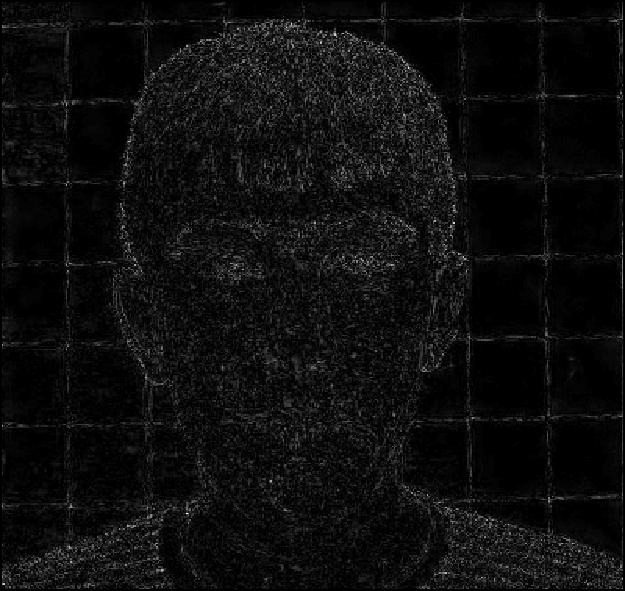} &\\
        Error after 10 layers & Error after 20 layers \\
        \includegraphics[width = 0.21\textwidth]{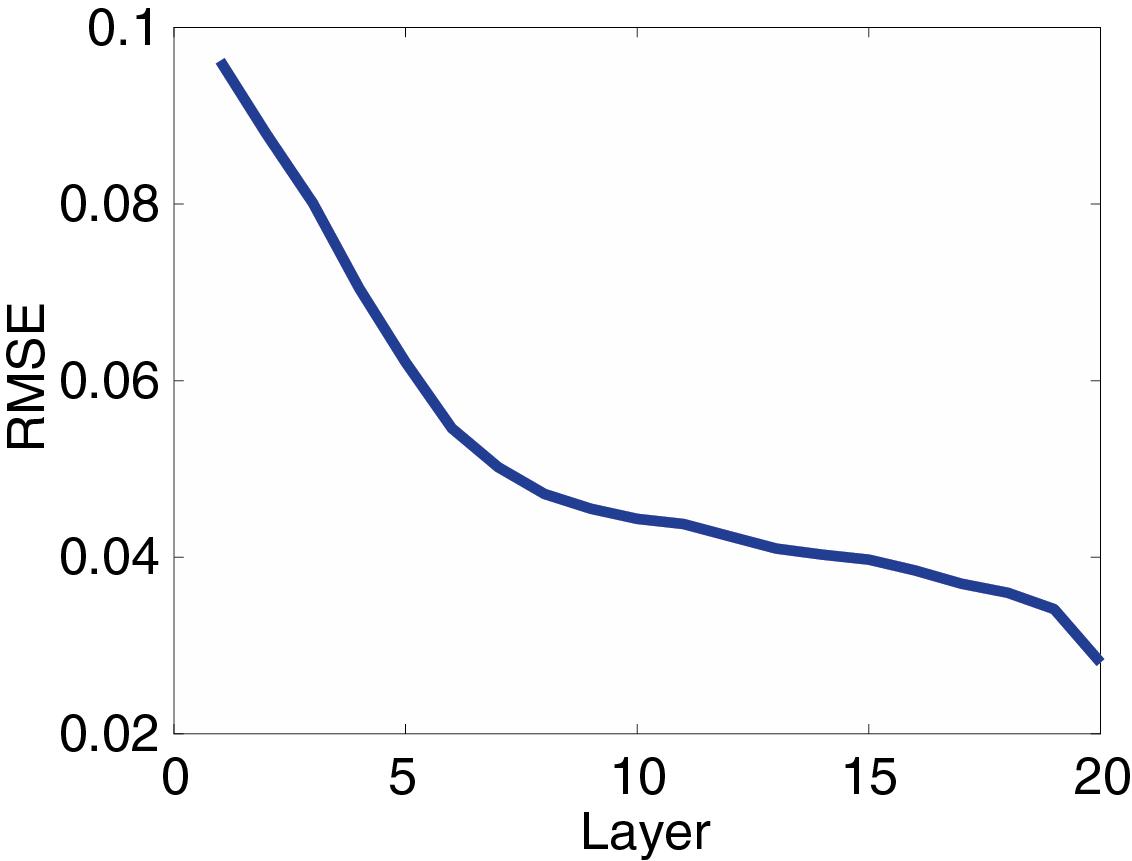} &     
		\includegraphics[height = 0.21\textwidth]{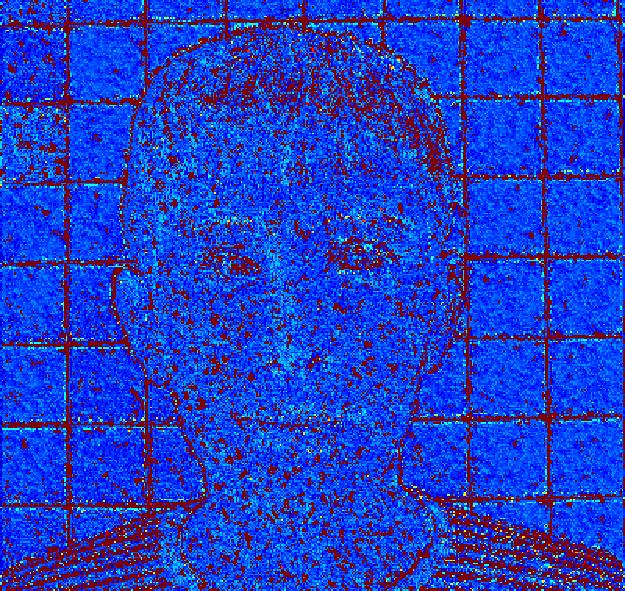} &
        \hspace{-1mm} \includegraphics[width = 0.0255\textwidth]{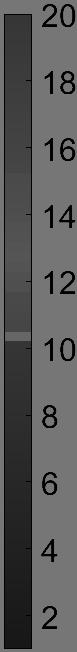} \\ 
        \parbox[b][4em][s]{0.13\textwidth}{RMSE per layer} & \parbox[b][4em][s]{0.16\textwidth}{Most significant layer}\\
        \vspace{-13mm}
                
	\end{tabular}   \\
    \end{centering}
	\caption{\small \textbf{Gradual error reduction.} 
    Top two rows: noisy image (Gaussian noise, $\sigma=25$) and the difference images evaluated after accumulating noise estimations from the first $5, 10,$ and $20$ layers. Bottom left: RMSE after each layer.
Bottom right: Each pixel is colored according to the depth of the layer in which its value changed the most.  
%
%
}
    \label{fig_layer_selection}
\end{figure}
One advantage of having intermediate noise estimates at intermediate layers is that it allows to examine the process of an otherwise black-box algorithm. In Fig.~\ref{fig_denoise_flow} we show an example of how an input image contaminated by Gaussian noise ($\sigma=25$) is being denoised as it flows through the network. It is evident that each layer of the network contributes differently to the noise removal process. Shallower layers (e.g. layer $5$ in the figure) seem to handle local noise statistics, while  deeper layers (e.g. layers $15, 20$) recover edges and enhance textures, which might have been degraded by the first layers. A possible explanation for this may reside in the receptive field sizes. Deeper layers correspond to larger receptive fields and therefore can better recover large patterns such as edges, contours, and textures, which might be indistinguishable from noise when viewed by smaller receptive fields of shallower layers.

In Fig.~\ref{fig_layer_selection} we inspect how the error between the ground truth and denoised image changes at different stages of the denoising process. We display the error after aggregating the noise estimation of the first $5, 10, $ or $20$ layers. It is evident that after about $10$ layers most of the smooth image areas are properly denoised. However, there are still non-negligible errors around edges and in textured areas. Combining the entire $20$ layers of the network, it is apparent that most of these errors are reduced significantly. 

This phenomenon is also evident when visualizing which layer was the most dominant in the denoising process of each pixel (Fig.~\ref{fig_layer_selection}, bottom right).
%
%
To further investigate the denoising process, we plot the RMSE after each layer. Surprisingly, even though it has not been explicitly enforced at training, the error monotonically decreases with the layer depth (bottom left). This non-trivial behavior is consistently produced by the network on the vast majority of test images. 
These results align with the ones depicted in Fig.~\ref{fig_denoise_flow}.

\begin{figure*}[tb]
\centering
\footnotesize
\begin{tabular}{ r@{\hspace{0.005\textwidth}}c@{\hspace{0.005\textwidth}}c@{\hspace{0.005\textwidth}}c@{\hspace{0.005\textwidth}}c@{\hspace{0.005\textwidth}}c}

    \vspace{0.5mm}Ground Truth &
    \raisebox{-.5\height}{\includegraphics[height=2.5cm]{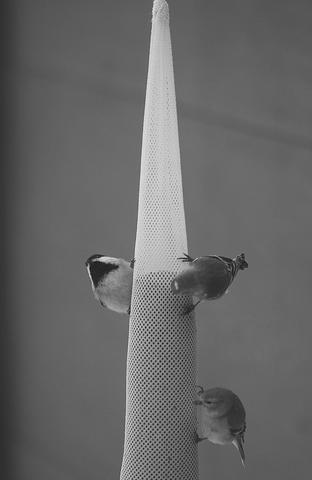}} &
    \raisebox{-.5\height}{\includegraphics[height=2.5cm]{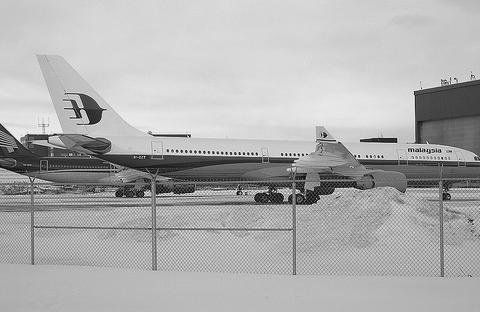}} &
    \raisebox{-.5\height}{\includegraphics[height=2.5cm]{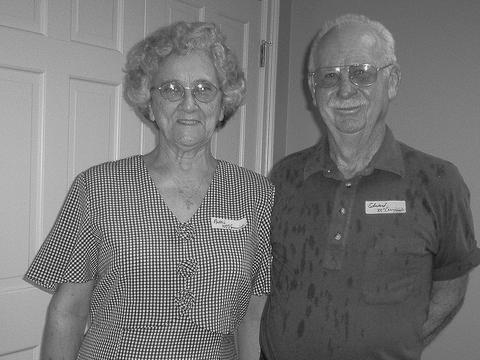}} &
    \raisebox{-.5\height}{\includegraphics[height=2.5cm]{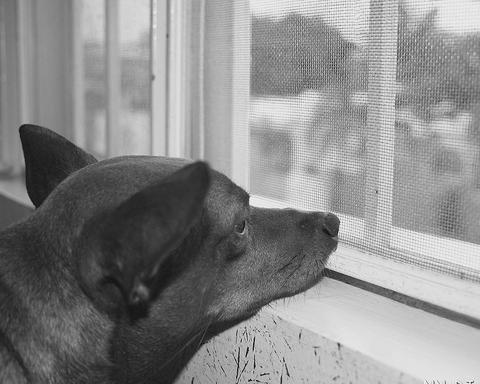}} &
    \raisebox{-.5\height}{\includegraphics[height=2.5cm]{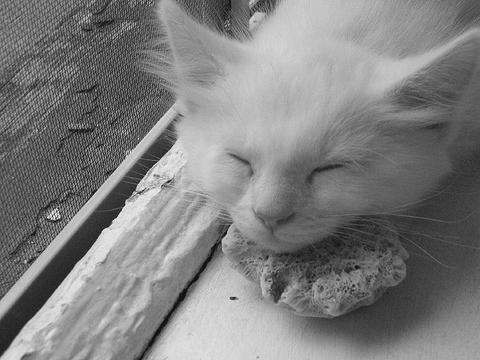}} \\
    &&&&&\\

    \vspace{0.5mm}BM3D &
    \raisebox{-.5\height}{\includegraphics[height=2.5cm]{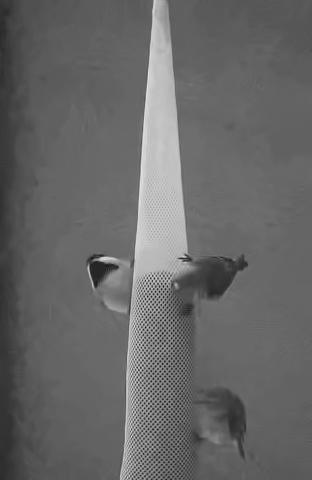}} &
    \raisebox{-.5\height}{\includegraphics[height=2.5cm]{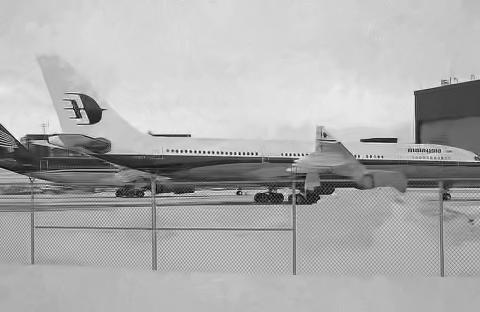}} &
    \raisebox{-.5\height}{\includegraphics[height=2.5cm]{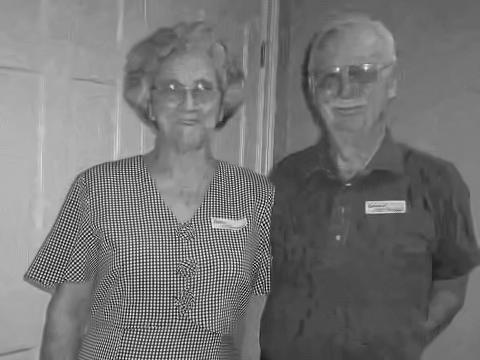}} &
    \raisebox{-.5\height}{\includegraphics[height=2.5cm]{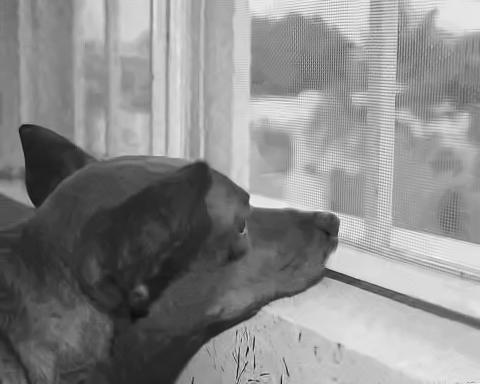}} &
    \raisebox{-.5\height}{\includegraphics[height=2.5cm]{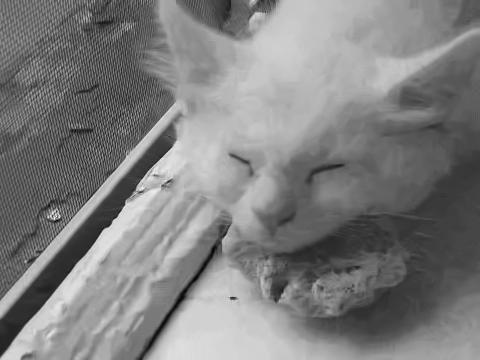}} \\
    & $34.08$ dB & $29.18$ dB & $28.80$ dB & $29.84$ dB & $29.10$ dB \\
    \vspace{0.5mm}Ours &
    \raisebox{-.5\height}{\includegraphics[height=2.5cm]{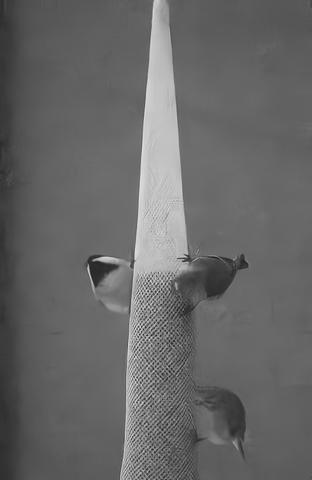}} &
    \raisebox{-.5\height}{\includegraphics[height=2.5cm]{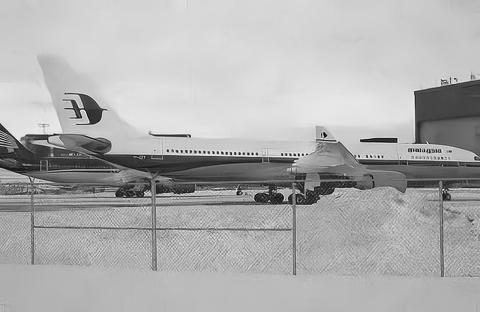}} &
    \raisebox{-.5\height}{\includegraphics[height=2.5cm]{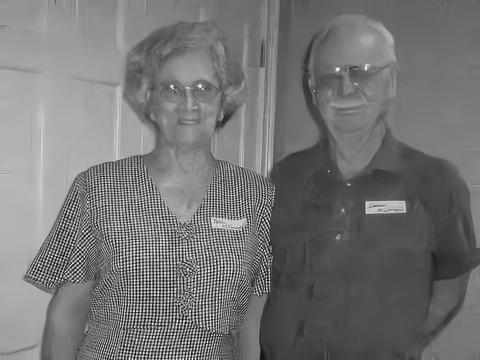}} &
    \raisebox{-.5\height}{\includegraphics[height=2.5cm]{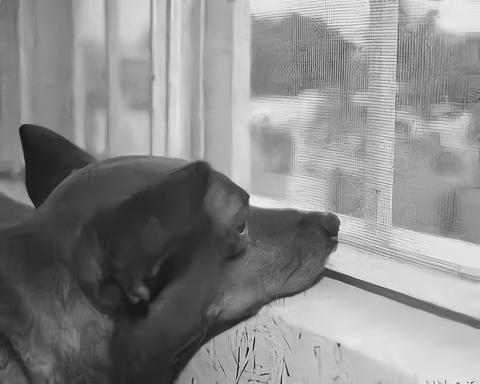}} &
    \raisebox{-.5\height}{\includegraphics[height=2.5cm]{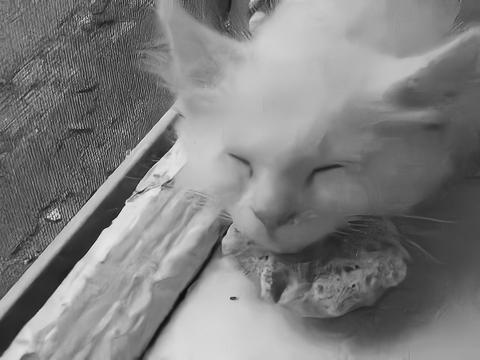}} \\ 
    & $31.28$ dB& $27.83$ dB& $27.49$ dB& $29.23$ dB& $ 28.54$ dB \\
\end{tabular}
\caption{\ed{\textbf{Failure examples.} Examples of failures of our method compared to BM3D, we selected the five worst images from the PASCAL test set (for Gaussian noise $\sigma=25$). Interestingly all of these imaged have repeated structures, such as nets or fences.}}
\label{figure:failures}
\end{figure*}

\ed{Fig.~\ref{figure:failures} presets failure cases of our class-agnostic method. It shows the five images with the worst performance compared to BM3D in the case of Gaussian noise with $\sigma=25$. Interestingly, in all five cases, big portions of the image contain repeating patterns at a high spatial frequency, e.g., nets and fences. Clearly, in such cases collaborative filtering techniques such as BM3D are expected to perform particularly well.}

\subsection{Class-aware denoising}
\label{sec_exp_aware}
As described in Section \ref{sec_method_aware}, our class-aware denoiser comprises a classifier and a set of denoisers fine-tuned to the image classes produced by the classifier.
Now we \ed{evaluate} the boost in performance gained by fine-tunning a denoiser on a set of images belonging to a particular class.

For this experiment, we collected images from ImageNet \cite{ImageNet15} belonging to the following five classes: \emph{face}, \emph{pet}, \emph{flower}, \emph{living room}, and \emph{street}. The $1500$ images per class were split into train ($60\%$), validation ($20\%$) and test ($20\%$) sets. We then trained a separate class-specific denoiser for each of the classes, initializing all weight to the values learned by our class-agnostic model and fine-tunning them to the specific class for $50K$ iterations. In addition, we trained a classifier to classify the noisy images as described in Section \ref{sec_method_aware}. 

We trained the classifier and the denoisers for two separate scenarios: (i) additive Gaussian noise with $\sigma=25$; and (ii) Poisson noise with peak value $8$.
We then compared on the test sets the performance of (a) our class specific denoisers using oracle class labels; (b) a full pipeline system that uses the learned classifier to determine the class of each image and applies the denoiser that belongs to that class; (c) our class-agnostic model trained on PASCAL; and (d) previous denoising methods using their available code and/or pretrained models. 
\ed{To evaluate the ability of other class-agnostic learning based methods to be transformed into class-aware ones as well as to have a fair comparison, we transformed IRCNN from being class-agnostic into class-aware using the exact same framework as ours (i.e., having specialized denoisers for each class and a classifier to select the image type). To this end we used our reimplementation of IRCNN and trained it using exactly the same regime as the one used to train our networks.}

\begin{figure}[bt]
    \begin{tabular}{c}
        \includegraphics[width = 0.39\textwidth]{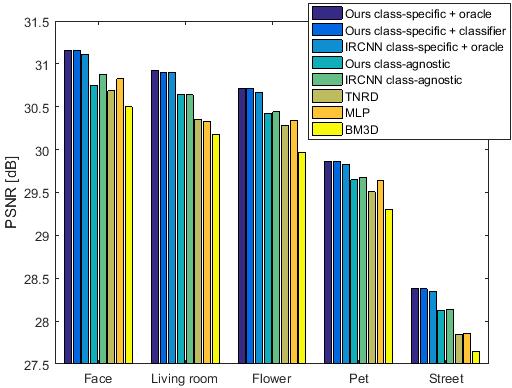} \\
        \includegraphics[width = 0.39\textwidth]{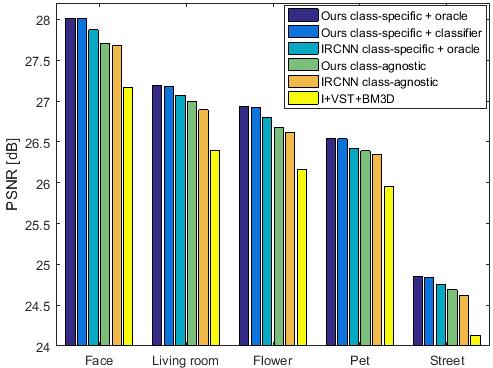} \\
	\end{tabular}   
	\caption{\small \ed{\textbf{Class-aware denoising on ImageNet.} Average PSNR values on images belonging to five semantic classes for Gaussian noise with $\sigma=25$ (top) and Poisson noise with peak $=8$ (bottom).}
    }
\label{fig_class_bar}
\end{figure}

Figure \ref{fig_class_bar} presents PSNR values for Gaussian and Poisson noise. It is evident that the class-specific models boost performance over their class-agnostic counterpart as well as other algorithms; \ed{this is true both for our method and for IRCNN}. Note also that using the full pipeline that includes a classifier exhibits similar behavior to the oracle. \ed{It is also evident that in almost all cases our method outperforms both the class-agnostic and the class-aware versions of IRCNN.} 
Table \ref{tab_classifier} presents the classifiers performance on the noisy test sets.
\begin{table}[b]
\centering
\small
  \begin{tabular}{ l@{\hskip 0.01\textwidth}|c@{\hskip 0.01\textwidth}c@{\hskip 0.01\textwidth}c@{\hskip 0.01\textwidth}c@{\hskip 0.01\textwidth}c@{\hskip 0.01\textwidth}c@{\hskip 0.01\textwidth}c@{\hskip 0.01\textwidth}c  }
    \hline \hline
    				& Face 		& Flower 	& Livingroom 	& Pet 		& Street 	\\ \hline
		Gaussian ($\sigma=25$)  	& $0.95$	& $0.97$ 	& $0.91$ 	 	& $0.92$ 	& $0.98$ 	\\
    	Poisson (peak$=8$)   	& $0.93$	& $0.98$ 	& $0.92$ 	 	& $0.80$ 	& $0.95$ 	\\
        
     \hline\hline
  \end{tabular}  
  \vspace{2mm}
\caption{\small \textbf{Correct classification rate of noisy images.} Performance of the classifiers trained for Gaussian and Poisson noise.
} 
\label{tab_classifier}	
\end{table}

\medskip 

\noindent\textbf{\em Cross class denoising.}
To further demonstrate the effect of refining a denoiser to a particular class, we tested each class-specific denoiser on images from other classes. The outcome of mismatching the image and denoiser classes is evident both qualitatively and quantitatively. 

\begin{figure}[]
	\centering
    \small
    \setlength{\tabcolsep}{0.2em}
	\begin{tabular}{ c@{\hskip 0.005\textwidth}c@{\hskip 0.005\textwidth}c }
    
	    {Noisy image} & {Correct denoiser} & {Wrong denoiser} \\ 
               
        &   \small{\textit{face} ($32.67$ dB)}& \small{\textit{street} ($32.09$ dB)} \\
        \includegraphics[width = 0.15\textwidth]{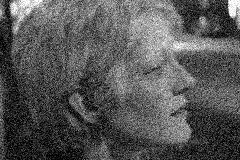} &
        \includegraphics[width = 0.15\textwidth]{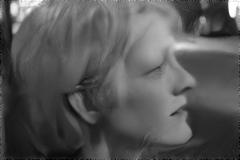} &
        \includegraphics[width = 0.15\textwidth]{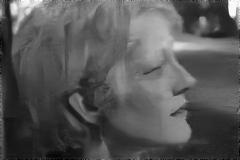} \\

		&   \small{\textit{pet} ($28.51$ dB)} & \small{\textit{living room} ($28.2$ dB)} \\        
        \includegraphics[width = 0.15\textwidth]{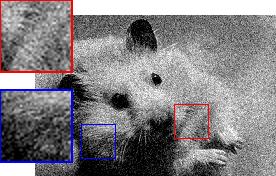} &
        \includegraphics[width = 0.15\textwidth]{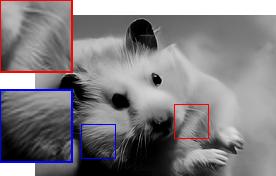} &
        \includegraphics[width = 0.15\textwidth]{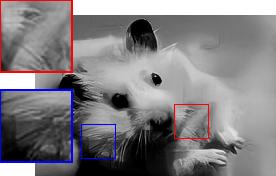}\\  

        & \small{\textit{street} ($32.63$ dB)} & \small{\textit{face} ($32.36$ dB)} \\          
        \includegraphics[width = 0.15\textwidth]{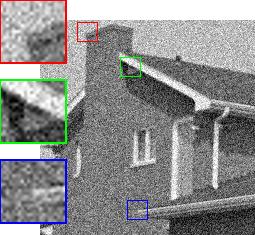} &
        \includegraphics[width = 0.15\textwidth]{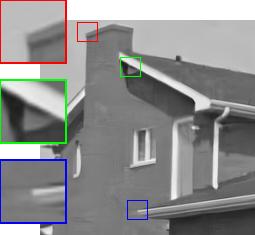} &
        \includegraphics[width = 0.15\textwidth]{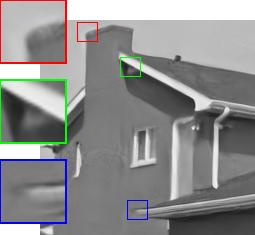} \\ 

        & \small{\textit{face} ($32.15$ dB)} & \small{\textit{street} ($31.94$ dB)} \\        
        \includegraphics[width = 0.15\textwidth]{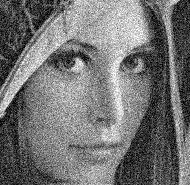} &
        \includegraphics[width = 0.15\textwidth]{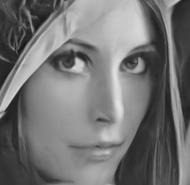} &                 					\includegraphics[width = 0.15\textwidth]{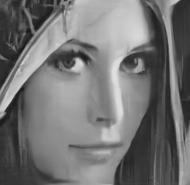} \\  
 
	\end{tabular}   \\
	\caption{\small \textbf{Cross-class denoising.} 
Examples of using correct (middle) and mismatched (right) class-specific denoisers on images with Gaussian noise, $\sigma=25$. 
PSNR values and denoiser type are listed above each denoised image. The reader is encouraged to zoom in for a better view of the artifacts.
}
    \label{fig_using_wrong_denoiser}
\end{figure}

The top row of Fig.~\ref{fig_using_wrong_denoiser} presents a comparison of denoisers fine-tuned to the \textit{street} and \textit{face} classes applied to a noisy face image. The street-specific denoiser produces noticeable artifacts around the eye, cheek and hair areas. Moreover, the edges appear too sharp and seem to favor horizontal and vertical edges. This is not very surprising as street images contain mainly man-made rectangle-shaped structures. In the second row, strong artifacts appear on the hamster's fur when the image is processed by the \textit{living room}-specific denoiser, while the \textit{pet}-specific denoiser produces a more naturally looking result.
Examples on the canonical images \textit{House} and \textit{Lena} are presented in the bottom two rows. Notice how the \textit{street}-specific denoiser reconstructs sharp boundaries of the building whereas the \textit{face}-specific counterpart smears them. 

\begin{figure}[tb]
    \begin{tabular}{ c c}
    	\includegraphics[width = 0.225\textwidth]{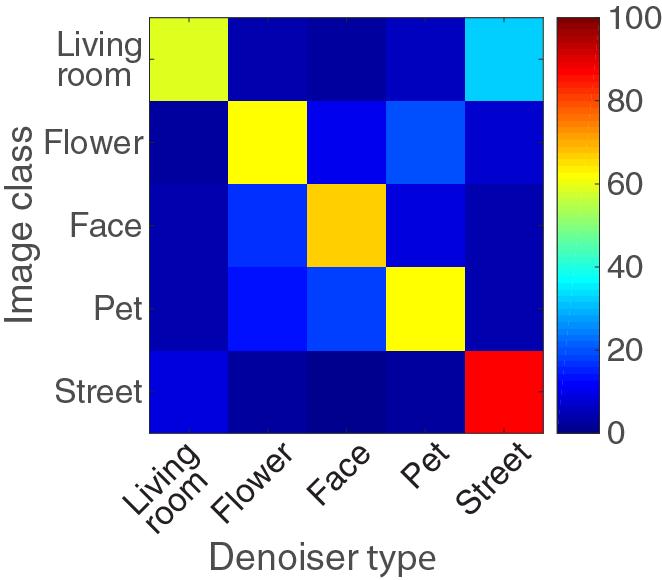} &
		\includegraphics[width = 0.225\textwidth]{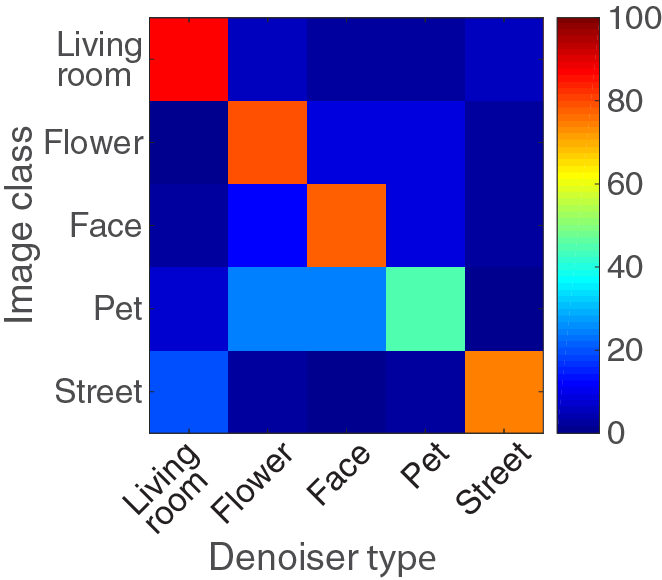} \\ 
	\end{tabular}   \\
    \vspace{-2mm}
	\caption{\textbf{Cross class denoising} for Gaussian noise with $\sigma=25$ (left) and Poisson noise with peak value $8$ (right). The rows represent the tested semantic class of images while the columns the class-aware denoisers used. The $(i,j)$-th element in the matrix shows the probability of the $j$-th class-aware denoiser to outperform all other class-aware denoisers on the $i$-th class of images.
    }
\label{fig_confusion_matrix}
\end{figure}

To quantify the effect of class mismatch, we evaluate the percentage of wins of every fine-tuned denoiser on each type of image class. A 'win' means that a particular denoiser produces the highest PSNR among all the others. Fig.~\ref{fig_confusion_matrix} presents a matrix with all the combinations of class-specific denoisers and image classes. 

\subsection{Video denoising}
\label{sec_exp_video}
This section demonstrates how our method can be extended to video denoising. To this end, we propose to alter our network architecture described in Fig.~\ref{fig_denoiseNet}. The main difference between our image and video architectures lies in how we provide the inputs to the network and what is the output we wish to obtain. As opposed to image denoising, where we feed a single gray-scale image as input, for videos we feed a sequence of $T$ stacked gray-scale images. We request the network to estimate the noise that should be subtracted from the central image of the input sequence. In practice, this only implies an increase in the depth of the kernels at the first layer of the network from 1 to $T$ and is therefore relatively inexpensive. When applying our model to RGB videos, we denoise each color channel independently.

To demonstrate the efficacy of our method for high-resolution videos contaminated by Poisson noise with peak values of 8 and 30 (independent for each color channel), we compared our architecture with 1, 5, and 11 frame-long sequences, using the following leading methods as a reference:  I+VST+BM3D \cite{Azzari16Variance} and VST+VBM4D \cite{Dabov07Video, Maggioni12Video}. 

Our training set was comprised of 17 YouTube videos $1920\times1080$ with varying content and length (between 4 and 10 minuted long) comprising 161K training frames. All algorithms were tested on 6 unseen videos, 2000 frames from each video (frames 20-2020 to give all temporal algorithms a warm-up period of 20 frames), 12000 frames in total. The full list of videos used in our experiments appears in Appendix~\ref{sec:video_list}.

I+VST+BM3D was applied to each image color channel separately. Due to the heavy memory usage of VBM4D, while testing this method, videos were sliced to blocks of 30 frames with size $384\times270$, which were stitched back after being denoised. Moreover, as this method was designed for Gaussian noise, the Anscombe transform was used to whiten the noise.

Tables~\ref{tab_video_peak8_psnr} and \ref{tab_video_peak30_psnr} \ed{present} quantitative results, in which a breakdown of PSNR performance is given by video. It is evident that our proposed technique outperforms the existing techniques for video denoising. As expected, our method improves when more frames are used for denoising.

\begin{figure*}[]
	\small
	\centering
    \begin{tabular}{c@{\hskip 0.01\textwidth}c@{\hskip 0.01\textwidth}c}
		\includegraphics[width = 0.32\textwidth]{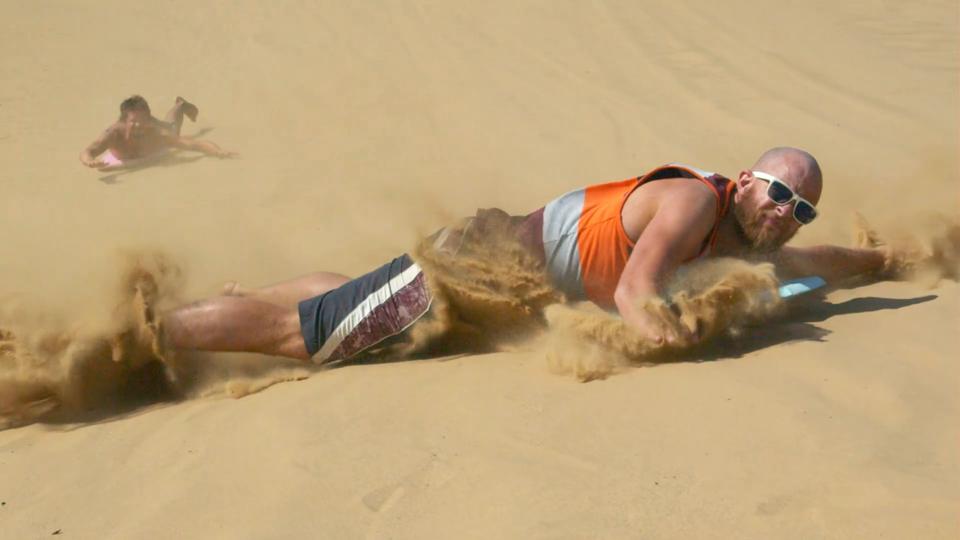} &
		\includegraphics[width = 0.32\textwidth]{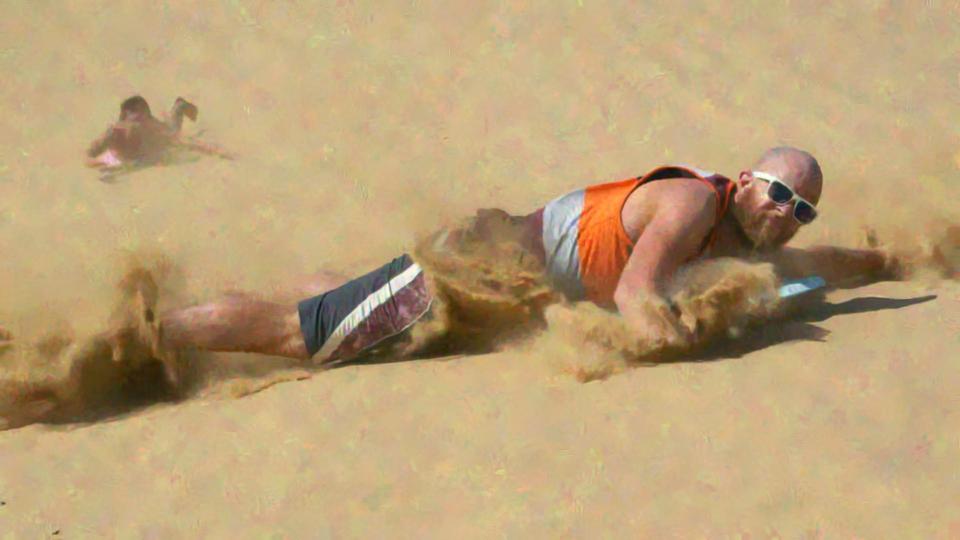} &
        \includegraphics[width = 0.32\textwidth]{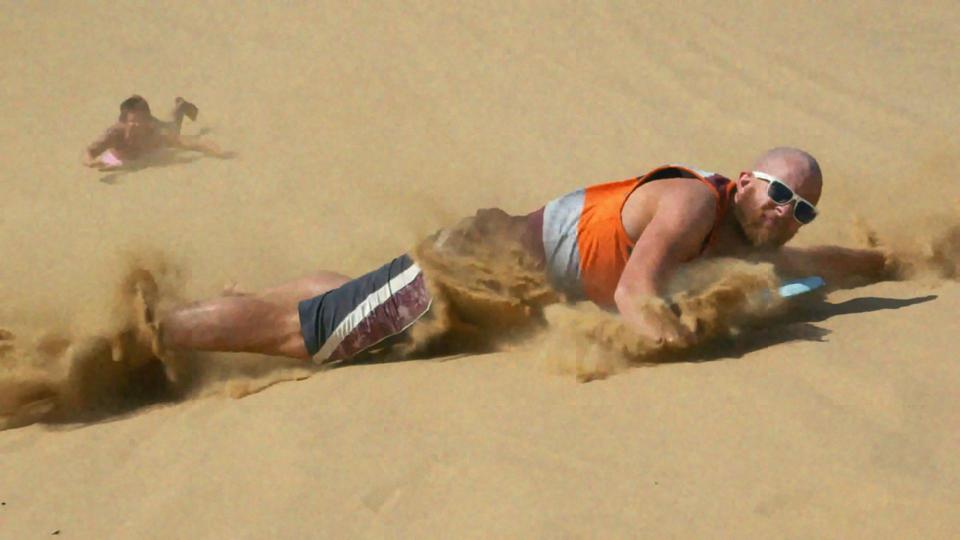} \\        
	\end{tabular}  
   	\begin{tabular}{c@{\hskip 0.01\textwidth}c@{\hskip 0.01\textwidth}c@{\hskip 0.01\textwidth}c@{\hskip 0.01\textwidth}c@{\hskip 0.01\textwidth}c}
		\includegraphics[width = 0.155\textwidth]{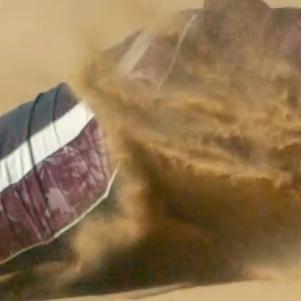} &
        \includegraphics[width = 0.155\textwidth]{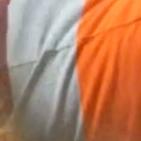} &
		\includegraphics[width = 0.155\textwidth]{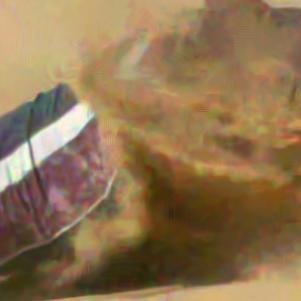} &
        \includegraphics[width = 0.155\textwidth]{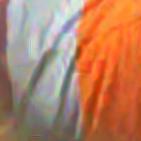} &
   		\includegraphics[width = 0.155\textwidth]{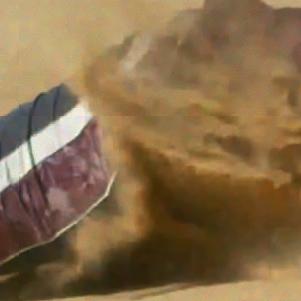} &
        \includegraphics[width = 0.155\textwidth]{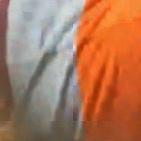} \\
        
		\includegraphics[width = 0.155\textwidth]{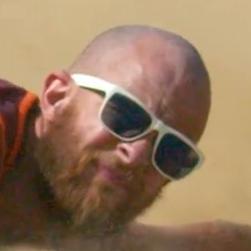} &
        \includegraphics[width = 0.155\textwidth]{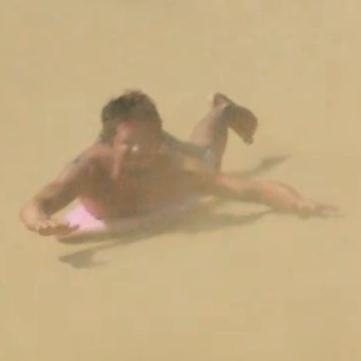} &
		\includegraphics[width = 0.155\textwidth]{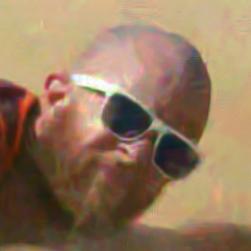} &
        \includegraphics[width = 0.155\textwidth]{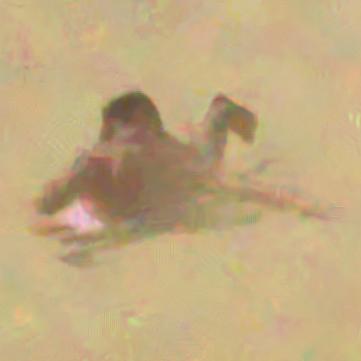} &
   		\includegraphics[width = 0.155\textwidth]{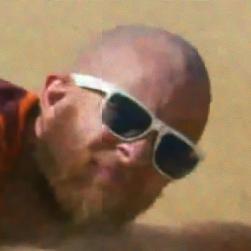} &
        \includegraphics[width = 0.155\textwidth]{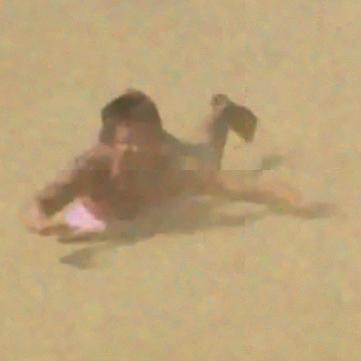} \\
        
        \multicolumn{2}{c}{Ground truth}&
        \multicolumn{2}{c}{I+VST+BM3D}&
        \multicolumn{2}{c}{vBM4D} \\
        
	\end{tabular}  
        \begin{tabular}{c@{\hskip 0.01\textwidth}c@{\hskip 0.01\textwidth}c}
		\includegraphics[width = 0.32\textwidth]{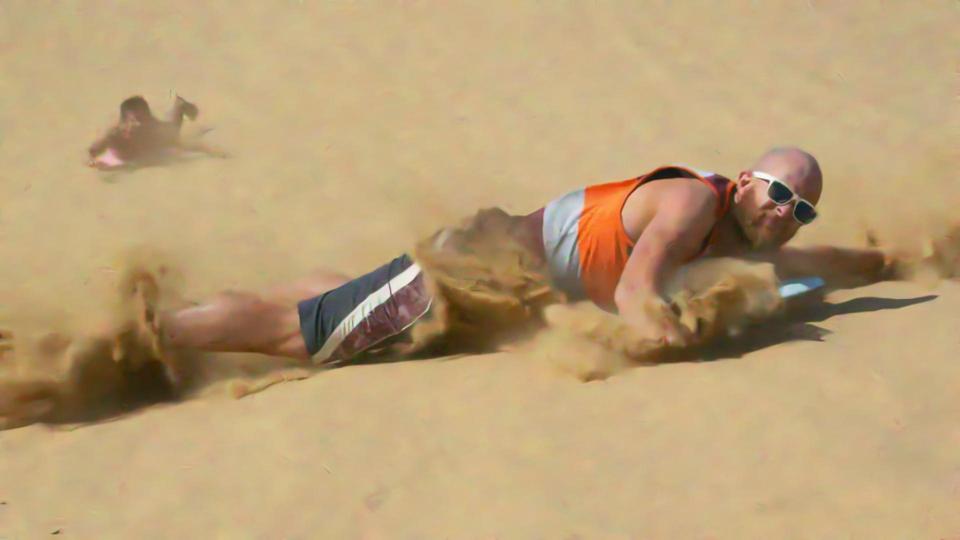} &
		\includegraphics[width = 0.32\textwidth]{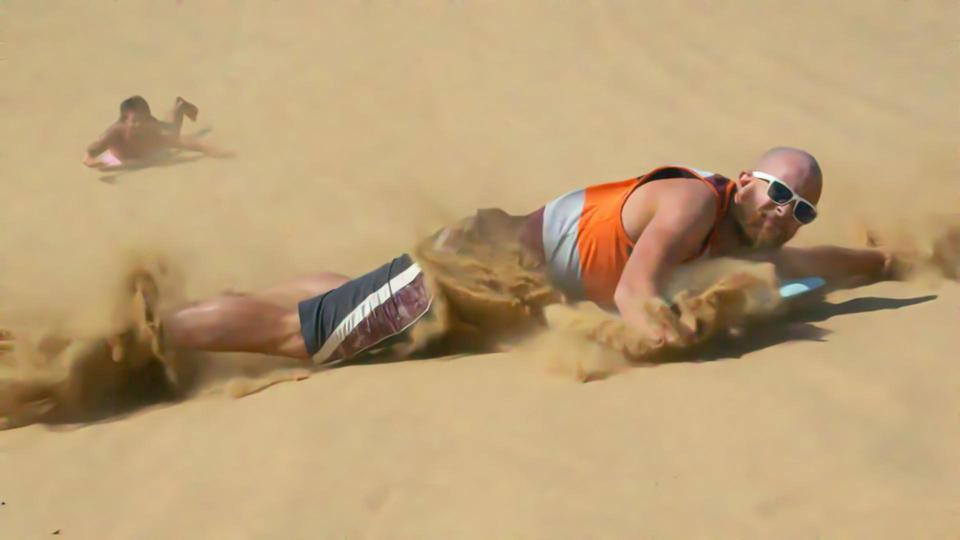} &
        \includegraphics[width = 0.32\textwidth]{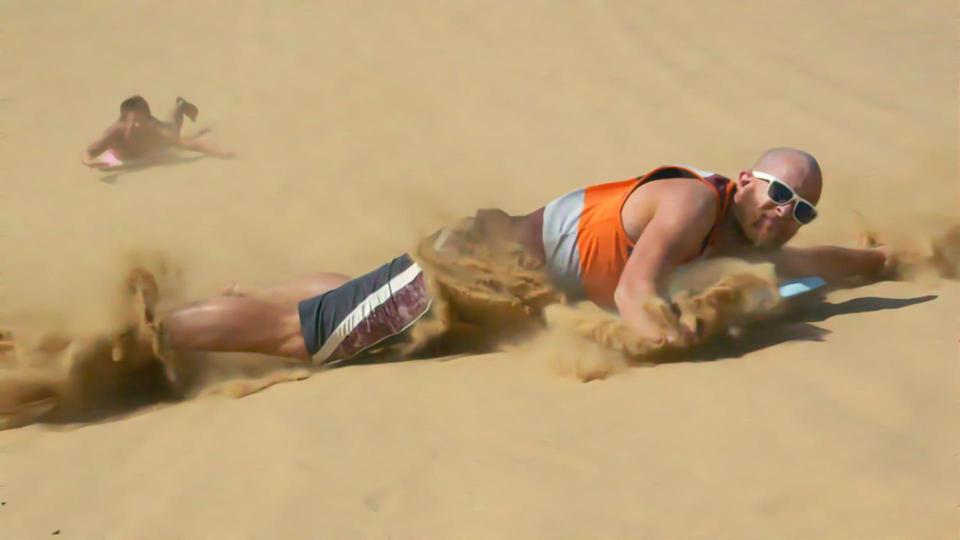} \\        
	\end{tabular}  
   	\begin{tabular}{c@{\hskip 0.01\textwidth}c@{\hskip 0.01\textwidth}c@{\hskip 0.01\textwidth}c@{\hskip 0.01\textwidth}c@{\hskip 0.01\textwidth}c}
		\includegraphics[width = 0.155\textwidth]{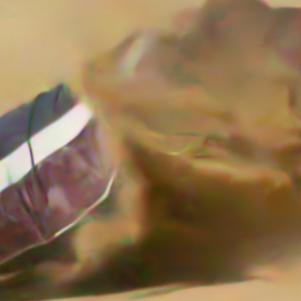} &
        \includegraphics[width = 0.155\textwidth]{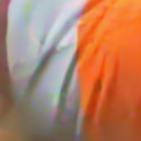} &
		\includegraphics[width = 0.155\textwidth]{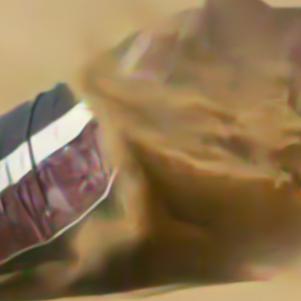} &
        \includegraphics[width = 0.155\textwidth]{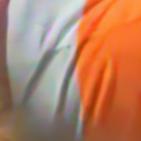} &
        \includegraphics[width = 0.155\textwidth]{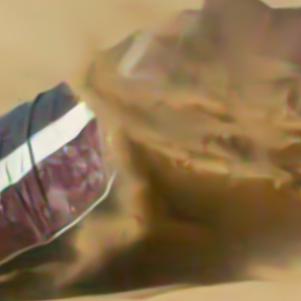} &
        \includegraphics[width = 0.155\textwidth]{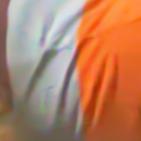} \\
        
		\includegraphics[width = 0.155\textwidth]{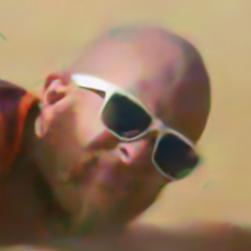} &
        \includegraphics[width = 0.155\textwidth]{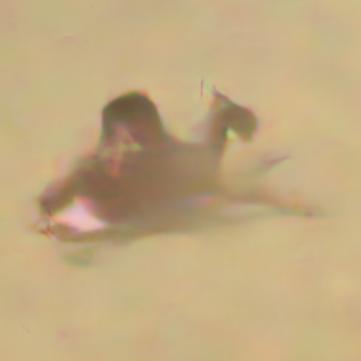} &
		\includegraphics[width = 0.155\textwidth]{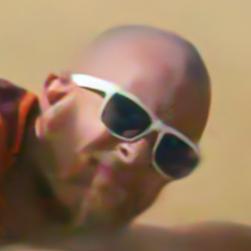} &
        \includegraphics[width = 0.155\textwidth]{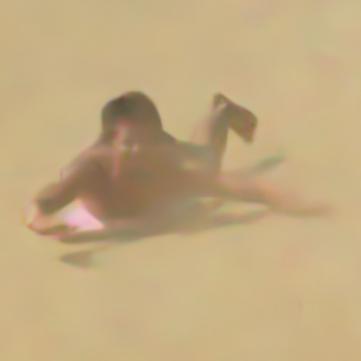} &
        \includegraphics[width = 0.155\textwidth]{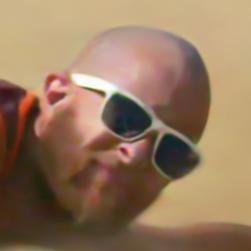} &
        \includegraphics[width = 0.155\textwidth]{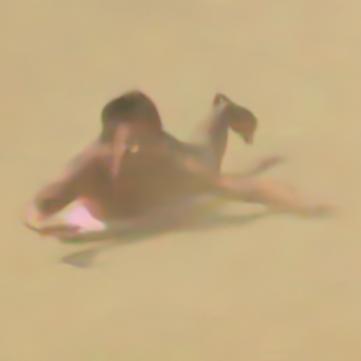} \\
        
        \multicolumn{2}{c}{Ours (1)} &
        \multicolumn{2}{c}{Ours (5)} &
        \multicolumn{2}{c}{Ours (11)} \\
        
	\end{tabular} 
    \smallskip 
	\caption{\textbf{Poisson Video Denoising Example. } Denoising comparison of a low-light video  with a peak value 8. Intensities were scaled for display purposes. }
\label{fig_vid2_peak8}
\end{figure*}

\begin{figure*}[]
	\small
	\centering
    \begin{tabular}{c@{\hskip 0.01\textwidth}c@{\hskip 0.01\textwidth}c}
		\includegraphics[width = 0.32\textwidth]{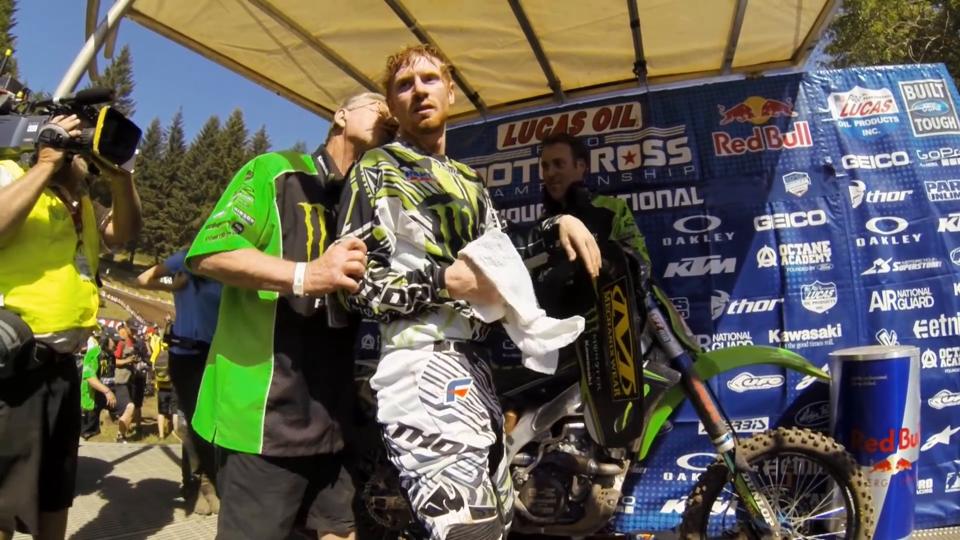} &
		\includegraphics[width = 0.32\textwidth]{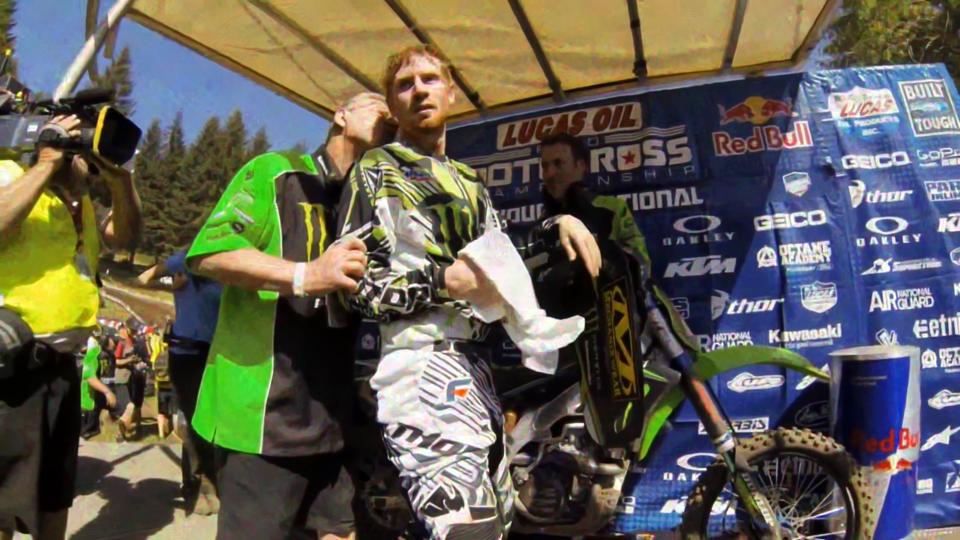} &
        \includegraphics[width = 0.32\textwidth]{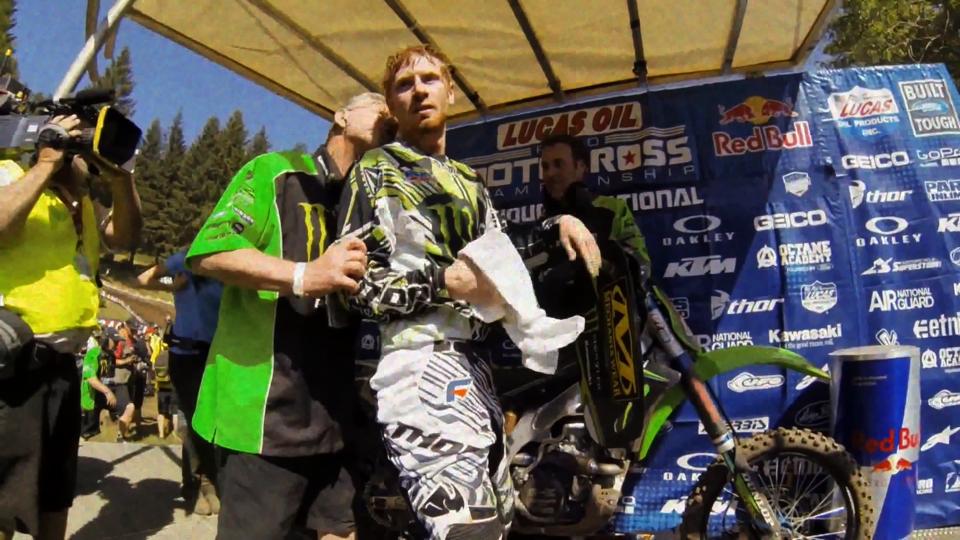} \\        
	\end{tabular}  
   	\begin{tabular}{c@{\hskip 0.01\textwidth}c@{\hskip 0.01\textwidth}c@{\hskip 0.01\textwidth}c@{\hskip 0.01\textwidth}c@{\hskip 0.01\textwidth}c}
		\includegraphics[width = 0.155\textwidth]{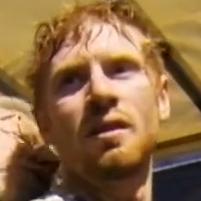} &
        \includegraphics[width = 0.155\textwidth]{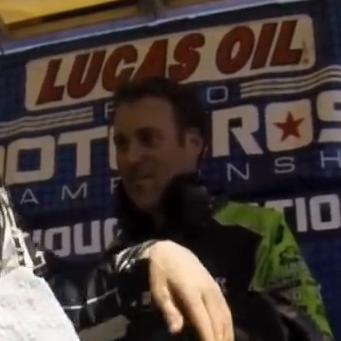} &
		\includegraphics[width = 0.155\textwidth]{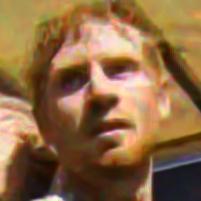} &
        \includegraphics[width = 0.155\textwidth]{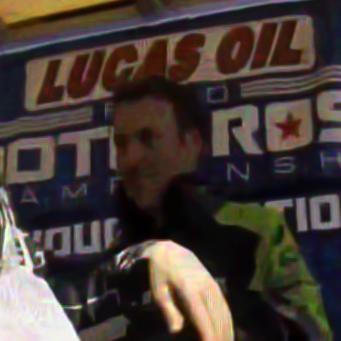} &
   		\includegraphics[width = 0.155\textwidth]{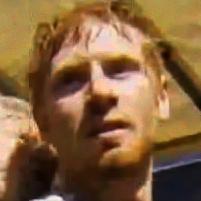} &
        \includegraphics[width = 0.155\textwidth]{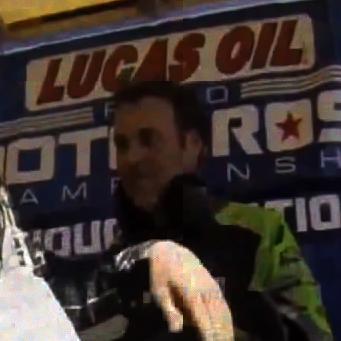} \\
        
		\includegraphics[width = 0.155\textwidth]{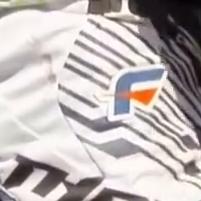} &
        \includegraphics[width = 0.155\textwidth]{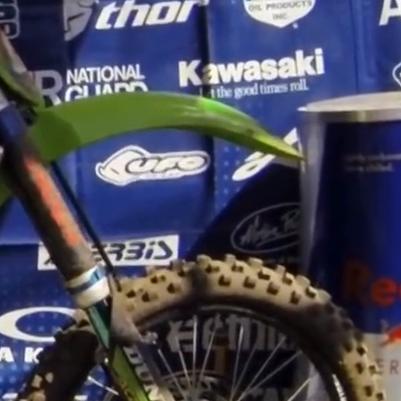} &
		\includegraphics[width = 0.155\textwidth]{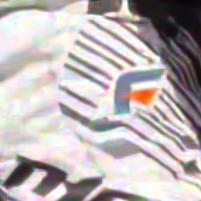} &
        \includegraphics[width = 0.155\textwidth]{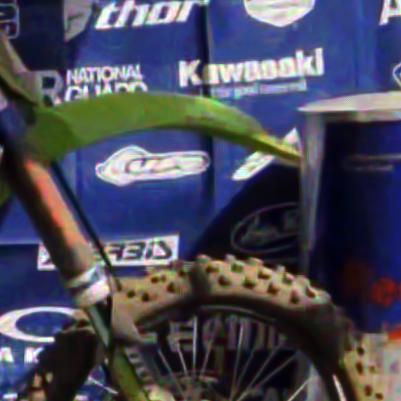} &
   		\includegraphics[width = 0.155\textwidth]{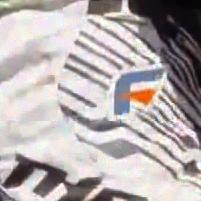} &
        \includegraphics[width = 0.155\textwidth]{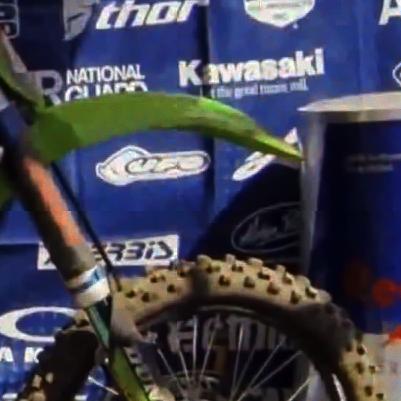} \\
        
        \multicolumn{2}{c}{Ground truth}&
        \multicolumn{2}{c}{I+VST+BM3D}&
        \multicolumn{2}{c}{vBM4D} \\
        
	\end{tabular}  
        \begin{tabular}{c@{\hskip 0.01\textwidth}c@{\hskip 0.01\textwidth}c}
		\includegraphics[width = 0.32\textwidth]{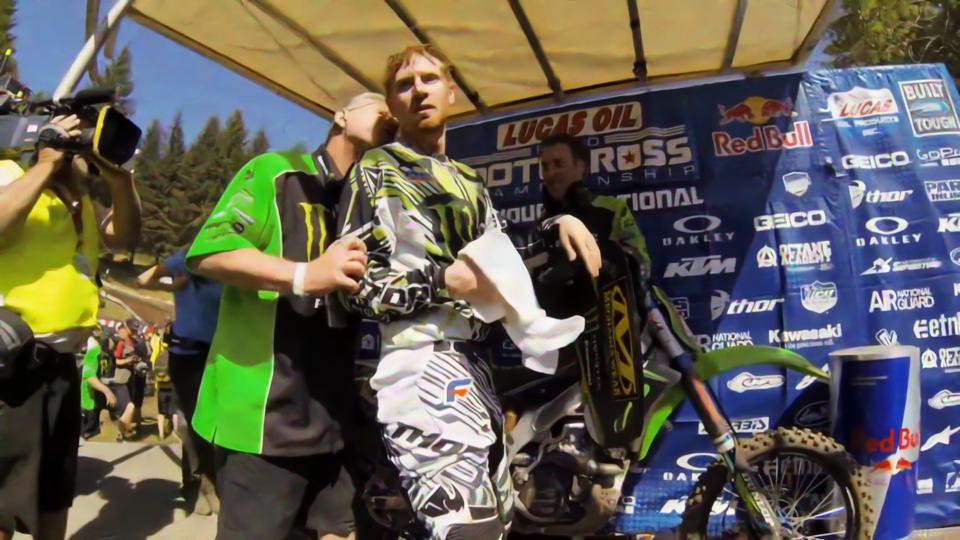} &
		\includegraphics[width = 0.32\textwidth]{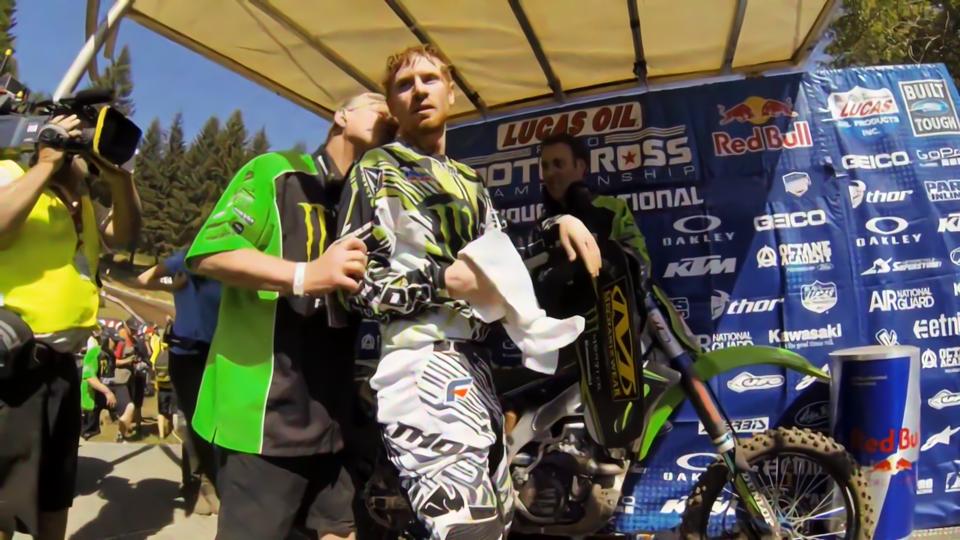} &
        \includegraphics[width = 0.32\textwidth]{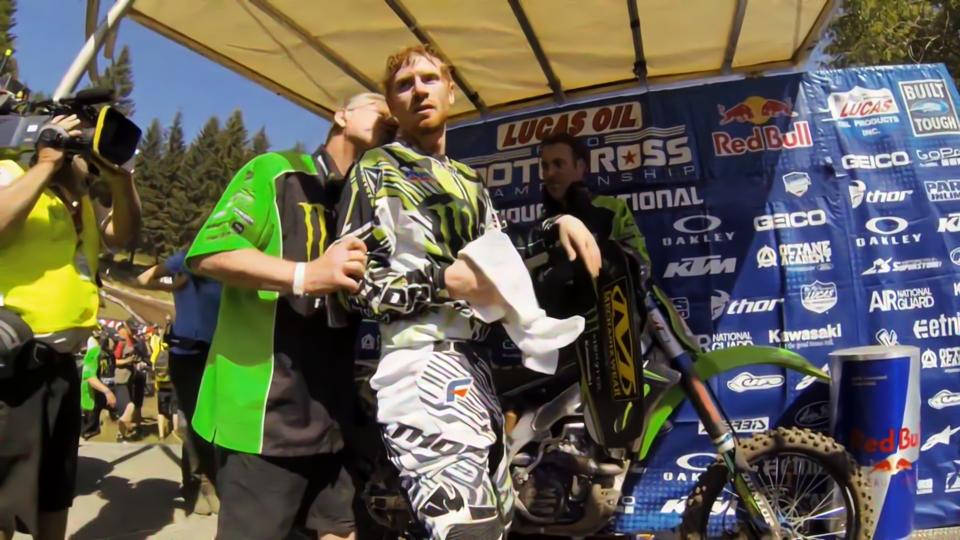} \\        
	\end{tabular}  
   	\begin{tabular}{c@{\hskip 0.01\textwidth}c@{\hskip 0.01\textwidth}c@{\hskip 0.01\textwidth}c@{\hskip 0.01\textwidth}c@{\hskip 0.01\textwidth}c}
		\includegraphics[width = 0.155\textwidth]{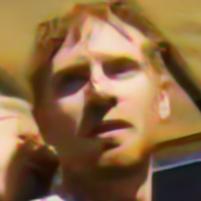} &
        \includegraphics[width = 0.155\textwidth]{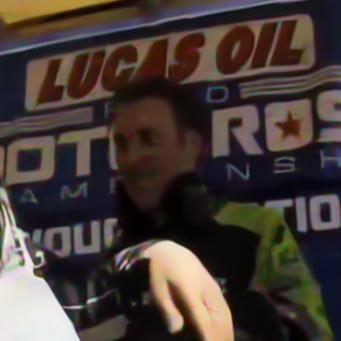} &
		\includegraphics[width = 0.155\textwidth]{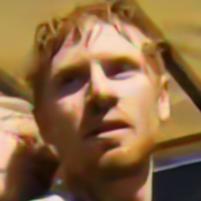} &
        \includegraphics[width = 0.155\textwidth]{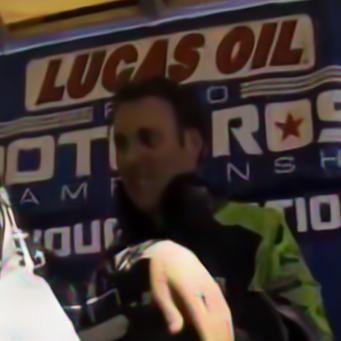} &
        \includegraphics[width = 0.155\textwidth]{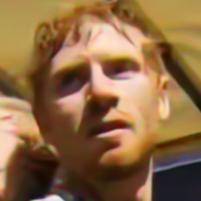} &
        \includegraphics[width = 0.155\textwidth]{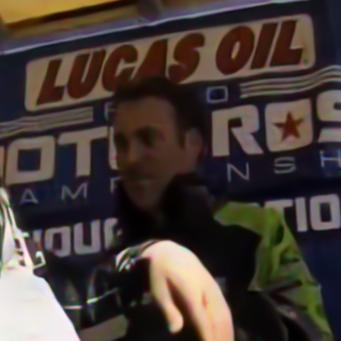} \\
        
		\includegraphics[width = 0.155\textwidth]{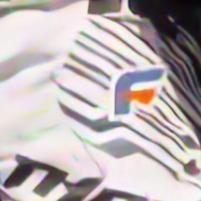} &
        \includegraphics[width = 0.155\textwidth]{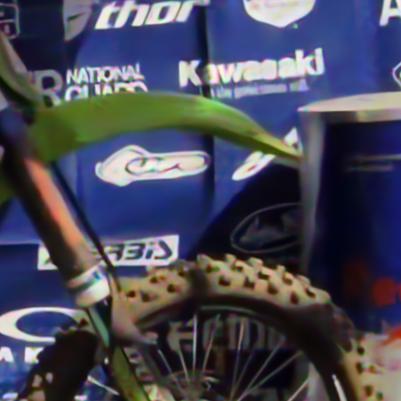} &
		\includegraphics[width = 0.155\textwidth]{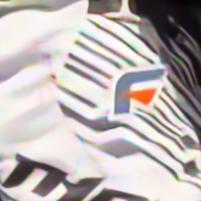} &
        \includegraphics[width = 0.155\textwidth]{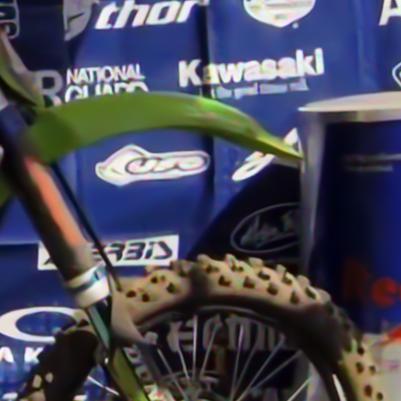} &
        \includegraphics[width = 0.155\textwidth]{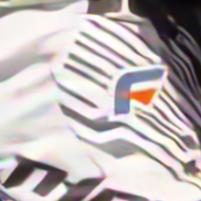} &
        \includegraphics[width = 0.155\textwidth]{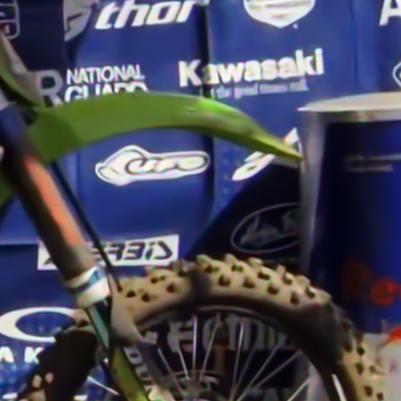} \\
        
        \multicolumn{2}{c}{Ours (1)} &
        \multicolumn{2}{c}{Ours (5)} &
        \multicolumn{2}{c}{Ours (11)} \\
        
	\end{tabular} 
    \smallskip 
	\caption{\textbf{Poisson Video Denoising Example. } Denoising comparison of a low-light video  with a peak value 8. Intensities were scaled for display purposes. }
\label{fig_vid24_peak8}
\end{figure*}

Figs.~\ref{fig_vid2_peak8} and \ref{fig_vid24_peak8} \ed{present} some qualitative examples. The time required per average frame are $450$ seconds for I+VST+BM3D, $97$ seconds for VBM3D, $3$, $3.9$, and $4.8$ seconds for our network with $T=1$, 5, and 11 receptively.

Training was done in a similar fashion to our image models while using sequences of length $T$ with random patches of size $128\times128$. We use Adam optimizer with a learning rate of 1$e$$-$4, 
patches are randomly flipped horizontally and the batch size is 64. The models with $T$=1, 5, and 11 converged after 130K,170K, and 330K iterations respectively.

\begin{table}[tb]
\centering
\small
  \begin{tabular}{ c@{\hskip 0.01\textwidth}|c@{\hskip 0.01\textwidth}c@{\hskip 0.01\textwidth}c@{\hskip 0.01\textwidth}c@{\hskip 0.01\textwidth}c}
    \hline \hline
    	Video 		& VBM4D 	& I+V+BM3D		& Ours (1)	& Ours (5)  	& Ours (11) 	\\ \hline
        $\#$1		& $27.07$	& $31.57$		& $32.76$		& $33.36$		& $\textbf{33.42}$	\\        
        $\#$2		& $26.92$	& $32.03$		& $34.04$		& $35.13$		& $\textbf{35.26}$	\\        
        $\#$3		& $28.91$	& $30.80$		& $31.88$		& $32.78$		& $\textbf{32.88}$	\\        
        $\#$4		& $26.32$	& $30.58$		& $32.06$		& $32.45$		& $\textbf{32.54}$	\\        
        $\#$5		& $27.35$	& $30.38$		& $31.99$		& $33.44$		& $\textbf{33.79}$	\\        
        $\#$6		& $26.85$	& $29.36$		& $30.17$		& $30.96$		& $\textbf{31.17}$	\\ \hline
        Avg.		& $27.24$	& $30.79$		& $32.15$		& $33.02$		& $\textbf{33.17}$	\\
        
     \hline\hline
  \end{tabular}  
  \vspace{2mm}
\caption{\small \textbf{Poisson video denoising for peak$=8$.} PSNR denoising performance for Poisson noise with peak value of 30. We compare our method with $T$= 1, $5$ and $11$ to VBM4D and I+VST+BM3D.
} 
\label{tab_video_peak8_psnr}	
\end{table}

\begin{table}[tb]
\centering
\small
  \begin{tabular}{ c@{\hskip 0.01\textwidth}|c@{\hskip 0.01\textwidth}c@{\hskip 0.01\textwidth}c@{\hskip 0.01\textwidth}c@{\hskip 0.01\textwidth}c}
    \hline \hline
    	Video 		& VBM4D 	& I+V+BM3D		& Ours (1)		& Ours (5)  	& Ours (11) 	\\ \hline
        $\#$1		& $33.01$	& $34.76$		& $35.86$		& $35.86$		& $\textbf{36.16}$		\\
        $\#$2		& $33.16$	& $35.16$		& $37.65$		& $37.68$		& $\textbf{38.20}$		\\
        $\#$3		& $34.30$	& $34.13$		& $35.35$		& $35.73$		& $\textbf{36.04}$		\\
        $\#$4		& $31.21$	& $33.23$		& $35.24$		& $35.17$		& $\textbf{35.72}$		\\
        $\#$5		& $34.08$	& $33.63$		& $35.42$		& $36.41$		& $\textbf{36.99}$		\\
        $\#$6		& $32.26$	& $32.76$		& $33.58$		& $34.07$		& $\textbf{34.46}$		\\ \hline
        Avg.		& $33.00$	& $33.94$		& $35.52$		& $35.82$		& $\textbf{36.26}$		\\
        
     \hline\hline
  \end{tabular}  
  \vspace{2mm}
\caption{\small \textbf{Poisson video denoising for peak$=30$.} PSNR denoising performance for Poisson noise with peak value of 30. We compare our method with $T$= 1, $5$ and $11$ to VBM4D and I+VST+BM3D.
} 
\label{tab_video_peak30_psnr}		
\end{table}

\subsection{Ablation study}
\label{sec_exp_ablation}
\ed{In the following experiments, we evaluated the effect of various hyper-parameters of our network on its performance. In all cases, we trained on the PASCAL training set with an additive Gaussian noise with $\sigma=25$ using the same methodology described in Section \ref{sec_implementation}. Testing was performed on the 1000 test images of PASCAL.

\noindent\paragraph{Batch-normalization}
We tested the effect of batch-normalization (BN) on our basic network with $20$ layers, $64$ kernels per layer, and intermediate skip connections at each layer. BN was added before each ReLU non-linearity. The PSNR on the test set was $30.25$ dB with BN and $30.33$ dB without it. Note that other recent works for various image processing tasks such as blind deblurring \cite{Nah17Deep} and burst denoising \cite{godard2017deep} also reported no advantage in the use of BN. 

\noindent\paragraph{Skip connections}
We evaluated the contribution of the intermediate skip connection at each layer (the dark blue blocks in Fig.~\ref{fig_denoiseNet}). To this end we trained two networks that are identical except that one has a skip connection at each layer (as in Fig.~\ref{fig_denoiseNet}), while the other have no skip connections at all (has $64$ feedforwad filters at each layer). Both networks had $20$ layers. 
The network without the skip connections achieved a PSNR of 
$30.26$ dB, while the one with the proposed skip connections achieved $30.33$ dB.


\noindent\paragraph{Network depth}
We tested the performance of our network for different depths ranging from $5$ layers up to $40$. In all cases the networks had 64 kernels per layer and intermediate skip connections. The results are presented in Table \ref{tab_net_depth}. We selected a depth of 20 layers as beyond it, the network has diminishing returns. With 20 layers we achieve a similar performance to a 40 layers network but with half the number of trainable parameters, memory and inference time.

\begin{table}[t]
\centering
\small
\begin{tabular}{ l c c c c  }
    \hline\hline
    depth & 5 & 10 & 20 & 40 \\ \hline
    PSNR & 29.25 & 30.11 & 30.33 & 30.37 \\
    \hline\hline
  \end{tabular}  
  \vspace{1mm}
\caption{\small \textbf{Network depth.} \ed{Average PSNR values on $1K$ PASCAL image test-set for different network depths.}}
\label{tab_net_depth}	
\end{table}

\noindent\paragraph{Number of kernels}

We tested the effect of the number kernels at each network layer for a network with 20 layers with skip connections. The values tested were 16, 32, 64, and 96 kernels per layer. The results are presented in Table~\ref{tab_net_kernels}. We selected to use 64 kernels per layer as this reduces by half the trainable parameters compared to having 96 kernels per layer and in addition we saw that adding more kernels usually resulted in poorer generalization of the network on other datasets such as BSD (when we did not train on those datasets but rather only tested on them). Interestingly, the authors of IRCNN found that the same number of kernels led to the best performance.

\begin{table}[t]
\centering
\small
\begin{tabular}{ l c c c c  }
    \hline\hline
    kernels & 16 & 32 & 64 & 96 \\ \hline
    PSNR & 29.84 & 30.17 & 30.33 & 30.40 \\
    \hline\hline
  \end{tabular}  
  \vspace{1mm}
\caption{\small \textbf{Number of kernels per layer.} \ed{Average PSNR values on $1K$ PASCAL image test-set for different numbers of kernels per network layer.}}
\label{tab_net_kernels}	
\end{table}

}

\begin{figure*}[]
	\centering    
    \begin{tabular}{c@{\hskip 0.005\textwidth}c@{\hskip 0.005\textwidth}c@{\hskip 0.005\textwidth}c@{\hskip 0.005\textwidth}c}    
            Ground Truth & Noisy & I+VST+BM3D & \ed{IRCNN} & Ours \\
            
		\includegraphics[width = 0.19\textwidth]{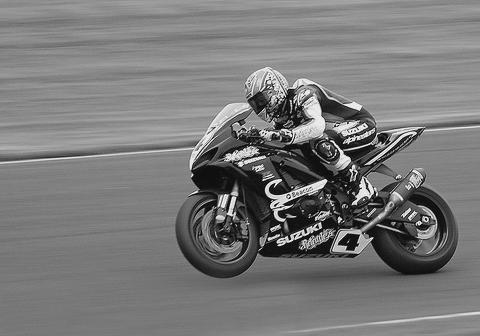} &
		\includegraphics[width = 0.19\textwidth]{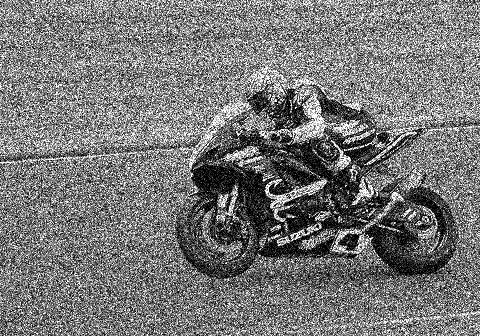} &
        \includegraphics[width = 0.19\textwidth]{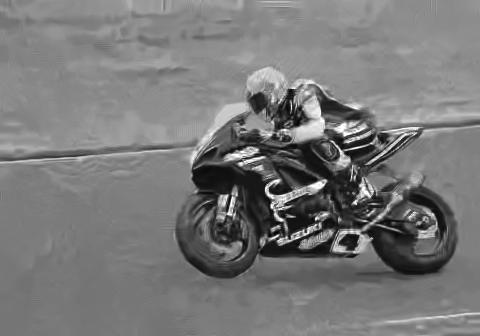} &
        \includegraphics[width = 0.19\textwidth]{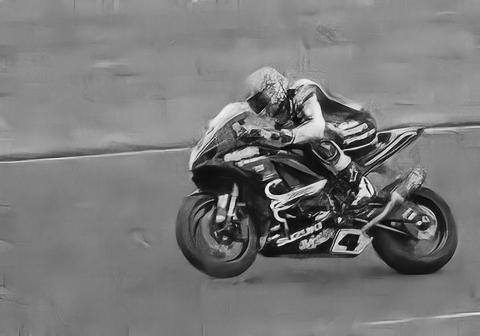} &
        \includegraphics[width = 0.19\textwidth]{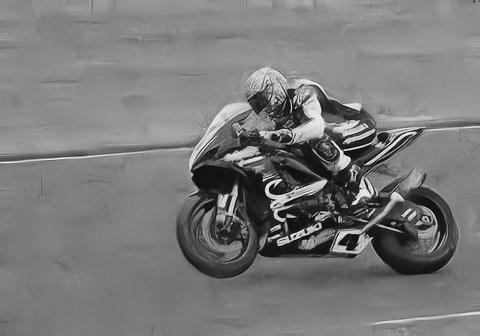}  \\ 
          &\medskip  & $24.56$ dB & \ed{$25.62$ dB} & $25.83$ dB \\          

		\includegraphics[width = 0.19\textwidth]{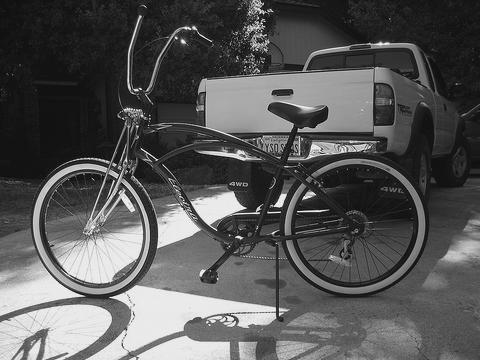} &
		\includegraphics[width = 0.19\textwidth]{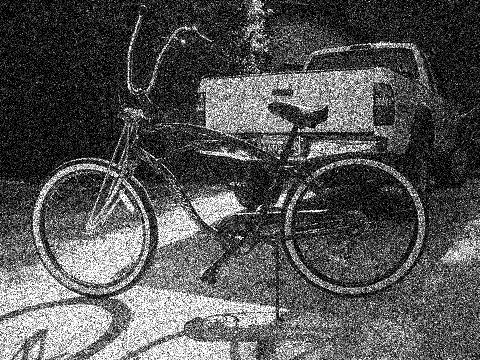} &
        \includegraphics[width = 0.19\textwidth]{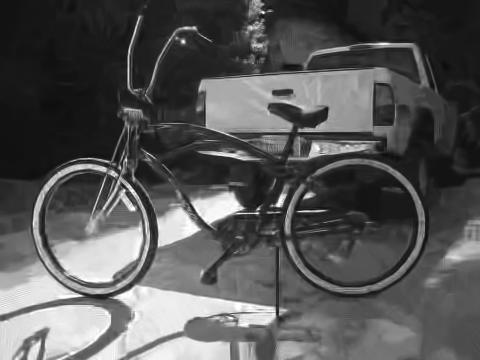} &
        \includegraphics[width = 0.19\textwidth]{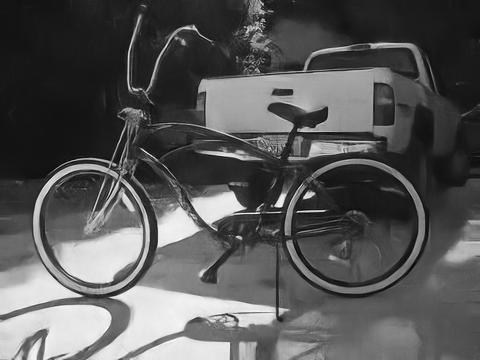} &
        \includegraphics[width = 0.19\textwidth]{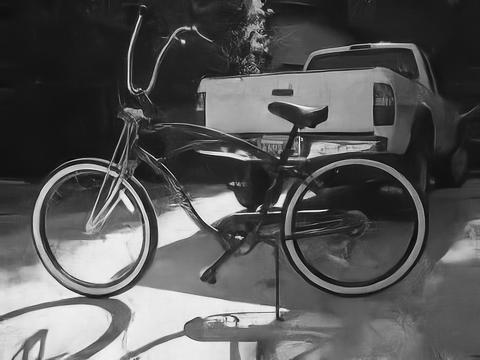}  \\ 
          &\medskip  & $24.76$ dB & \ed{$25.78$} dB & $26.00$ dB \\ 

		\includegraphics[width = 0.19\textwidth]{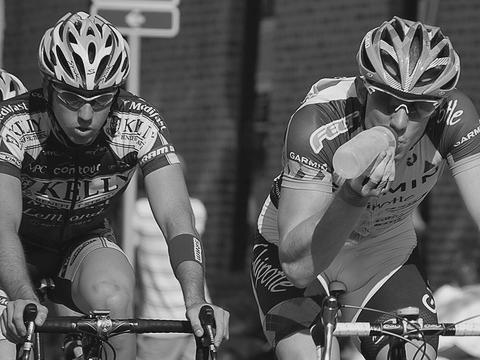} &
		\includegraphics[width = 0.19\textwidth]{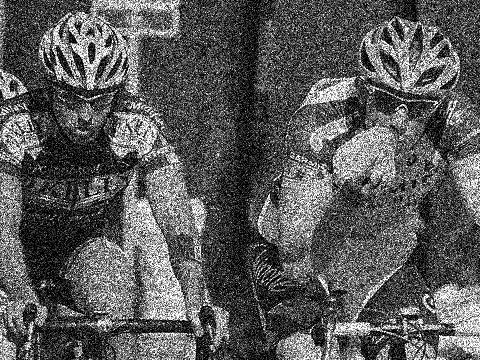} &
        \includegraphics[width = 0.19\textwidth]{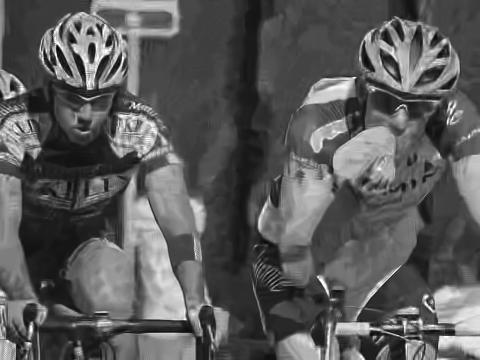} &
        \includegraphics[width = 0.19\textwidth]{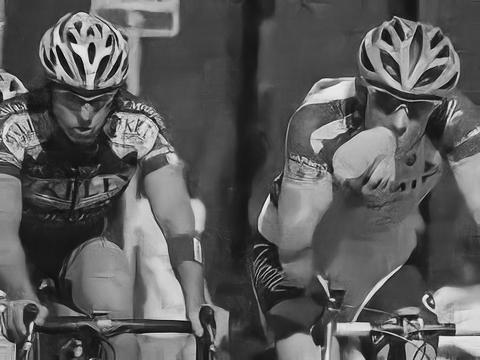} &
        \includegraphics[width = 0.19\textwidth]{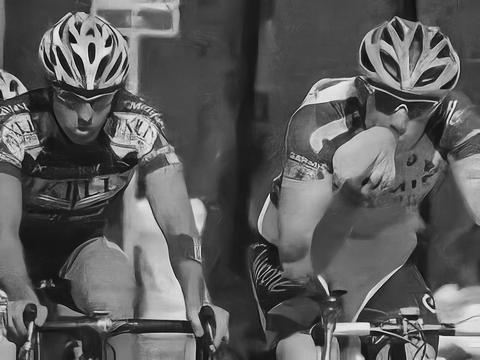}  \\ 
         \medskip &  & $23.40$ dB & \ed{$24.47$} dB & $24.66$ dB \\ 

		\includegraphics[width = 0.19\textwidth]{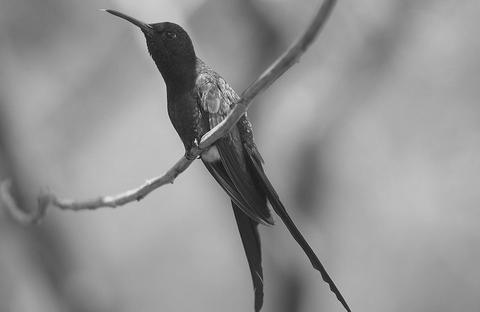} &
		\includegraphics[width = 0.19\textwidth]{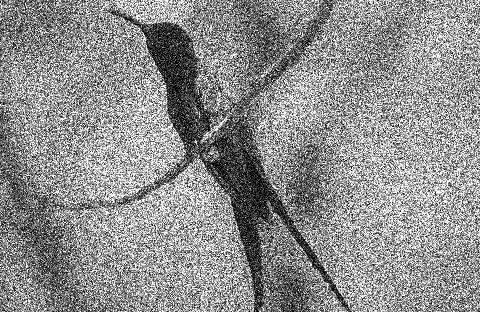} &
        \includegraphics[width = 0.19\textwidth]{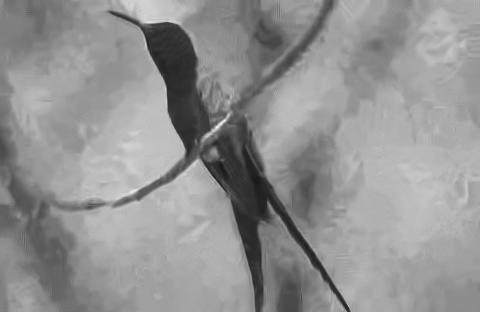} &
        \includegraphics[width = 0.19\textwidth]{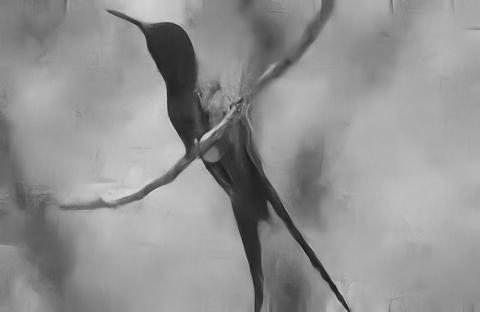} &
        \includegraphics[width = 0.19\textwidth]{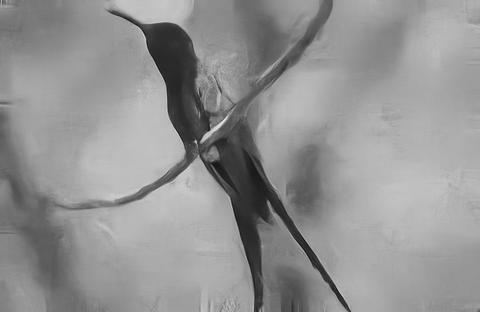}  \\ 
          &  & $30.14$ dB & \ed{$31.43$dB} & $31.61$ dB \\

	\end{tabular} 
    \caption{\ed{\textbf{Poisson denoising examples from PASCAL.} For Poisson noise with peak $8$. PSNR values appear below each of the images.}
\label{fig_large_pascal}}
\end{figure*}
\begin{figure*}[h]
	\centering   
    \tiny

   	\begin{tabular}{c@{\hskip 0.005\textwidth}c@{\hskip 0.005\textwidth}c@{\hskip 0.005\textwidth}c@{\hskip 0.005\textwidth}c@{\hskip 0.005\textwidth}c@{\hskip 0.005\textwidth}c@{\hskip 0.005\textwidth}c}
    		
    		\includegraphics[width = 0.115\textwidth]{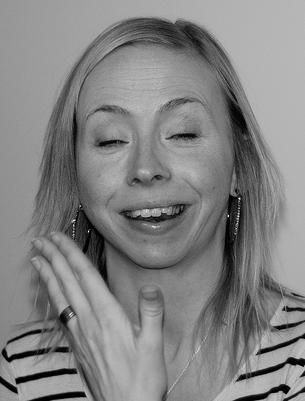} &
    		\includegraphics[width = 0.115\textwidth]{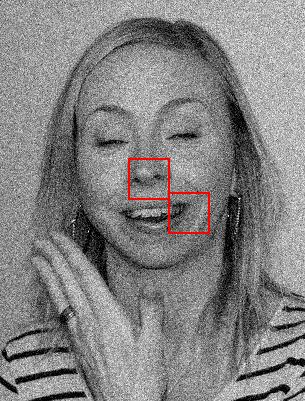} &
    		\includegraphics[width = 0.115\textwidth]{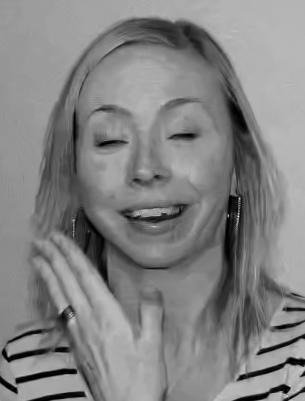} &   		
    		\includegraphics[width = 0.115\textwidth]{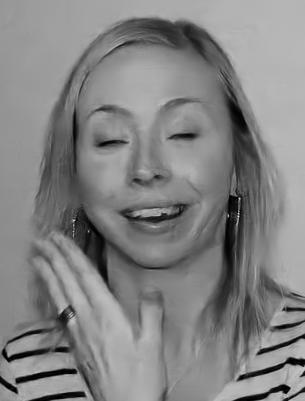} &
    		\includegraphics[width = 0.115\textwidth]{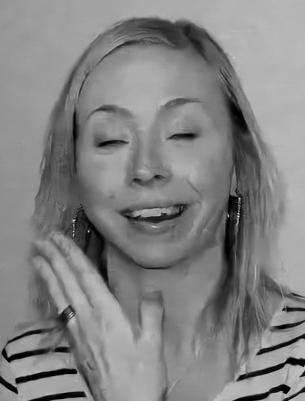} &
            \includegraphics[width = 0.115\textwidth]{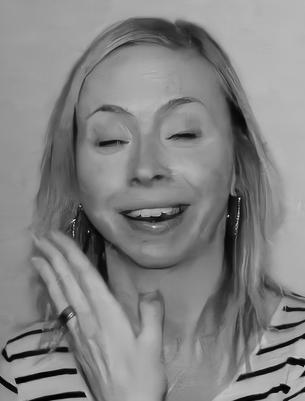} &
            \includegraphics[width = 0.115\textwidth]{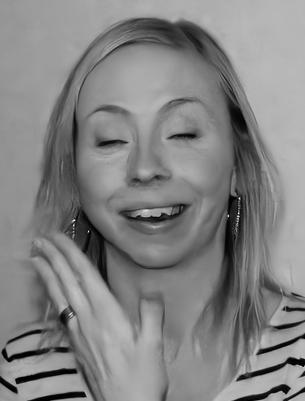} &
    		\includegraphics[width = 0.115\textwidth]{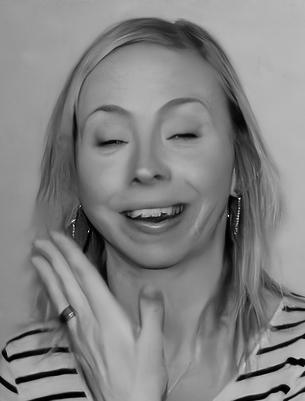} \\
    		
    	\end{tabular}  
    	\begin{tabular}{c@{\hskip 0.005\textwidth}c@{\hskip 0.005\textwidth}c@{\hskip 0.005\textwidth}c@{\hskip 0.005\textwidth}c@{\hskip 0.005\textwidth}c@{\hskip 0.005\textwidth}c@{\hskip 0.005\textwidth}c@{\hskip 0.005\textwidth}c@{\hskip 0.005\textwidth}c@{\hskip 0.005\textwidth}c@{\hskip 0.005\textwidth}c@{\hskip 0.005\textwidth}c@{\hskip 0.005\textwidth}c@{\hskip 0.005\textwidth}c@{\hskip 0.005\textwidth}c}
    		
    		\includegraphics[width = 0.055\textwidth]{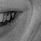} &
    		\includegraphics[width = 0.055\textwidth]{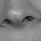} &
    		\includegraphics[width = 0.055\textwidth]{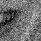} &
    		\includegraphics[width = 0.055\textwidth]{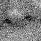} &
    		\includegraphics[width = 0.055\textwidth]{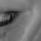} &
    		\includegraphics[width = 0.055\textwidth]{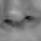} &
			\includegraphics[width = 0.055\textwidth]{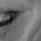} &
    		\includegraphics[width = 0.055\textwidth]{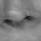} &
    		\includegraphics[width = 0.055\textwidth]{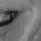} &
    		\includegraphics[width = 0.055\textwidth]{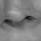} &
    		\includegraphics[width = 0.055\textwidth]{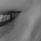} &
    		\includegraphics[width = 0.055\textwidth]{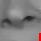} &            
    		\includegraphics[width = 0.055\textwidth]{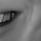} &
    		\includegraphics[width = 0.055\textwidth]{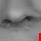} &                
    		\includegraphics[width = 0.055\textwidth]{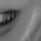} &
    		\includegraphics[width = 0.055\textwidth]{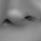} \\    	
            
    		\multicolumn{2}{c}{Ground truth}&
    		\multicolumn{2}{c}{Noisy image}&
    		\multicolumn{2}{c}{BM3D}&
    		\multicolumn{2}{c}{MLP}&
    		\multicolumn{2}{c}{TNRD}&
            \multicolumn{2}{c}{\ed{IRCNN}}&
            \multicolumn{2}{c}{\ed{IRCNN \textit{face}-specific}}&
    		\multicolumn{2}{c}{Our \textit{face}-specific}\\
            
    		\multicolumn{2}{c}{$$}&
    		\multicolumn{2}{c}{$$}&
    		\multicolumn{2}{c}{$31.63dB$} &
    		\multicolumn{2}{c}{$31.68dB$} &
    		\multicolumn{2}{c}{$31.73dB$} &
            \multicolumn{2}{c}{\ed{$32.17dB$}} &
            \multicolumn{2}{c}{\ed{$32.38dB$}} &
    		\multicolumn{2}{c}{$32.46dB$} \\     		
    		\smallskip
    	\end{tabular}    	

    	\begin{tabular}{c@{\hskip 0.005\textwidth}c@{\hskip 0.005\textwidth}c@{\hskip 0.005\textwidth}c@{\hskip 0.005\textwidth}c@{\hskip 0.005\textwidth}c@{\hskip 0.005\textwidth}c@{\hskip 0.005\textwidth}c}
    		
    		\includegraphics[width = 0.115\textwidth]{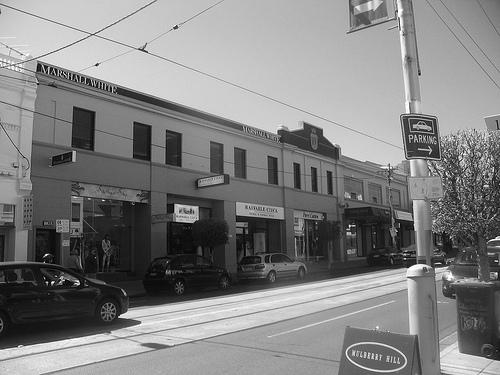} &
    		\includegraphics[width = 0.115\textwidth]{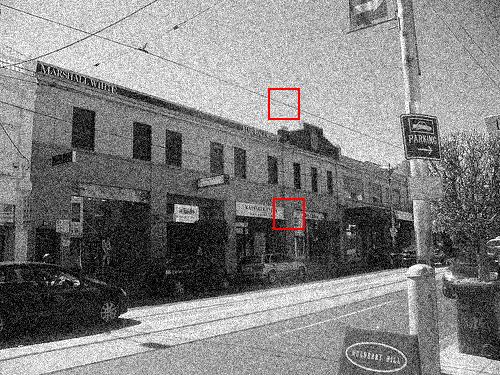} &
    		\includegraphics[width = 0.115\textwidth]{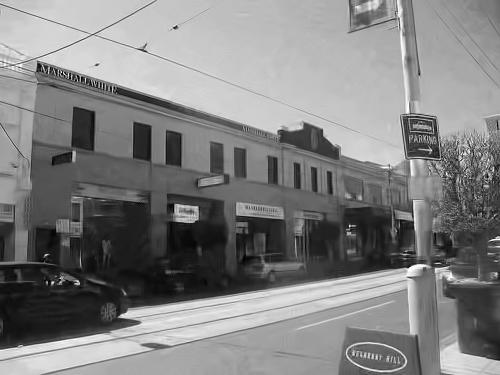} &
    		\includegraphics[width = 0.115\textwidth]{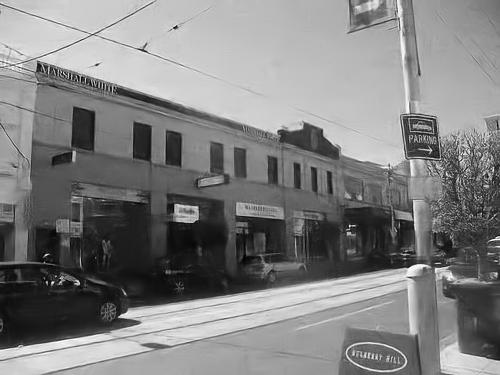} &
    		\includegraphics[width = 0.115\textwidth]{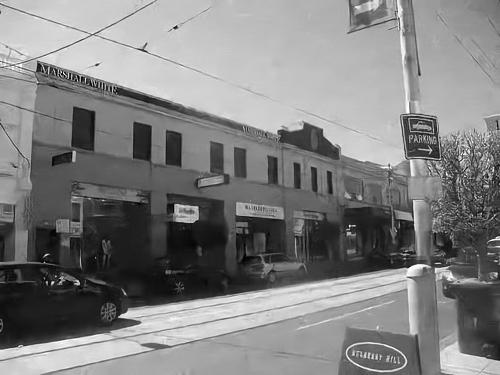} &
            \includegraphics[width = 0.115\textwidth]{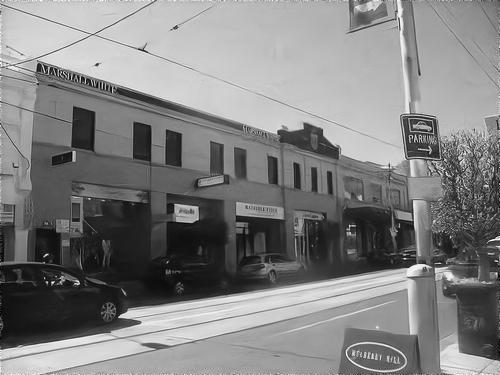} &
            \includegraphics[width = 0.115\textwidth]{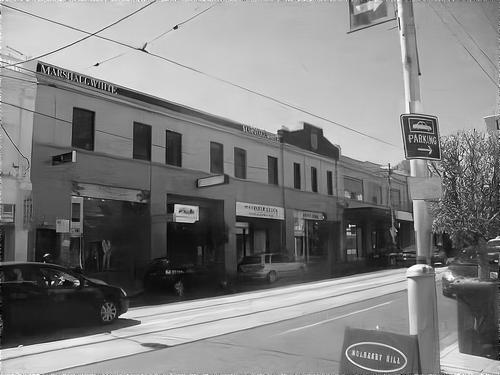} &
    		\includegraphics[width = 0.115\textwidth]{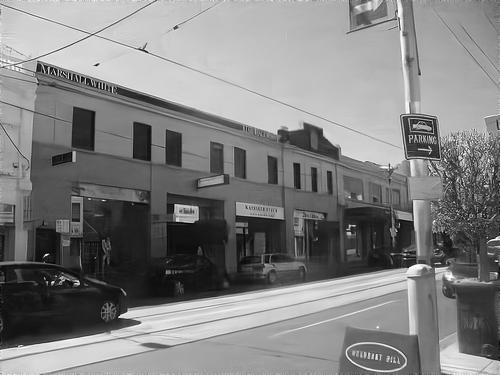} \\
    		
    	\end{tabular}  
    	\begin{tabular}{c@{\hskip 0.005\textwidth}c@{\hskip 0.005\textwidth}c@{\hskip 0.005\textwidth}c@{\hskip 0.005\textwidth}c@{\hskip 0.005\textwidth}c@{\hskip 0.005\textwidth}c@{\hskip 0.005\textwidth}c@{\hskip 0.005\textwidth}c@{\hskip 0.005\textwidth}c@{\hskip 0.005\textwidth}c@{\hskip 0.005\textwidth}c@{\hskip 0.005\textwidth}c@{\hskip 0.005\textwidth}c@{\hskip 0.005\textwidth}c@{\hskip 0.005\textwidth}c}
    		
    		\includegraphics[width = 0.055\textwidth]{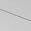} &
    		\includegraphics[width = 0.055\textwidth]{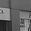} &
    		\includegraphics[width = 0.055\textwidth]{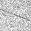} &
    		\includegraphics[width = 0.055\textwidth]{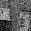} &
    		\includegraphics[width = 0.055\textwidth]{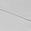} &
    		\includegraphics[width = 0.055\textwidth]{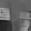} &
			\includegraphics[width = 0.055\textwidth]{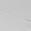} &
    		\includegraphics[width = 0.055\textwidth]{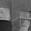} &
    		\includegraphics[width = 0.055\textwidth]{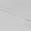} &
    		\includegraphics[width = 0.055\textwidth]{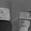} &
            \includegraphics[width = 0.055\textwidth]{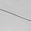} &
    		\includegraphics[width = 0.055\textwidth]{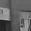} & 
            \includegraphics[width = 0.055\textwidth]{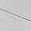} &
    		\includegraphics[width = 0.055\textwidth]{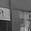} &             
    		\includegraphics[width = 0.055\textwidth]{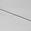} &
    		\includegraphics[width = 0.055\textwidth]{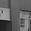} \\
    		
    		\multicolumn{2}{c}{Ground truth}&
    		\multicolumn{2}{c}{Noisy image}&
    		\multicolumn{2}{c}{BM3D}&
    		\multicolumn{2}{c}{MLP}&
    		\multicolumn{2}{c}{TNRD}&
            \multicolumn{2}{c}{\ed{IRCNN}}&
            \multicolumn{2}{c}{\ed{IRCNN \textit{street}-specific}}&
    		\multicolumn{2}{c}{Our \textit{street}-specific} \\
    		
    		\multicolumn{2}{c}{$$}&
    		\multicolumn{2}{c}{$$}&
    		\multicolumn{2}{c}{$28.67dB$}&
     		\multicolumn{2}{c}{$28.71dB$}&
    		\multicolumn{2}{c}{$28.76dB$}&
            \multicolumn{2}{c}{\ed{$29.67dB$}}&
             \multicolumn{2}{c}{\ed{$29.81dB$}}&
    		\multicolumn{2}{c}{$29.76dB$}\\  
			
    		\smallskip        
    	\end{tabular}

    \begin{tabular}{c@{\hskip 0.005\textwidth}c@{\hskip 0.005\textwidth}c@{\hskip 0.005\textwidth}c@{\hskip 0.005\textwidth}c@{\hskip 0.005\textwidth}c@{\hskip 0.005\textwidth}c@{\hskip 0.005\textwidth}c}
    
		\includegraphics[width = 0.115\textwidth]{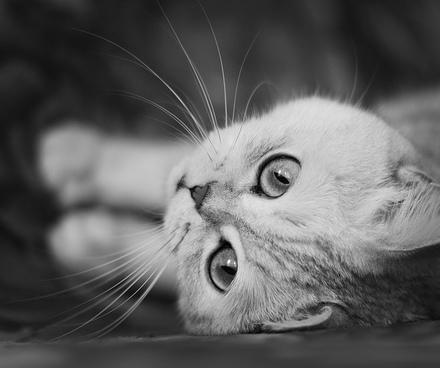} &
		\includegraphics[width = 0.115\textwidth]{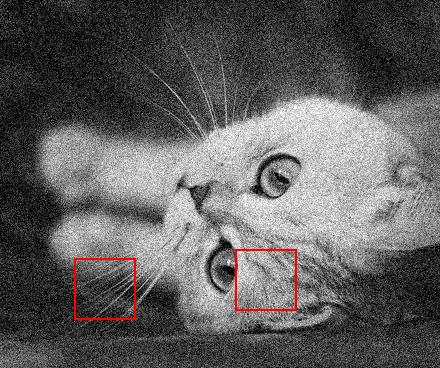} &
        \includegraphics[width = 0.115\textwidth]{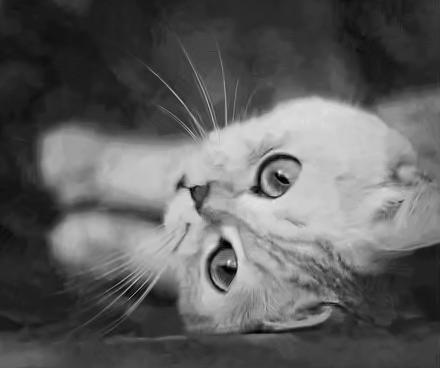} &
		\includegraphics[width = 0.115\textwidth]{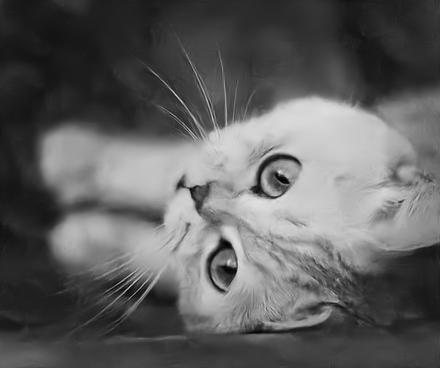} &
        \includegraphics[width = 0.115\textwidth]{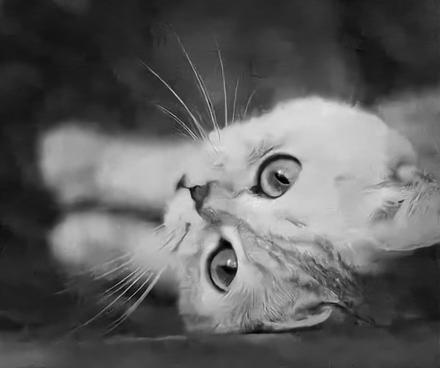} &
        \includegraphics[width = 0.115\textwidth]{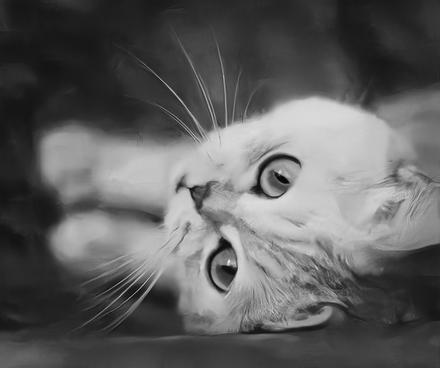} &
        \includegraphics[width = 0.115\textwidth]{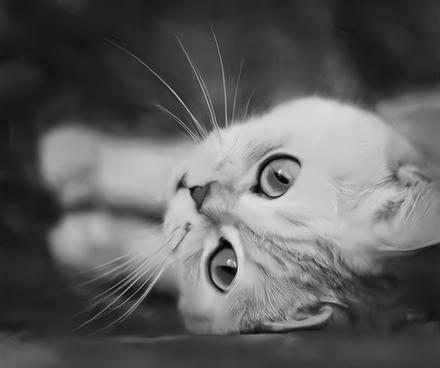} &
		\includegraphics[width = 0.115\textwidth]{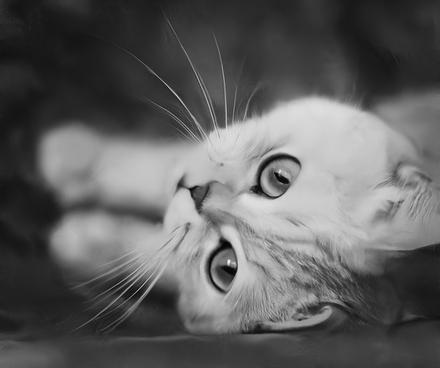} \\
 
 	\end{tabular}  
   	\begin{tabular}{c@{\hskip 0.005\textwidth}c@{\hskip 0.005\textwidth}c@{\hskip 0.005\textwidth}c@{\hskip 0.005\textwidth}c@{\hskip 0.005\textwidth}c@{\hskip 0.005\textwidth}c@{\hskip 0.005\textwidth}c@{\hskip 0.005\textwidth}c@{\hskip 0.005\textwidth}c@{\hskip 0.005\textwidth}c@{\hskip 0.005\textwidth}c@{\hskip 0.005\textwidth}c@{\hskip 0.005\textwidth}c@{\hskip 0.005\textwidth}c@{\hskip 0.005\textwidth}c}
    
		\includegraphics[width = 0.055\textwidth]{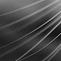} &
        \includegraphics[width = 0.055\textwidth]{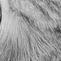} &
		\includegraphics[width = 0.055\textwidth]{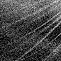} &
        \includegraphics[width = 0.055\textwidth]{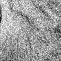} &
   		\includegraphics[width = 0.055\textwidth]{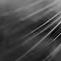} &
        \includegraphics[width = 0.055\textwidth]{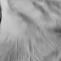} &
		\includegraphics[width = 0.055\textwidth]{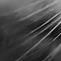} &
        \includegraphics[width = 0.055\textwidth]{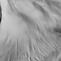} &
		\includegraphics[width = 0.055\textwidth]{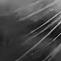} &
        \includegraphics[width = 0.055\textwidth]{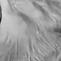} &
        \includegraphics[width = 0.055\textwidth]{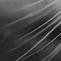} &
        \includegraphics[width = 0.055\textwidth]{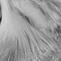} & 
        \includegraphics[width = 0.055\textwidth]{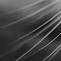} &
        \includegraphics[width = 0.055\textwidth]{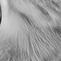} & 
		\includegraphics[width = 0.055\textwidth]{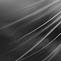} &
        \includegraphics[width = 0.055\textwidth]{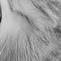} \\  
        
        \multicolumn{2}{c}{Ground truth}&
        \multicolumn{2}{c}{Noisy image}&
        \multicolumn{2}{c}{BM3D}& 
        \multicolumn{2}{c}{MLP}&
        \multicolumn{2}{c}{TNRD}&
        \multicolumn{2}{c}{\ed{IRCNN}}&
         \multicolumn{2}{c}{\ed{IRCNN \textit{pet}-specific}} &
        \multicolumn{2}{c}{Our \textit{pet}-specific}\\  
                
        \multicolumn{2}{c}{$$}&
        \multicolumn{2}{c}{$$}&
        \multicolumn{2}{c}{$31.80dB$}&  
        \multicolumn{2}{c}{$29.97dB$ }&
        \multicolumn{2}{c}{$31.28dB$ }&
        \multicolumn{2}{c}{\ed{$31.99dB$}}&
        \multicolumn{2}{c}{\ed{$32.34dB$}}&
        \multicolumn{2}{c}{$32.37dB$}\\   
    \smallskip
	\end{tabular}

    \caption{\ed{\textbf{Gaussian ($\sigma=25$) denoising examples from ImageNet.} We present four class-agnostic methods and two class-aware ones. PSNR values appear below each of the images.}}
\label{fig_large_gaussian_imagenet}
\end{figure*}

\begin{figure*}[h]
	\centering   
    \footnotesize
    \begin{tabular}{c@{\hskip 0.01\textwidth}c@{\hskip 0.01\textwidth}c@{\hskip 0.01\textwidth}c@{\hskip 0.01\textwidth}c@{\hskip 0.01\textwidth}c}    
            Ground Truth & I+VST+BM3D & \ed{IRCNN class-agnostic} & \ed{IRCNN class-specific} & Our class-agnostic & Our class-specific\\
		\includegraphics[width =  0.15\textwidth]{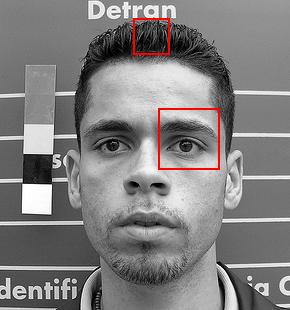} &
        \includegraphics[width =  0.15\textwidth]{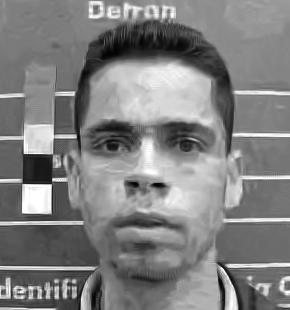} &
        \includegraphics[width =  0.15\textwidth]{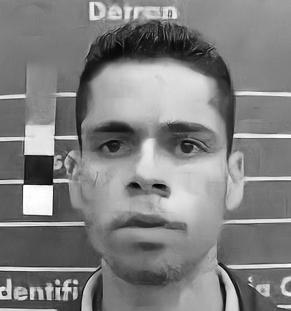}  &
        \includegraphics[width =  0.15\textwidth]{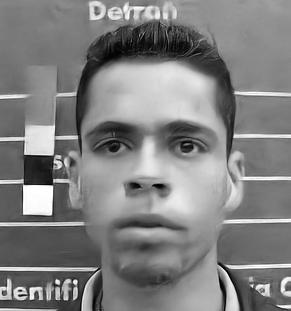}  &
        \includegraphics[width =  0.15\textwidth]{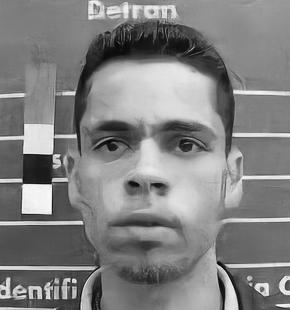}  &
        \includegraphics[width =  0.15\textwidth]{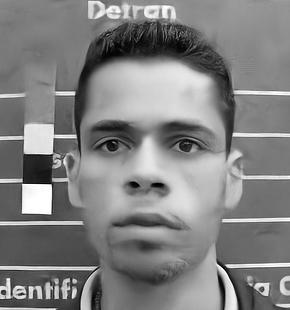} \\  
\end{tabular}  
   	\begin{tabular}{c@{\hskip 0.005\textwidth}c@{\hskip 0.01\textwidth}c@{\hskip 0.005\textwidth}c@{\hskip 0.01\textwidth}c@{\hskip 0.005\textwidth}c@{\hskip 0.01\textwidth}c@{\hskip 0.005\textwidth}c@{\hskip 0.01\textwidth}c@{\hskip 0.005\textwidth}c@{\hskip 0.01\textwidth}c@{\hskip 0.005\textwidth}c}
    
		\includegraphics[width = 0.0725\textwidth]{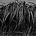} &
        \includegraphics[width = 0.0725\textwidth]{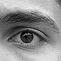} &
		\includegraphics[width = 0.0725\textwidth]{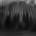} &
        \includegraphics[width = 0.0725\textwidth]{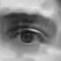} &
        \includegraphics[width = 0.0725\textwidth]{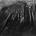}  &
        \includegraphics[width = 0.0725\textwidth]{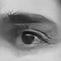}  &
        \includegraphics[width = 0.0725\textwidth]{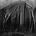}  &        
        \includegraphics[width = 0.0725\textwidth]{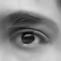}  &
		\includegraphics[width = 0.0725\textwidth]{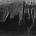} &
        \includegraphics[width = 0.0725\textwidth]{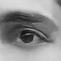} &
        \includegraphics[width = 0.0725\textwidth]{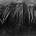} &
        \includegraphics[width = 0.0725\textwidth]{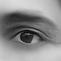} \\
   
		 \multicolumn{2}{c}{\smallskip} &
         \multicolumn{2}{c}{$25.85$ dB} & 
         \multicolumn{2}{c}{\ed{$26.51$ dB}} &
         \multicolumn{2}{c}{\ed{$26.64$ dB}} &
         \multicolumn{2}{c}{$26.79$ dB} & 
         \multicolumn{2}{c}{$26.96$ dB} \\
	\end{tabular}  

    \begin{tabular}{c@{\hskip 0.01\textwidth}c@{\hskip 0.01\textwidth}c@{\hskip 0.01\textwidth}c@{\hskip 0.01\textwidth}c@{\hskip 0.01\textwidth}c}  

		\includegraphics[width =  0.15\textwidth]{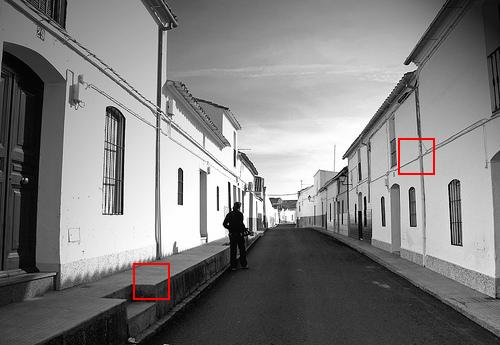} &
        \includegraphics[width =  0.15\textwidth]{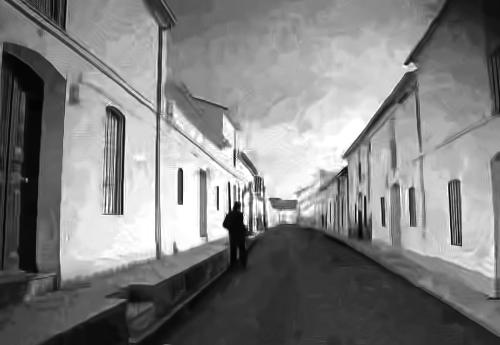} &
        \includegraphics[width =  0.15\textwidth]{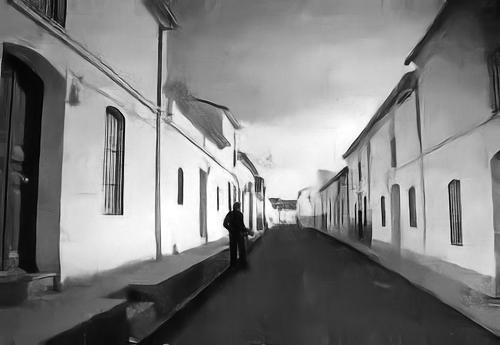} &
        \includegraphics[width =  0.15\textwidth]{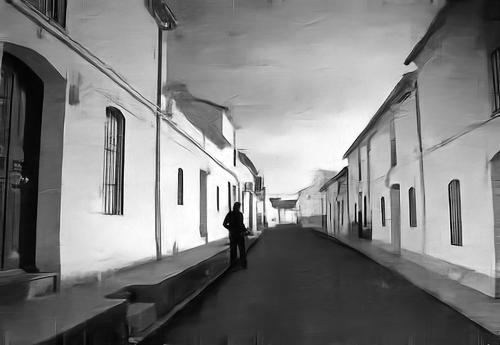} &
        \includegraphics[width =  0.15\textwidth]{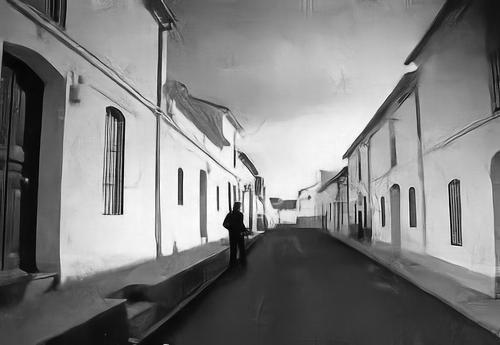}  &
        \includegraphics[width =  0.15\textwidth]{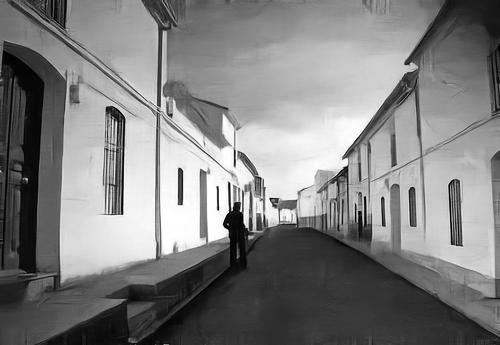} \\ 
	\end{tabular}  
   	\begin{tabular}{c@{\hskip 0.005\textwidth}c@{\hskip 0.01\textwidth}c@{\hskip 0.005\textwidth}c@{\hskip 0.01\textwidth}c@{\hskip 0.005\textwidth}c@{\hskip 0.01\textwidth}c@{\hskip 0.005\textwidth}c@{\hskip 0.01\textwidth}c@{\hskip 0.005\textwidth}c@{\hskip 0.01\textwidth}c@{\hskip 0.005\textwidth}c}
    
		\includegraphics[width = 0.0725\textwidth]{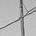} &
        \includegraphics[width = 0.0725\textwidth]{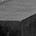} &
		\includegraphics[width = 0.0725\textwidth]{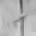} &
        \includegraphics[width = 0.0725\textwidth]{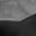} &
        \includegraphics[width = 0.0725\textwidth]{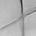}  &
        \includegraphics[width = 0.0725\textwidth]{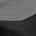}  &
        \includegraphics[width = 0.0725\textwidth]{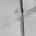}  &        
        \includegraphics[width = 0.0725\textwidth]{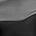}  &        
		\includegraphics[width = 0.0725\textwidth]{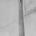} &
        \includegraphics[width = 0.0725\textwidth]{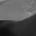} &
        \includegraphics[width = 0.0725\textwidth]{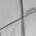} &
        \includegraphics[width = 0.0725\textwidth]{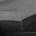} \\
   
		 \multicolumn{2}{c}{\smallskip} &
         \multicolumn{2}{c}{$25.56$ dB} & 
         \multicolumn{2}{c}{\ed{$26.31$ dB}} &
         \multicolumn{2}{c}{\ed{$26.56$ dB}} &
         \multicolumn{2}{c}{$26.47$ dB} & 
         \multicolumn{2}{c}{$26.75$ dB} \\
	\end{tabular}  

    \begin{tabular}{c@{\hskip 0.01\textwidth}c@{\hskip 0.01\textwidth}c@{\hskip 0.01\textwidth}c@{\hskip 0.01\textwidth}c@{\hskip 0.01\textwidth}c}   
		\includegraphics[width =  0.15\textwidth]{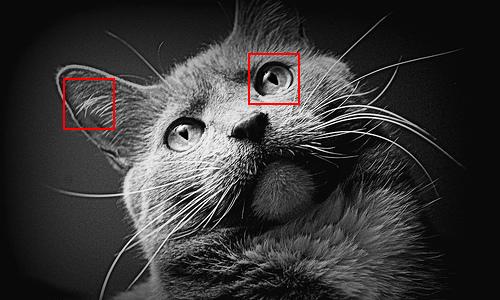} &
        \includegraphics[width =  0.15\textwidth]{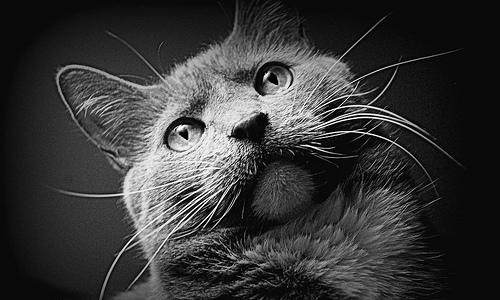} &
        \includegraphics[width =  0.15\textwidth]{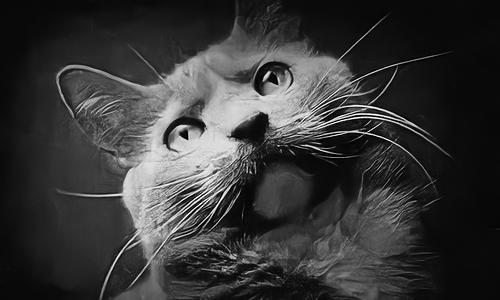} &
        \includegraphics[width =  0.15\textwidth]{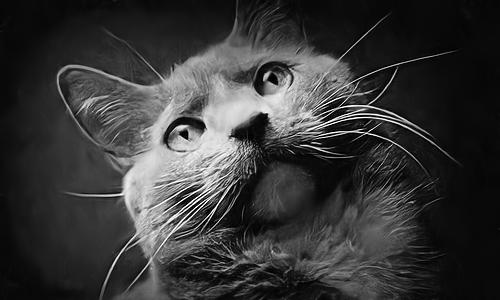} &        
        \includegraphics[width =  0.15\textwidth]{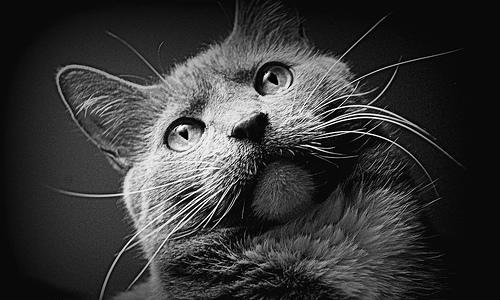}  &
        \includegraphics[width =  0.15\textwidth]{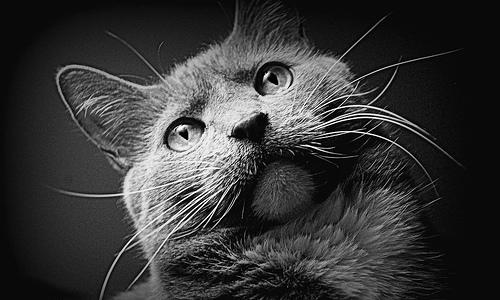}  \\ 
	\end{tabular}  
   	\begin{tabular}{c@{\hskip 0.005\textwidth}c@{\hskip 0.01\textwidth}c@{\hskip 0.005\textwidth}c@{\hskip 0.01\textwidth}c@{\hskip 0.005\textwidth}c@{\hskip 0.01\textwidth}c@{\hskip 0.005\textwidth}c@{\hskip 0.01\textwidth}c@{\hskip 0.005\textwidth}c@{\hskip 0.01\textwidth}c@{\hskip 0.005\textwidth}c}
    
		\includegraphics[width = 0.0725\textwidth]{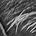} &
        \includegraphics[width = 0.0725\textwidth]{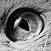} &
		\includegraphics[width = 0.0725\textwidth]{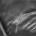} &
        \includegraphics[width = 0.0725\textwidth]{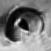} &
        \includegraphics[width = 0.0725\textwidth]{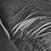}  &
        \includegraphics[width = 0.0725\textwidth]{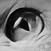}  &
        \includegraphics[width = 0.0725\textwidth]{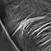}  &        
        \includegraphics[width = 0.0725\textwidth]{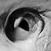}  &         
		\includegraphics[width = 0.0725\textwidth]{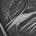} &
        \includegraphics[width = 0.0725\textwidth]{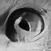} &
        \includegraphics[width = 0.0725\textwidth]{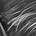} &
        \includegraphics[width = 0.0725\textwidth]{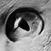} \\
   
		 \multicolumn{2}{c}{\smallskip} &
         \multicolumn{2}{c}{$23.53$ dB} & 
         \multicolumn{2}{c}{\ed{$24.48$ dB}} & 
         \multicolumn{2}{c}{\ed{$24.66$ dB}} & 
         \multicolumn{2}{c}{$24.60$ dB} & 
         \multicolumn{2}{c}{$24.78$ dB} \\
	\end{tabular}   
    \caption{\ed{\textbf{Poisson (peak$=8$) denoising  examples from ImageNet.} PSNR values appear below each of the images.}}
\label{fig_large_poisson_imagenet}
\end{figure*}

\section{Discussion}
\label{sec_discussion}

We introduced a novel fully convolutional neural network for Gaussian and Poisson image denoising with performance often exceeding the state-of-the-art. Our network architecture is inherently transparent, in the sense that all intermediate layers are extracted and directly contribute to the noise estimation.
This is helpful in overcoming the ``black-box'' nature of neural networks as it allows us to gain several interesting insights on the denoising process.

In addition, we showed that using a classifier to route the input image to a class-specific network is preferable to a universal filter, achieving an additional boost of up to $0.4$dB PSNR over a class-agnostic denoiser. 
That said, the decision to split according to a global semantic class as done in this paper was made due to the immediate availability of labeled data and off-the-shelf classifiers that are relatively resilient to noise. Yet, this splitting scheme may be sub-optimal and other choices for data partitioning could be made. In particular, the splitting scheme could be learned automatically by incorporating it into the network architecture. It would then lose its simple semantic interpretation, and instead yield abstract classes, perhaps varying across different regions of the image. 

\medskip 

\noindent\textbf{Limitations.} In most denoising methods, some set of parameters is adjusted according to the noise type and level. While in algorithms such as BM3D this set is small and requires a small amount of memory, in deep learning methods like MLP, TNRD, IRCNN and ours, it is much larger. Furthermore, in our class aware denoising framework, it grows linearly with the number of classes. \ed{Clearly, being specific to one class deteriorates the performance to the other ones as shown in Fig.~\ref{fig_using_wrong_denoiser} and therefore a denoiser is required per each class}. 
Integrating the splitting scheme into the network might mitigate this issue. \ed{Note also that when a class specific denoiser is trained, one should verify that the variety of images used for training is large and diverse enough to avoid overfitting.}

Another limitation is a decrease in the performance gain of our method with the decrease of the peak value (i.e., in the presence of stronger noise). A possible explanation for this phenomenon might lie in the fact that extremely noisy images no longer resemble natural ones and small convolution kernels have a harder time learning such patterns. A different architecture might be more adequate for such scenarios.

\ed{An additional limitation we leave to be solved in future work is the ability to handle the noise in real images (see for example \cite{plotz2017benchmarking}), where the noise is of a mixed type and its level cannot be assumed to be known prior to the denoising process.}

\section{Acknowledgments}
This research is partially supported by ERC-StG RAPID PI Bronstein and ERC-StG SPADE PI Giryes.

\appendices
\section{List of Videos}
\label{sec:video_list}
The list of videos used for training, validation, and testing of the proposed video denoising network is provided below. 

\subsection{Video training set}
\begin{enumerate}

\item Extreme Bungy Jumping with Cliff Jump Shenanigans! 
4K! \url{youtu.be/l9m4cW2yxy0}

\item Motorized Drift Trike and Blokart in 4K! \url{youtu.be/Mf2wAEYtrlo}

\item Robotic Dolphin and Flying Water Car - In 4K! 
\url{youtu.be/dkpzgS9rdTk}

\item World's Best Basketball Freestyle Dunks - 
4k \url{youtu.be/5rd_WwD_aOQ}

\item Hoverboard in Real Life! In 4K! \url{youtu.be/gMaDhkNJA2g}

\item Barefoot Skiing behind Airplane in 4K - Insane!
\url{youtu.be/vdTrr_VRKgU}

\item Assassin's Creed Unity Meets Parkour in Real Life - 4K!  \url{youtu.be/S8b1zWOgOKA}

\item Top 5 moments in Roland Garros - Best rallies \url{youtu.be/a7O54L_vcyo}

\item The Nexus interrupt the main event and reap destruction  Raw 
\url{youtu.be/vVVtqoqzgNw}

\item May Win Compilation 2014 MW \url{youtu.be/UgY9eR3Zd0k}

\item Insane Human Skeeball! \url{youtu.be/nyMgJ3z8U9I}

\item GoPro  Sketchy Cornice Rappel \url{youtu.be/7fGGEwY3qVk}

\item Dubai in 4K - City of Gold \url{youtu.be/SLaYPmhse30}

\item Sky Racers! - Dubai! 
4K
\url{youtu.be/cTQvYxELaFI}

\item GoPro HD HERO camera  Base Jump Movie \url{youtu.be/mRzhBkZNQFI}

\item GoPro  Combing Valparaiso's Hills \url{youtu.be/cN-YTcSnE6c}

\item GoPro Moab Towers \& Magic Backpacks \url{youtu.be/fVcV9ItdZ8w}

\end{enumerate}

\subsection{Video validation set}
\begin{enumerate}

\item DevinSupertramp - Best of 2014 \url{youtu.be/bpz-Pq61DLw}

\item 41-Man Battle Royal for a Championship Match 
\url{youtu.be/n3PswaGIZt4}

\end{enumerate}

\subsection{Video test set}
\begin{enumerate}

\item Sandboarding Supertramp Style - 
4K!   
\url{youtu.be/0ENviLq-XfI}

\item World's Craziest Teeterboard Flips - Streaks Show in 4K! \url{youtu.be/95f4kb5XR3U}

\item John Cena \& Randy Orton battle the entire Raw roster  Raw 
\url{youtu.be/ndgVcE7w0uo}

\item GoPro  Rocky Cliff Huck In The French Alps \url{youtu.be/lEOHQQBZbJw}

\item GoPro  The Untold Story of Ryan Villopoto 
\url{youtu.be/4oJc8IF2Gpc}

\item GoPro  Mountain Bike River Jump \url{youtu.be/G4EAswzxKJs}

\end{enumerate}

%

%
%

\ifCLASSOPTIONcaptionsoff
  \newpage
\fi

\bibliographystyle{IEEEtran}
\bibliography{bib}



%

\end{document}